\documentclass{article}

 \usepackage[preprint,nonatbib]{neurips_2026}

\usepackage[utf8]{inputenc}      
\usepackage[T1]{fontenc}         
\usepackage{hyperref}            
\usepackage{url}                 
\usepackage{booktabs}            
\usepackage{amsfonts}            
\usepackage{nicefrac}            
\usepackage{microtype}           
\usepackage{graphicx}            
\usepackage{subcaption}          
\usepackage{multirow}            
\usepackage{makecell}            
\usepackage{fontawesome5}        
\usepackage{amsmath, amssymb}    
\usepackage{amsthm, mathtools}   
\usepackage{bm}                  
\usepackage{helvet}              
\usepackage{enumitem}            
\usepackage{titletoc}            
\usepackage{etoolbox}            
\usepackage{tikz}                
\usepackage{algorithm}           
\usepackage{algorithmic}         
\usepackage{tcolorbox}           
\usepackage{tabularx}            

\usepackage[table,dvipsnames]{xcolor}        
\usepackage[skip=0.5\baselineskip]{caption}  
\usepackage[capitalize,noabbrev]{cleveref}   

\PassOptionsToPackage{colorlinks=true, linkcolor=blue}{hyperref}

\BeforeBeginEnvironment{tikzpicture}{\sffamily} 

\DeclareMathOperator*{\argmax}{argmax}

\usepackage[numbers]{natbib}

\newcommand{\appendixtoc}{
  \clearpage
  \startcontents[appendix]
  
  \begin{center}
    {\huge\bfseries \textsc{Zatom-1} Appendices}
    \vspace{8pt}
    \hrule height 1.5pt
    \vspace{16pt}
  \end{center}

  \titlecontents{section}[2em] 
    {\addvspace{12pt}\bfseries\sffamily\large} 
    {\contentslabel{2em}} 
    {} 
    {\titlerule*[0.6pc]{.}\contentspage} 
    [\addvspace{2pt}]

  \titlecontents{subsection}[4.5em] 
    {\addvspace{2pt}\sffamily\small} 
    {\contentslabel{2.5em}}
    {}
    {\titlerule*[0.6pc]{.}\contentspage}
    
  \printcontents[appendix]{}{1}{\setcounter{tocdepth}{2}}
  
  \vspace{24pt}
  \hrule height 0.5pt
}

\usetikzlibrary{backgrounds, fit, positioning, arrows.meta, calc}
\tikzset{
    input/.style={
        rectangle,
        rounded corners,
        draw=blue!80,
        fill=blue!10,
        minimum width=2.5cm,
        minimum height=0.8cm,
        align=center
    },
    block/.style={
        rectangle,
        draw=black,
        fill=orange!20,
        minimum width=3cm,
        minimum height=1cm,
        align=center,
        text width=2.8cm
    },
    aux_block/.style={
        rectangle,
        draw=black,
        fill=green!20,
        minimum width=2.5cm,
        minimum height=0.8cm,
        align=center,
        text width=2.3cm
    },
    output/.style={
        rectangle,
        rounded corners,
        draw=red!80,
        fill=red!10,
        minimum width=2.5cm,
        minimum height=1cm,
        align=center
    },
    arrow/.style={
        ->,
        thick,
        >=stealth
    }
}

\definecolor{inputgray}{HTML}{EAEAEA}
\definecolor{bordergray}{HTML}{AAAAAA}

\definecolor{processblue}{HTML}{D6EAF8}
\definecolor{borderblue}{HTML}{5499C7}

\definecolor{modelpurple}{HTML}{E8DAEF}
\definecolor{borderpurple}{HTML}{8E44AD}

\definecolor{taskteal}{HTML}{D1F2EB}
\definecolor{borderteal}{HTML}{48C9B0}

\definecolor{outputgreen}{HTML}{D5F5E3}
\definecolor{borderoutputgreen}{HTML}{58D68D}

\definecolor{slate-grey}{RGB}{112, 128, 144}

\definecolor{commentcolor}{HTML}{C46F21}

\definecolor{frontiertitlebg}{HTML}{E6EEF4} 

\begin{document}


\begin{tcolorbox}[
    colback=frontiertitlebg,
    colframe=frontiertitlebg,
    arc=4mm,          
    boxrule=0pt,      
    left=15pt,        
    right=15pt,
    top=15pt,
    bottom=15pt
]

\begin{flushleft}
{\LARGE \bfseries \textsc{Zatom-1}: Towards a Multimodal Foundation Model for 3D Molecules and Materials \par}
\vspace{1em}

{\large \textbf{Alex Morehead}$^{1,2,*}$, \textbf{Miruna Cretu}$^{3,*}$, \textbf{Antonia Panescu}$^{4,*}$, \textbf{Rishabh Anand}$^{4,*}$, \textbf{Maurice Weiler}$^{5,*}$, \textbf{Tynan Perez}$^{5,*}$, \textbf{Samuel Blau}$^1$, \textbf{Steven Farrell}$^1$, \textbf{Wahid Bhimji}$^1$, \textbf{Anubhav Jain}$^1$, \textbf{Hrushikesh Sahasrabuddhe}$^{1,6}$, \textbf{Pietro Liò}$^3$, \textbf{Tommi Jaakkola}$^5$, \textbf{Rafael Gómez-Bombarelli}$^5$, \textbf{Rex Ying}$^{4,\dagger}$, \textbf{N. Benjamin Erichson}$^{1,2,\dagger}$, \textbf{Michael W. Mahoney}$^{6,1,2,\dagger}$ \par}
\vspace{0.8em}

{\small
$^1$LBNL \quad $^2$ICSI \quad $^3$University of Cambridge \quad $^4$Yale University \quad $^5$MIT \quad $^6$UC Berkeley \par
\vspace{0.3em}
$^*$Equal contribution \quad $^\dagger$Equal advising \par
}
\end{flushleft}

\vspace{1em}

\noindent General-purpose 3D modeling in chemistry encompasses molecules and materials, requiring both generative and predictive capabilities. However, most existing AI approaches are optimized for a single domain (molecules or materials) and a single task (generation or prediction), which limits representation sharing and transfer. We introduce \textsc{Zatom-1}, a cross-domain, general-purpose model architecture that unifies generative and predictive learning of 3D molecules and materials. \textsc{Zatom-1} is a deliberately simplified Transformer trained with a multimodal flow matching objective that jointly models discrete atom types and continuous 3D geometries. This approach supports scalable pretraining with predictable gains as model capacity increases, while enabling fast and stable sampling. We use cross-domain generative pretraining as a universal initialization for downstream multi-task prediction of properties, energies, and forces. Empirically, \textsc{Zatom-1} outperforms or competes with specialized baselines on both multi-task generative and predictive benchmarks in data-controlled settings, while improving generative inference speed by more than an order of magnitude. Our experiments demonstrate positive predictive transfer between data domains from joint generative pretraining: modeling materials during generative pretraining improves molecular property prediction accuracy. Open-source code and model weights are freely available.

\vspace{1.5em}

\noindent{\small
\textbf{Correspondence:} AM: \texttt{acmwhb@lbl.gov}; MWM: \texttt{mmahoney@stat.berkeley.edu} \\
\textbf{Code:} \url{https://github.com/Zatom-AI/zatom} \\
\textbf{Keywords:} 3D Molecules, Materials Design, Foundation Models, Multimodal Flow Matching
}

\end{tcolorbox}

\vspace{2em}



\section{Introduction}
\label{section:introduction}

Life, materials, and many modern functions, ranging from small molecule therapeutics to consumer electronic devices, are driven by chemistry. However, navigating the design space of chemical systems for high-impact downstream applications is notoriously time-consuming and error-prone \citep{blanco2023role, venkatasubramanian2019promise, wang2023scientific}. As such, developing our understanding of chemical systems with artificial intelligence (AI) methods, in particular foundation models equipped with neural scaling \citep{bommasani2021opportunities, kaplan2020scaling, frey2023neural}, represents a promising area of investment for deep learning research, given that related computational techniques have recently led to significant breakthroughs across adjacent scientific fields \citep{jumper2021highly, abramson2024accurate, hayes2025simulating}.

Recently, deep learning has begun to unify the generation of new 3D molecules and materials \citep{cheng2024response, joshi2025allatom, zhang2025unigenx, lu2026unified}, and several specialized prediction models currently exist for advanced molecule and material property prediction \citep{liao2023equiformer, qu2024escaip, perez2026self}. 
Nonetheless, no prior works offer a general-purpose model capable of both generative modeling and representation learning of 3D molecules and materials via cross-domain pretraining. To build such a unified model, several design choices must be addressed, including how to tokenize its inputs as well as what and how to learn using these tokens.

\begin{figure*}[!t]
    \centering
    \resizebox{\textwidth}{!}{%
        \input{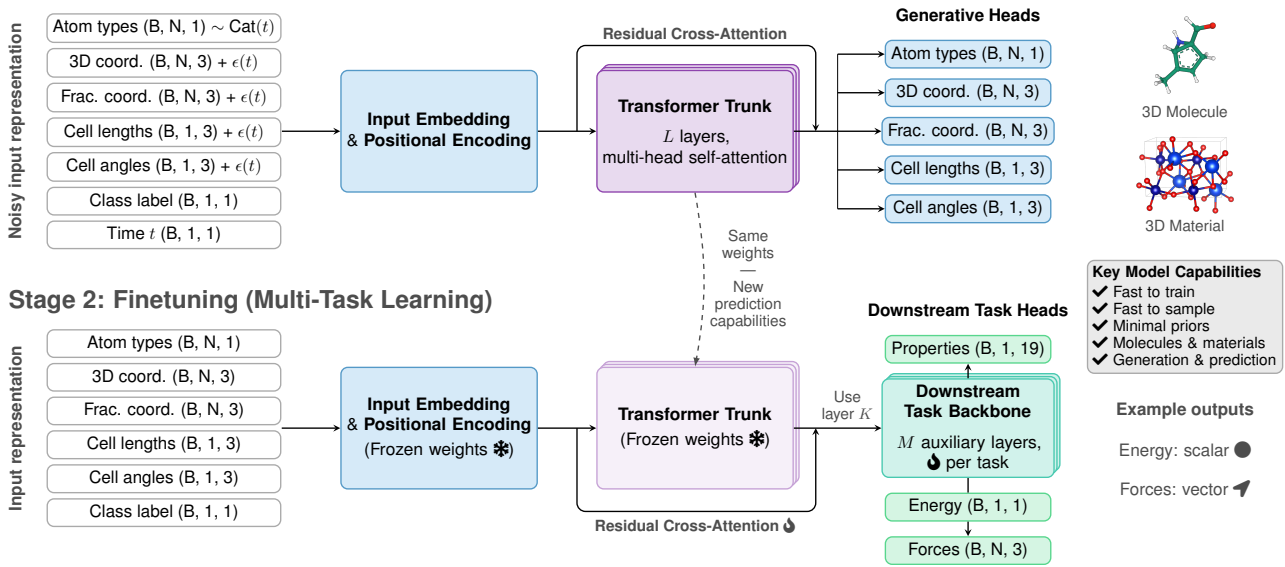}
    } 
    \caption{\textbf{The \textsc{Zatom-1} architecture for 3D molecules and materials.} A unified Transformer trunk processes (noisy) atomic, structural, and conditional inputs. The model features two training stages: (1) multimodal flow pretraining using the final trunk layer $L$'s representations to create new molecules or materials; and (2) multi-task finetuning of additional downstream task backbones and heads that use a specified trunk layer $K$'s representations to predict properties, energies, and forces.
    }
    \label{figure:zatom_1_overview}
\end{figure*}

Inspired by previous research on scientific foundation models \citep{subramanian2023scifm, abramson2024accurate} and neural scaling \citep{subramanian2026neural}, we argue for the importance of tokenizing atomistic systems at the fundamental level of atomic granularity: 3D atom coordinates and element types. The next question is what to learn with these tokens. Previous works have shown that image and video diffusion models, employing standardized deep learning architectures, can learn rich embeddings for downstream prediction tasks in a self-supervised manner \citep{li2023image, velez2025video} without assuming a canonical token ordering, as is common in autoregressive models \citep{cheng2025scalable}. In addition, scientific machine learning (SciML) is increasingly adopting standardized deep learning architectures to train highly scalable models \citep{abramson2024accurate, kreiman2025transformers,wadell2025foundation}. As such, it is clear that non-autoregressive generative modeling, due to its self-supervised (i.e., label-free) nature, has the potential to serve as a powerful and flexible pretraining method for standardized atomistic model architectures.

In this spirit, we propose a question: \textit{How much chemical knowledge can we represent with a 3D generative model?} \textbf{\textsc{Zatom-1}}, our answer to this question, is illustrated in \cref{figure:zatom_1_overview}.
It is a general-purpose, multimodal flow architecture for 3D molecules and materials, based on these ideas:
\begin{enumerate}
    \item \textit{Ambient all-atom generative modeling.}
    We represent 3D molecules and materials in $\mathbb{R}^3$ and perform generative flow matching directly within this Euclidean space. Importantly, with this approach, no pretrained autoencoders are necessary, e.g., for latent diffusion \citep{joshi2025allatom}, delivering notable speed-ups in model training and inference.
    \item \textit{Representation learning using generative models.}
    We frame generative modeling of 3D molecules and materials as an ideal pretraining task for chemical representation learning \citep{briq2025amazing} using modern, standardized Transformers \citep{vaswani2017attention}.
\end{enumerate}

\textbf{Contributions.} Based on these ideas, our main contributions in this work are as follows:
\begin{itemize} 
    \item We introduce \textsc{Zatom-1}, a general-purpose model architecture for diverse 3D chemical tasks, including \textit{end-to-end} generation of periodic 3D materials and non-periodic 3D molecules, multi-task (and for materials, zero-shot) property prediction, and energy and force modeling.
    \item \textsc{Zatom-1}: (\textbf{1}) achieves \textit{state-of-the-art} for 3D materials (molecule) generation with LeMat-GenBench (QM9 \& GEOM-Drugs); (\textbf{2}) provides state-of-the-art multi-task QM9 property predictions; (\textbf{3}) yields a 3x reduction in GPU training hours compared to latent diffusion \citep{joshi2025allatom}; and (\textbf{4}) demonstrates positive transfer learning through cross-domain pretraining.
    \item The \textsc{Zatom-1} architecture features a standardized Transformer with Query-Key Normalization \citep{wortsman2024smallscale}, Flash Attention \citep{dao2024flashattention}, and SwiGLU feed-forward networks \citep{shazeer2020glu}, yielding predictably improving performance through parameter scaling. Its design enables a \textbf{12.5x} speed-up over a 500M-parameter latent diffusion baseline \citep{joshi2025allatom}, with \textsc{Zatom-1} (300M) generating 10,000 samples in under 4 minutes on a single NVIDIA A100 GPU.
\end{itemize}


\section{\textsc{Zatom-1}}
\label{section:zatom_1}

In this section, we introduce the \textsc{Zatom-1} model architecture for generative modeling and representation learning of 3D molecules and materials. Namely, we describe its unified input representation and architectural features for such data types and characterize how we pretrain and finetune \textsc{Zatom-1} for a variety of tasks in 3D chemistry.

In particular, \textsc{Zatom-1} is trained in two stages: (1) generative pretraining for 3D molecule and material generation; and (2) predictive finetuning for energy, force, and property prediction tasks. 
For generative pretraining, we use conditional flow matching, a scalable training technique for normalizing flows \citep{lipman2023flow}. As a \textit{non-autoregressive} generative modeling paradigm \citep{nguyen2025oneflow, nielarge2025}, conditional flow matching trains a neural network to approximate a time-conditioned velocity field required to transport samples from a source distribution to a target distribution using ordinary differential equation (ODE) integration. Importantly, time integration is only required during sampling, not model training, offering scalable training speeds as a result \citep{morehead2026flow}.

To efficiently model the multimodal nature of 3D atomic systems, we use multimodal flow matching over a (discrete) atom type modality and four (continuous) geometric modalities, yielding a joint distribution of these modalities for generative modeling. 
Previous works for 3D materials generation have attempted multimodal flow matching over similar modality structures \citep{miller2024flowmm}. However, we argue that reliance on sparse graph neural network architectures, as well as hand-crafted generative priors, has hindered performance and training/inference speed for generative modeling in this domain \citep{krishnapriyan2021characterizing, qu2024escaip}. Instead, we adopt a standardized Transformer architecture and Gaussian multimodal flow matching for scalable joint modeling of 3D molecules and materials. We discuss related works and \textsc{Zatom-1}'s placement in the literature in detail in Appendix \ref{appendix:related_work}.

\subsection{Problem formulation}
\label{subsection:problem_formulation}

We aim to learn a joint generative model of 3D molecules and materials and to obtain rich neural embeddings of both data domains. Towards this end, we represent a 3D molecule or material with $N$ atoms using five unified modalities:
\begin{tcolorbox}[
    colframe=gray!50, 
    colback=gray!5, 
    coltitle=black, 
    fonttitle=\bfseries\small,
    boxrule=0.5pt, 
    arc=1mm, 
    title=Unified Input Representation,
    top=1mm, bottom=1mm, left=1mm, right=1mm
]
\small
\begin{tabularx}{\linewidth}{l l X r}
    $\bm{A} = \{a_i\}_{i=1}^N \in \mathbb{Z}^{1 \times N}$ & (Atom types) & & \\
    $\bm{X} = \{x_i\}_{i=1}^N \in \mathbb{R}^{3 \times N}$ & (3D coordinates) & \hfill & 
    \smash{\begin{tabular}[c]{@{}ll@{}} 
        $\bm{L}_{\text{len}} = \{l_{\text{len}}\} \in \mathbb{R}^{3 \times 1}$ & (Lattice lengths) \\
        $\bm{L}_{\text{ang}} = \{l_{\text{ang}}\} \in \mathbb{R}^{3 \times 1}$ & (Lattice angles)
    \end{tabular}} \\
    $\bm{F} = \{f_i\}_{i=1}^N \in [0, 1)^{3 \times N}$ & (Fractional coordinates) & & 
\end{tabularx}
\end{tcolorbox}
Atom types $\bm{A}$ are a discrete modality defined over the set of integers $\mathbb{Z}$, while 3D coordinates $\bm{X}$ are a continuous modality of $\mathbb{R}^3$ expressed in Angstroms. Fractional coordinates $\bm{F} = \bm{L}^{-1}\bm{X}$ reside in the range $[0, 1)$, and rotation-invariant lattice side lengths $\{l_\text{len}\} = \{a, b, c\} \in \mathbb{R}^{3 \times 1}$ and the three internal angles between them $\{l_\text{ang}\} = \{\alpha, \beta, \gamma\} \in [60^{\circ}, 120^{\circ}]^{3 \times 1}$ parametrize a parallelepiped $\bm{L} \in \mathbb{R}^{3 \times 3}$ \citep{grosse2004numerically} characterizing the repeating unit cell \citep{miller2024flowmm}. For the sake of numerical stability, during training and sampling, lattice lengths are normalized by $\sqrt[3]{N}$, and lattice angles are converted from degrees to radians. Further, during training and sampling, for non-periodic molecules the lattice lengths, angles, and fractional coordinate inputs are masked to null values $\varnothing$. For periodic materials 3D coordinate inputs are masked to null values $\varnothing$ to enforce materials rotation invariance by default (with training losses masked accordingly).

\begin{figure}[tb]
\centering
\resizebox{0.51\textwidth}{!}{
\begin{minipage}[t]{0.54\textwidth}
\begin{algorithm}[H]
\scriptsize
\caption{Pseudocode for \textsc{TFT} model $\mathcal{T}$}
\label{algorithm:tft_model_pseudocode}
\textbf{Input:}\quad Noisy 3D atoms $(\{a_i\}, \{x_i\}, \{f_i\}, l_\text{len}, l_\text{ang})$,\ \ class $c$,\ \ time $t$ \\[2pt]
\textbf{Output:}\ \ Denoised 3D atoms $(\{a'_i\}, \{x'_i\}, \{f'_i\}, l'_\text{len}, l'_\text{ang})$, \\
\phantom{\textbf{Output:}}\ \
predictions $\{p^\text{props'}, p^\text{energy'}, p_i^\text{forces'}\}$
\begin{algorithmic}[1]
  \STATE $\text{Lin}^\text{B},\ \text{Lin}^\text{NB}\ =\,\ \text{Linear(bias=True)},\ \text{Linear(bias=False)}$
  \item[] \textcolor{commentcolor}{\# Mask inputs and project to $d_{\text{model}}$}
  \STATE $x_i,\, (f_i, l_\text{len}, l_\text{ang})\ =\ x_i \mkern-4mu\cdot\mkern-2mu \mathbb{I}_\textup{molecule},\, (f_i, l_\text{len}, l_\text{ang}) \mkern-4mu\cdot\mkern-2mu \mathbb{I}_\textup{material}$
  \STATE $h_i \,=\, \text{Embed}(a_i) \,+\, \text{Embed}(c) \,+\, \text{Embed}(t) \,+$ \\
  $\mkern44mu \sum_{v \in \{x_i, f_i, l_\text{len}, l_\text{ang}\}} \text{Lin}^\text{NB}_{v}(v)$ \hfill $h_i \in \mathbb{R}^{d_{\text{model}}}$

  \item[] \textcolor{commentcolor}{\# Apply encoder trunk \ ($L$ layer transformer)}
  \STATE $\{z_i,z_i^\text{K}\} = \text{TransformerEncoder}^L(\{h_i\})$

  \item[] \textcolor{commentcolor}{\# Apply residual cross-attention to input embeddings}
  \STATE $\{h_i^a, h_i^x, h_i^f, h_i^{l_\text{len}}, h_i^{l_\text{ang}}\} = \text{TransformerDecoderBlock}(\{z_i, h_i\})$

  \item[] \textcolor{commentcolor}{\# Denoise inputs}
  \STATE $a'_i = \argmax(\text{Lin}^\text{B}(\text{SiLU}(\text{Lin}^\text{B}(h_i^a))))$ \hfill $a'_i \in \mathbb{Z}$
  \STATE $x'_i = \text{Lin}^\text{NB}(\text{LayerNorm}(h_i^x))$ \hfill $x'_i \in \mathbb{R}^3$
  \STATE $f'_i = \text{Lin}^\text{NB}(\text{LayerNorm}(h_i^f))$ \hfill $f'_i \in \mathbb{R}^3$
  \STATE $l'_\text{len} = \text{Lin}^\text{NB}\left(\text{LayerNorm}\left(\frac{1}{N} \sum_{i=1}^{N} h_i^{l_\text{len}}\right)\right)$ \hfill $l' \in \mathbb{R}^3$
  \STATE $l'_\text{ang} = \text{Lin}^\text{NB}\left(\text{LayerNorm}\left(\frac{1}{N} \sum_{i=1}^{N} h_i^{l_\text{ang}}\right)\right)$ \hfill $l' \in \mathbb{R}^3$

  \item[] \textcolor{commentcolor}{\# Predict outputs \ (cross-attention + $M$ layer transformer)}
  \STATE $\{h_i^\text{props}, h_i^\text{energy}, h_i^\text{forces}\} = \text{TransformerDecoderBlock}_y(\{z_i^\text{K}, h_i\})$,
  \\ \hspace{1ex} for $y \in \{\text{props, energy, forces}\}$
  \STATE Update $\{h_i^y\} \leftarrow \text{TransformerEncoder}_\text{y}^M(\{h_i^y\})$
  \STATE $p^\text{props'} = \text{Lin}^\text{B}\left(\frac{1}{N} \sum_{i=1}^{N} h_i^\text{props}\right)$ \hfill $p \in \mathbb{R}^{19}$
  \STATE $p^\text{energy'} = \text{Lin}^\text{B}\left(\frac{1}{N} \sum_{i=1}^{N} h_i^\text{energy}\right)$ \hfill $p \in \mathbb{R}^1$
  \STATE $p_i^\text{forces'} = \text{Lin}^\text{NB}\left(h_i^\text{forces}\right)$ \hfill $p \in \mathbb{R}^3$
\end{algorithmic}
\end{algorithm}
\end{minipage}%
}%
\hfill
\resizebox{0.48\textwidth}{!}{%
\begin{minipage}[t]{0.57\textwidth}
\begin{algorithm}[H]
\scriptsize
\caption{Pseudocode for multimodal sampling}
\label{algorithm:multimodal_sampling}
\begin{algorithmic}[1]
  \STATE \textbf{Input:} Denoising model $\mathcal{T}$, number of atoms $N$, sampling steps $\{t_i\}_{i=0}^S$
  \STATE \textbf{Output:} Sampled 3D atoms $(\mathbf{a}, \mathbf{x}, \mathbf{f}, \mathbf{l}_\text{len}, \mathbf{l}_\text{ang})$
  \vspace{0.1in}

  \STATE \textbf{def} \textsc{DiscreteFlowStep}($\mathbf{a}_t, \mathbf{p}'_{t, a}, t, \Delta t$):
    \item[] \quad \textcolor{commentcolor}{\# NOTE: All indexed ops are vectorized}
    \STATE \quad $\mathbf{r}_t(i, \cdot) \gets \frac{\Delta t}{1-t} \mathbf{p}'_{t, a}(i)$ \citep{campbell24dfm}
    \STATE \quad $\mathbf{r}_t(i, \mathbf{a}_t(i)) \gets -\sum_{j \neq \mathbf{a}_t(i)} \mathbf{r}_t(i, j)$
    \STATE \quad $\mathbf{p}_{t+\Delta t}(i, j) \gets 1_{\mathbf{a}_t(i)=j} + \mathbf{r}_t(i,j)$
    \STATE \quad $\mathbf{a}_t(i) \gets \text{Categorical}(\mathbf{p}_{t+\Delta t}(i, \cdot))$
    \STATE \quad \textbf{return} $\mathbf{a}_t$
  \vspace{0.1in}

  \STATE \textbf{def} \textsc{EuclideanStep}($\mathbf{z}_t, \mathbf{z}'_t, t, \Delta t, g(\cdot), \gamma_g$):
    \STATE \quad $\mathbf{v}_t \gets \frac{1}{1-t}(\mathbf{z}'_t - \mathbf{z}_t)$ \citep{geffner2025proteina}
    \STATE \quad $\mathbf{s}_t \gets g(t)\frac{t \mathbf{v}_t - \mathbf{z}_t}{1-t}$
    \STATE \quad $d\mathbf{W}_t \gets \sqrt{2\,\gamma_g\,g(t)}\,\mathcal{N}(0, I)$
    \STATE \quad $\mathbf{z}_t \gets (\mathbf{v}_t + \mathbf{s}_t + d\mathbf{W}_t)\Delta t$
    \STATE \quad \textbf{return} $\mathbf{z}_t$
  \vspace{0.1in}

  \item[] \textcolor{commentcolor}{\# Initialize all modalities with random noise}
  \STATE $\mathbf{a}_t \gets \text{Categorical}(\delta(\frac{1}{\# \text{ atom types}}))$
  \STATE Initialize $\mathbf{z}_t \gets \mathcal{N}(0, I)$ for each continuous modality $\mathbf{z} \in \{\mathbf{x}, \mathbf{f}, \mathbf{l}_\text{len}, \mathbf{l}_\text{ang}\}$
  
  \item[] \textcolor{commentcolor}{\# Iteratively denoise over sampling steps}
  \FOR{$i = 1 \text{ to } S$}
    \STATE \quad $\Delta t \gets t_i - t_{i-1}$
    \item[] \quad \textcolor{commentcolor}{\# Predict denoised endpoints using the model $\mathcal{T}$}
    \STATE \quad $(\mathbf{p}'_{t, a}, \mathbf{x}'_t, \mathbf{f}'_t, \mathbf{l}'_{t, \text{len}}, \mathbf{l}'_{t, \text{ang}}) \gets \mathcal{T}(\mathbf{a}_t, \mathbf{x}_t, \mathbf{f}_t, \mathbf{l}_{t, \text{len}}, \mathbf{l}_{t, \text{ang}}, t_i)$
    
    \item[] \quad \textcolor{commentcolor}{\# Perform one sampling step for each modality}
    \STATE \quad $\mathbf{a}_t \gets \textsc{DiscreteFlowStep}(\mathbf{a}_t, \mathbf{p}'_{t, a}, t_i, \Delta t)$
    \STATE \quad Update $\mathbf{z}_t \gets \textsc{EuclideanStep}(\mathbf{z}_t, \mathbf{z}'_t, t_i, \Delta t)$ for each $\mathbf{z}$
  \ENDFOR

  \item[] \textcolor{commentcolor}{\# Return final denoised sample}
  \STATE \textbf{return} $(\mathbf{a}, \mathbf{x}, \mathbf{f}, \mathbf{l}_\text{len}, \mathbf{l}_\text{ang})$
\end{algorithmic}
\end{algorithm}
\end{minipage}
}%
\end{figure}

\subsection{Network architecture}
\label{subsection:network_architecture}

We use a standard Transformer design, titled the \textbf{Trunk-based Flow Transformer (\textsc{TFT})}, to learn a shared latent representation of each molecular and material modality. For a given 3D atomic system $\left(\bm{A}, \bm{X}, \bm{F}, \bm{L}_\text{len}, \bm{L}_\text{ang}\right)$ with a periodic (material) or non-periodic (molecule) class label $c$, denoising timestep index $t$, and (optional) additive noise, a \textsc{TFT} model $\mathcal{T}$ with $L$ Transformer trunk layers learns latent representations $\bm{Z} = \{z_i\}_{i=1}^N \in \mathbb{R}^{d \times N}$ from each molecular and material modality to predict denoised modalities
$\bm{S} = \left(\bm{A}', \bm{X}', \bm{F}', \bm{L}_\text{len}', \bm{L}_\text{ang}'\right)$
and auxiliary properties
$\bm{P} = \left(\bm{P}^\text{props'}, \bm{P}^\text{energy'}, \bm{P}^\text{forces'} \right) = \left(p^\text{props'} \in \mathbb{R}^{19},\ p^\text{energy'} \in \mathbb{R}^{1},\ \{p_i^\text{forces'}\}_{i=1}^N \in \mathbb{R}^{3 \times N} \right)$:
\begin{equation}
    \label{equation:tft_embeddings}
    \bm{S}, \bm{P}\ =\ \mathcal{T}\left(\bm{A}, \bm{X}, \bm{F}, \bm{L}_\text{len}, \bm{L}_\text{ang}, c, t\right),
\end{equation}
where a 10\% drop probability is applied to $c$ during training to support (optional) classifier-free guidance during inference, and additional (e.g., property) conditioning can be added likewise.

\cref{algorithm:tft_model_pseudocode} provides pseudocode for the forward pass of a \textsc{TFT} model with $L$ trunk layers, where layer $K \le L$ is used for auxiliary task representations.
Using random data augmentations during training, the model learns the rotational and periodic symmetries of molecules and materials, respectively \citep{abramson2024accurate, joshi2025allatom}.
Additionally, inspired by the Platonic Transformer \citep{islam2025platonic}, we experiment with an O(3)-equivariant version of \textsc{Zatom-1} (\textsc{Platom-1}) featuring a \textsc{TFT} model variant titled the \textbf{Trunk-based Flow Platoformer (\textsc{TFP})},
which we describe in more detail in Appendix~\ref{appendix:tfp_main}.

\subsection{Stage 1: Multimodal flow pretraining}
\label{subsection:multimodal_flow_pretraining}

Conceptually, given a pair of data samples ($\mathbf{x}_0$, $\mathbf{x}_1$) from their respective source and target distributions, flow matching (FM) offers a scalable technique for training generative models. Namely, using FM, one can train a neural network $v_\theta(\mathbf{x}_t, t)$ to sample data from the target distribution by matching the expected velocities transporting samples from the (often simpler) source to the target distribution:
\begin{equation}
    \mathcal{L}_\text{FM} = \mathbb{E}_{t, (\mathbf{x}_0, \mathbf{x}_1)} [\lVert v_\theta(\mathbf{x}_t, t) - u_t \rVert_2^2],
\end{equation}
where $\mathbf{x}_t = (1 - t)\mathbf{x}_0 + t\mathbf{x}_1$ and $u_t = \mathbf{x}_1 - \mathbf{x}_0$. In this work, to optimize continuous modalities, we use Euclidean diffusion/conditional flow matching (CFM) \citep{sohl2015deep,song2019generative,ho2020denoising,lipman2023flow,albergo2023building,tong2024improving}, leveraging the equivalence between these two methods \citep{diffusionflow2024}. For optimizing atom types, we employ Discrete CFM, which is parameterized within the framework of Discrete Flow Models \citep{campbell24dfm}.

Now, specifically consider a molecule or material comprised of $N$ atoms, characterized by zero-centered ground-truth 3D coordinates $\bm{X}$, fractional coordinates $\bm{F}$, lattice lengths $\bm{L}_\text{len}$, angles $\bm{L}_\text{ang}$, and atom types $\bm{A}$. The continuous 3D coordinates and fractional coordinates, along with lattice lengths and angles, undergo partial noise perturbation represented as $\bm{X}_t = t \cdot \bm{X} + (1-t) \cdot \epsilon$, where the noise $\epsilon$ is drawn from a Gaussian distribution $\mathcal{N}(0, I)$. For the discrete modality, the noisy atom types $\bm{A}_t$ are derived by interpolating between atom type probabilities and sampling from a categorical distribution, expressed as $\bm{A}_t \sim \text{Cat}(t \cdot \delta(\bm{A}) + (1-t) \cdot \delta(\frac{1}{\#\ \text{atom types}}))$, where $\delta(\cdot)$ generates a one-hot encoding \citep{campbell24dfm}. During the training phase, the model receives $\left(\bm{X}_t, \bm{F}_t, \bm{L}_{t_\text{len}}, \bm{L}_{t_\text{ang}}, \bm{A}_t\right)$ as input and learns to predict the terminal points of the trajectory.
\paragraph{Flow matching losses.} The objective functions for each continuous and discrete modality, respectively, are defined as follows:
\begin{equation}
    \mathcal{L}_{\text{metric}}(\bm{X}) = \mathbb{E}_{\epsilon,t} \left[ \frac{1}{N} \| \bm{X}' - \bm{X} \|_2^2 \right] \qquad \mathcal{L}_{\text{discrete}}(\bm{A}) = \mathbb{E}_{t} \left[ -\sum_{i} a_i \log(a_i') \right]
\end{equation}
These objectives are consolidated into a multi-objective framework, employing a weighting factor $\lambda_{\text{discrete}} \in [0, 1]$, expressed as
\begin{equation}
    \begin{split}
        \mathcal{L}_{\text{total}} &= \mathcal{L}_{\text{metric}}(\bm{X}) + \mathcal{L}_{\text{metric}}(\bm{F}) + \mathcal{L}_{\text{metric}}(\bm{L}_\text{len}) + \mathcal{L}_{\text{metric}}(\bm{L}_\text{ang}) + \lambda_{\text{discrete}} \cdot \mathcal{L}_{\text{discrete}}(\bm{A}).
    \end{split}
\end{equation}
During training, we set $\lambda_{\text{discrete}} = 0.1$ and sample from $t \sim \text{Beta}(\alpha_t, 1)$, where $\alpha_t=1.8$, in line with prior studies \citep{geffner2025proteina, vonessen2026tabasco}. Further, to learn data symmetries and accelerate convergence during training, we randomly rotate and (for periodic materials) translate multiple copies of each input example, which are noised to different levels according to batch-wise $t$ \citep{abramson2024accurate}. As $t \to 1$, the model resembles an identity function, since we adopt an endpoint formulation for flow matching due to its empirical performance for scientific applications \citep{stark2024harmonic}. To enable the model to effectively learn accurate atom placements even when $t \to 1$ causes the losses to approach zero, based on the sampled time $t$, we scale the loss with $\beta(t) \cdot \mathcal{L}_{\text{total}}(\bm{X}_t, \bm{F}_t, \bm{L}_{t_\text{len}}, \bm{L}_{t_\text{ang}}, \bm{A}_t)$, where $\beta(t) = \min \left\{ 100,\ \frac{1}{(1-t)^2} \right\}$.

\paragraph{Multimodal sampling.} \cref{algorithm:multimodal_sampling} describes how we perform unconditional sampling of each continuous and discrete modality to generate 3D molecules and materials. Concisely, we sample random atom types or Gaussian noise for discrete and continuous modalities, respectively comprising $N$ atoms (where $N$ is sampled according to training dataset statistics \citep{hoogeboom2022equivariant}), and iteratively denoise each modality using a series of Euler steps operating on the multimodal predictions of the TFT model $\mathcal{T}$. Notably, other ODE solvers can be used, though we find that multimodal Euler steps provide strong empirical performance without considerable hyperparameter tuning.

\subsection{Stage 2: Finetuning}
\label{subsection:finetuning}

As illustrated in \cref{figure:zatom_1_overview}, we treat conditional flow matching of the modalities $\left(\bm{A}, \bm{X}, \bm{F}, \bm{L}_\text{len}, \bm{L}_\text{ang}\right)$ as a self-supervised pretraining task akin to bidirectional modeling used to pretrain non-autoregressive BERT-style models \citep{devlin2019bert} for enhanced representation learning \citep{zhang2025diffusion} and \textit{label-free} transfer learning. Once the model weights listed up to Line 10 of \cref{algorithm:tft_model_pseudocode} have been (pre)trained, for finetuning, (based on ablations described in Table \ref{appendix:table:qm9_dnmg_finetuning_results} of Appendix \ref{appendix:additional_results}) we freeze these weights and enable training of the $M$-layer auxiliary Transformer trunk weights following Line 10 using trunk layer $K$'s frozen embeddings. By default, $K=L$, a choice that we ablate in Table \ref{table:qm9_matbench_property_prediction_results} of \cref{section:results}.

\paragraph{Predictive Losses.} For property prediction, we supervise the task's embeddings \( p_b^\text{props'} \) using the mean absolute error (MAE) over each batch element \( b \) as follows:
\begin{equation}
    \mathcal{L}_{\text{MAE}}(\bm{P}^\text{props'}) = \frac{1}{B} \sum\nolimits_b \big|p_b^\text{props'} - p_b^\text{props}\big|,
\end{equation}
where \( B \) is the number of samples in a given mini-batch, \( p_b^\text{props'} \) denotes the predicted property (or properties) for the \( b \)-th sample, and \( p_b^\text{props} \) denotes the true property for the same sample. Similarly, energy embeddings $p_b^\text{energy'}$ and force embeddings $p_b^\text{forces'}$ are supervised using the batch-wise mean squared error (MSE), with the loss weight $\lambda_\text{forces}$ specifically applied for forces:
\begin{equation}
    \mathcal{L}_{\text{MSE}}(\bm{P}^\text{energy'}) = \frac{1}{B} \sum\nolimits_b \big(p_b^\text{energy'} - p_b^\text{energy}\big)^2 \quad \mathcal{L}_{\text{MSE}}(\bm{P}^\text{forces'}) = \mathbb{E} \left[ \frac{1}{N} \| \bm{P}^\text{forces'} - \bm{P}^\text{forces} \|_2^2 \right]
\end{equation}

\section{Experimental Setup}
\label{section:experimental_setup}

In this section, we describe how we train and evaluate \textsc{Zatom-1}'s performance for generative modeling and representation learning of 3D molecules and materials. Additionally, we discuss the representative baselines against which we comprehensively compare \textsc{Zatom-1}.

\textbf{Datasets.} To fairly compare \textsc{Zatom-1} to previous methods \citep{joshi2025allatom}, we train models on periodic materials from MP20 and non-periodic molecules from QM9 for our baseline generative pretraining experiments. MP20 \citep{xie2022crystal} is a collection of 45,231 metastable crystal structures from the Materials Project \citep{jain2013commentary}, with up to 20 atoms in each material's unit cell spanning 89 different atom/element types. QM9 \citep{wu2018moleculenet} is comprised of 130,000 stable small molecules with up to 9 heavy atoms (C, N, O, F) along with hydrogens. To ensure fair comparisons, we split the data according to prior work \citep{xie2022crystal, hoogeboom2022equivariant}. Additionally, we report generative pretraining results for GEOM-Drugs \citep{axelrod2022geom}, a collection of 430,000 large molecules with up to 180 atoms, following past benchmarking conventions \citep{vignac2023midi, joshi2025allatom}. In Table \ref{appendix:table:mof_generation_results} of Appendix \ref{appendix:additional_results}, we also employ the QMOF dataset of 14,000 metal-organic framework (MOF) structures \citep{rosen2021machine} after removing structures with more than 150 atoms \citep{joshi2025allatom}.

For predictive finetuning, we use the QM9 and Matbench datasets, the latter supporting several fundamental property prediction tasks in materials science \citep{dunn2020benchmarking}. Further, in Table \ref{appendix:table:mptrj_omol25_finetuned_mlip_results} of Appendix \ref{appendix:additional_results}, we adopt the OMol25 (4M) \citep{levine2025open} and MPtrj \citep{deng2023chgnet} datasets for molecule and material energy and force prediction, respectively. Data splits follow prior work for fair evaluation \citep{liao2023equiformer, dunn2020benchmarking, levine2025open, deng2023chgnet}. Notably, we are also the \textit{first} to adopt OMol25 (4M) for \textit{larger-scale} generative pretraining experiments.

\textbf{Training and hyperparameters.} Our generative pretraining procedure is described in \cref{appendix:table:mp20_qm9_pretraining_hyperparams,appendix:table:geom_qmof_omol25_pretraining_hyperparams} of Appendix \ref{appendix:additional_model_details}. Predictive finetuning is outlined in \cref{appendix:table:finetuning_hyperparams} of Appendix \ref{appendix:additional_model_details}, during which we finetune the weights dedicated to a particular task individually to minimize time to convergence.

Key inference hyperparameters include the white noise scale $\gamma_g$ of each modality (following \citet{geffner2025proteina}) and the number of integration steps $T$. Through an extensive hyperparameter sweep illustrated in \cref{appendix:figure:crystals_molecules_mp20_qm9_inference_sweep} of Appendix \ref{appendix:additional_model_details}, we find that $T=50$ or $100$ with $\gamma_g = 0.01$ generally works well for molecule and material generation, except for QM9-sized small molecules, which benefit from adding additional white noise (e.g., 50) to non-periodic atom positions to improve sample uniqueness/diversity. Like \citet{vonessen2026tabasco}, $g(t) = \frac{1}{t+0.01}$ for noise scaling by default.

\textbf{Evaluation metrics.} Following previous work \citep{hoogeboom2022equivariant, joshi2025allatom}, for generative pretraining, we evaluate \textsc{Zatom-1}'s ability to generate valid and realistic molecules and materials by sampling 10,000 molecules and 2,500 materials and computing validity, uniqueness, and PoseBusters sanity checks for molecules \citep{buttenschoen2024posebusters} and calculating validity, energy, (meta)stability, uniqueness, and novelty (i.e., (M)SUN) metrics using LeMat-GenBench \citep{duval2025lemat}. These generative metrics are described in further detail in Appendix \ref{appendix:evaluation_metrics}. For predictive finetuning, we assess \textsc{Zatom-1}'s MAE for property, energy, and force prediction tasks, scaled/unit-converted according to literature benchmarking conventions.


\begin{table}[t]
  \centering
  \caption{\textbf{Crystal generation results on LeMat-GenBench with MP20 training.} Core model evaluation metrics are calculated for 2,500 sampled crystals using an \textit{MLIP ensemble} with \textit{LeMat-Bulk} as the reference set. Best values are shown in \textbf{bold}, second-best values are \underline{underlined}. $\bm{\star}$ denotes aggregate metrics that, summed together, constitute the primary objective to optimize for this benchmark (i.e., SUN + MSUN). Notably, jointly trained \textsc{Zatom-1} (80M) achieves its results using $\sim$400 GPU training hours, 3x faster than jointly trained ADiT (180M). Further, jointly trained \textsc{Zatom-1-WD} achieves state-of-the-art SUN (and MSUN) rates for materials discovery in both relaxed (and unrelaxed) method categories \citep{deng2023chgnet} through aggressive weight decay ($1e^{-2}$) and early stopping ($1/5$ into training, before molecule validation performance has fully converged).}
  \label{table:mp20_dng}
  \resizebox{\textwidth}{!}{%
  \begin{tabular}{@{}lrrrrrrrrrrr@{}}
    \toprule
    \textbf{Model} & \textbf{\# Params} & \textbf{Valid\% $\bm{\uparrow}$} & \textbf{Unique\% $\bm{\uparrow}$} & \textbf{Novel\% $\bm{\uparrow}$} & \multicolumn{3}{c}{\textbf{Energy-based}} & \multicolumn{2}{c}{\textbf{Stability}} & \multicolumn{2}{c}{\textbf{Metastability}} \\
    \cmidrule(lr){6-8} \cmidrule(lr){9-10} \cmidrule(lr){11-12}
    & & & & & $\bm{E_f}$ \textbf{(Std)} $\bm{\downarrow}$ & $\bm{E}_\text{\textbf{hull}}$ \textbf{(Std)} $\bm{\downarrow}$ & \textbf{RMSD (Std)} $\bm{\downarrow}$ & \textbf{Stable\%} $\bm{\uparrow}$ & $\bm{\star}$ \textbf{SUN\%} $\bm{\uparrow}$ & \textbf{Metastable\%} $\bm{\uparrow}$ & $\bm{\star}$ \textbf{MSUN\%} $\bm{\uparrow}$ \\
    \midrule
    \textbf{With relaxation} \\
    Crystalite  & 67M & \textbf{97.2}             & \textbf{95.8} & 53.2 & $-0.89$ & $0.09$ & $0.13$ & \textbf{12.7} & \underline{1.5 (38)} & 51.6 & \underline{22.6} \\
    MatterGen     & 47M & 95.7 & \underline{95.1} & \underline{70.5} & $-0.70 \pm 0.79$ & 0.18 $\pm$ 0.18 & $0.39 \pm 0.50$ & 2.0 & 0.2 (6) & 33.4 & 15.0 \\
    PLaID++      & 7B & \underline{96.0}    & 77.8          & 24.2          & $-0.50 \pm 0.44$ & 0.09 $\pm$ 0.16 & \textbf{0.13} $\pm$ \textbf{0.29} & \underline{12.4} & 1.0 (26) & 60.7 & 7.6 \\
    WyFormer      & 8B & 93.4             & 93.0          & 66.4 & $-0.43 \pm 0.95$ & $0.50 \pm 0.51$ & $0.81 \pm 0.98$ & 0.5 & 0.1 (3) & 15.7 & 1.9 \\
    WyFormer-DFT  & 8B & 95.2             & 95.0 & 66.4 & $-0.67 \pm 0.91$ & $0.27 \pm 0.36$ & $0.42 \pm 0.60$ & 3.7 & 0.4 (9) & 24.8 & 7.8 \\
    \midrule
    \textbf{Without relaxation} \\
    Jointly trained ADiT          & 180M & 90.6             & 87.8          & 26.0          & $-0.73 \pm 0.93$ & $0.33 \pm 0.45$ & $0.38 \pm 0.40$ & 0.4 & 0.0 (0) & 36.5 & 1.0 \\
    Crystal-GFN & 1M   & 51.7             & 51.7          & 51.7          & \textbf{--1.30} $\pm$ \textbf{2.63} & $2.09 \pm 2.38$ & $1.87 \pm 0.97$ & 0.0 & 0.0 (0) & 0.0 & 0.0 \\
    Crystalformer & 5M & 69.9             & 69.4          & 31.8          & $-0.17 \pm 1.48$ & $0.70 \pm 1.33$ & $0.66 \pm 1.05$ & 1.4 & 0.0 (0) & 28.8 & 3.1 \\
    DiffCSP & 12M       & 95.7 & 94.8          & 66.2          & $-0.64 \pm 0.96$ & $0.28 \pm 0.56$ & $0.59 \pm 0.63$ & 2.3 & 0.1 (3) & 29.8 & 8.5 \\
    DiffCSP++ & 12M     & 95.3             & 95.1 & 62.0          & $-0.52 \pm 0.87$ & $0.41 \pm 0.44$ & $0.69 \pm 0.82$ & 1.0 & 0.2 (4) & 26.4 & 5.0 \\
    LLaMat2-CIF & 7B   & 84.4             & 81.4          & 30.0          & $-0.47 \pm 1.01$ & $0.44 \pm 0.71$ & $0.54 \pm 0.71$ & 0.7 & 0.1 (3) & 34.7 & 2.1 \\
    LLaMat3-CIF & 8B   & 15.4             & 15.2          & 10.5          & $0.74 \pm 2.69$  & $1.71 \pm 2.48$ & $1.01 \pm 1.10$ & 0.1 & 0.0 (0) & 2.1  & 0.2 \\
    SymmCD & 12M        & 73.4             & 73.0          & 47.0          & $-0.02 \pm 1.22$ & $0.88 \pm 1.07$ & $0.87 \pm 1.09$ & 1.4 & 0.1 (2) & 18.6 & 2.4 \\
    \midrule
    \textbf{Ours: Without relaxation} \\
    MP20-only \textsc{Zatom-1} & 80M   & 73.2                & 70.4          & 21.0          & $-0.53 \pm 0.97$ & $0.44 \pm 0.64$ & $0.42 \pm 0.49$ & 0.6 & 0.00 (0) & 32.7 & 0.20 (5)   \\
    Jointly trained \textsc{Zatom-1} & 80M             & 88.5             & 84.4          & 8.1          & $-0.84 \pm 0.89$ & $0.19 \pm 0.37$ & $0.21 \pm 0.27$ & 2.7 & 0.04 (1) & 51.8 & 0.36 (9) \\
    \rowcolor{orange!10} Jointly trained \textsc{Zatom-1-L} & 160M           & 95.0                & 90.4          & 3.7          & $-0.87 \pm 0.86$ & \textbf{0.09} $\pm$ \textbf{0.21} & \underline{0.14 $\pm$ 0.20} & 3.1 & 0.04 (1) & \textbf{71.5} & 0.64 (16) \\
    Jointly trained \textsc{Zatom-1-XL} & 300M          & 94.9                & 90.1          & 4.8          & \underline{$-0.91 \pm 0.87$} & $0.11 \pm 0.23$ & $0.16 \pm 0.19$ & 3.3 & 0.00 (0) & \underline{64.3} & 0.36 (9) \\
    \midrule
    \rowcolor{orange!30} Jointly trained \textsc{Zatom-1-WD} & 80M             & 67.2             & 65.9          & 67.1          & $-0.37 \pm 1.01$ & $0.52 \pm 0.67$ & $0.47 \pm 0.50$ & 0.2 & 0.2 (5) & 22.8 & 21.7 (542) \\
    \midrule
    \textbf{Ours: With relaxation} \\
    \rowcolor{orange!40} Jointly trained \textsc{Zatom-1-WD} & 80M             & 88.3             & 86.4          & \textbf{87.9}          & $-0.82 \pm 0.84$ & \underline{$0.11 \pm 0.14$} & $0.32 \pm 0.40$ & 4.6 & \textbf{4.4} (\textbf{110}) & 50.6 & \textbf{48.9} (\textbf{1,222}) \\
    \bottomrule
  \end{tabular}%
  }
\end{table}


\begin{table}[t]
\centering
\caption{\textbf{Molecule generation results on QM9.} We report (a) validity and uniqueness rates as well as (b) \% pass rates on 7 sanity checks from PoseBusters for 10,000 sampled molecules. All models explicitly generate hydrogen atoms. Notably, jointly trained \textsc{Zatom-1} (80M) achieves its results using $\sim$400 GPU training hours compared to jointly trained ADiT (180M), which requires $\sim$1,200 GPU pretraining + finetuning hours.}
\label{table:qm9_dng}
\vspace{0.5em}

\begin{minipage}[t]{0.45\columnwidth}
    \centering
    (a) \footnotesize \textbf{Validity results}
    \vspace{0.3em}
    \resizebox{\linewidth}{!}{%
    \begin{tabular}{lccc}
    \toprule
    \textbf{Model} & \textbf{\# Params} & \textbf{Validity (\%) $\uparrow$} & \textbf{Unique (\%) $\uparrow$} \\
    \midrule
    \multicolumn{3}{l}{\textbf{One-stage training}} \\
    Equivariant Diffusion & 20M & 91.90 & \underline{98.69} \\
    Symphony & 2M & 83.50 & 97.98 \\
    \midrule
    \multicolumn{3}{l}{\textbf{Two-stage training}} \\
    GeoLDM & 20M & 93.80 & \textbf{98.82} \\
    QM9-only ADiT & 180M & 92.19 & 97.90 \\
    Jointly trained ADiT & 180M & 94.45 & 97.82 \\
    \midrule
    \multicolumn{3}{l}{\textbf{Ours: One-stage training}} \\
    QM9-only \textsc{Zatom-1} & 80M & 92.88 & 97.71 \\
    \midrule
    \rowcolor{orange!10} Jointly trained \textsc{Zatom-1} & 80M & 94.94 & 97.16 \\
    Jointly trained \textsc{Zatom-1-L} & 160M & \textbf{95.26} & 96.84 \\
    Jointly trained \textsc{Zatom-1-XL} & 300M & \underline{95.19} & 97.10 \\
    \midrule
    \rowcolor{orange!30} Jointly trained \textsc{Zatom-1-WD} & 80M & 94.62 & 97.44 \\
    \bottomrule
    \end{tabular}%
    }
\end{minipage}
\begin{minipage}[t]{0.54\columnwidth}
    \centering
    (b) \footnotesize \textbf{PoseBusters results}
    \vspace{0.3em}
    \resizebox{\linewidth}{!}{%
    \begin{tabular}{lccccc}
    \toprule
    \textbf{Test (\% pass) $\uparrow$} & \textbf{Symphony} & \textbf{Eq. Diff.} & \textbf{ADiT} & \textbf{\textsc{Zatom-1}} &   \textbf{\textsc{Zatom-1-WD}} \\
    \midrule
    Atoms connected & 99.92 & 99.88 & 99.70 & \underline{99.98} & \textbf{100.00} \\
    Bond angles & 99.56 & \textbf{99.98} & 99.85 & \underline{99.95} & 99.91 \\
    Bond lengths & 98.72 & \textbf{100.00} & 99.41 & \underline{99.97} & 99.94 \\
    Aromatic ring flat & \textbf{100.00} & \textbf{100.00} & \textbf{100.00} & \textbf{100.00} & \textbf{100.00} \\
    Double bond flat & 99.07 & 98.58 & 99.98 & \underline{99.99} & \textbf{100.00} \\
    Internal energy & 95.65 & 94.88 & 95.86 & \underline{99.78} & \textbf{99.79} \\
    No steric clash & 98.16 & 99.79 & 99.79 & \underline{99.81} & \textbf{99.84} \\
    \bottomrule
    \end{tabular}%
    }
\end{minipage}
\end{table}

\textbf{Baselines.} For generative pretraining, we compare \textsc{Zatom-1} trained jointly on MP20 and QM9 to material-only and molecule-only variants, as well as state-of-the-art baselines for both data sources trained in one or two-stage configurations. Generated samples, as well as qualitative and quantitative evidence of \textsc{Zatom-1}'s cross-domain transfer learning capabilities, are shown in Appendix \ref{appendix:visualization}.

Our experiments for materials generation using LeMat-GenBench include comparisons to: (1) five equivariant/invariant generative models adopting handcrafted multimodal product manifolds: MatterGen \citep{zeni2025generative}, Crystal-GFN \citep{hernandez2023crystal}, DiffCSP \citep{jiao2023crystal}, DiffCSP++ \citep{jiao2024space}, and SymmCD \citep{levy2025symmcd}; (2) two non-equivariant Transformer-based generative models: ADiT \citep{joshi2025allatom} and Crystalite \citep{veljkovic2026crystalite}; and (3) four language model-based generative methods for materials: PLaID++ \citep{xu2025plaid++}, WyFormer \citep{kazeev2025wyckoff}, Crystalformer \citep{cao2025space}, and LLaMat \citep{mishra2024foundational}.

Our molecule generation experiments compare \textsc{Zatom-1} to: (1) four generative methods operating on an equivariant multimodal product manifold: Equivariant Diffusion \citep{hoogeboom2022equivariant}, Symphony \citep{daigavane2024symphony}, EQGAT-diff \citep{le2024navigating}, and SemlaFlow \citep{irwin2025semlaflow}; (2) two latent diffusion/flow matching generative models: GeoLDM \citep{xu2023geometric} and ADiT \citep{joshi2025allatom}; and (3) non-equivariant TABASCO \citep{vonessen2026tabasco} which learns symmetries from data.

For predictive finetuning, we compare \textsc{Zatom-1} to: (1) 11 highly-optimized (i.e., hyperparameter-tuned) property prediction methods: DimeNet++ \citep{gasteiger2020fast}, EGNN \citep{satorras2021n}, PaiNN \citep{schutt2021equivariant}, TorchMD-NET \citep{tholke2022equivariant}, SphereNet \citep{liu2022spherical}, SEGNN \citep{brandstetter2022geometric}, EQGAT \citep{le2022representation}, Equiformer \citep{liao2023equiformer}, EquiformerV2 \citep{liao2024equiformerv}, P$\Theta$NITA \citep{bekkers2024fast}, and the Platonic Transformer \citep{islam2025platonic}; (2) an unoptimized (i.e., non-hyperparameter-tuned) single-task property prediction method: AIM-STL \citep{minot2025aim}; and (3) two unoptimized \textit{multi}-task property prediction methods: AIM-MTL-Scalar and AIM-MTL-Matrix \citep{minot2025aim}. Lastly, our energy and force prediction experiments for molecules (materials) compare \textsc{Zatom-1} to equivariant eSEN \citep{fu2025learning} (Orb-v1 \citep{neumann2024orb}).


\begin{figure}[t]
    \centering
    
    \begin{subfigure}[b]{0.32\textwidth}
        \centering
        \includegraphics[width=\linewidth]{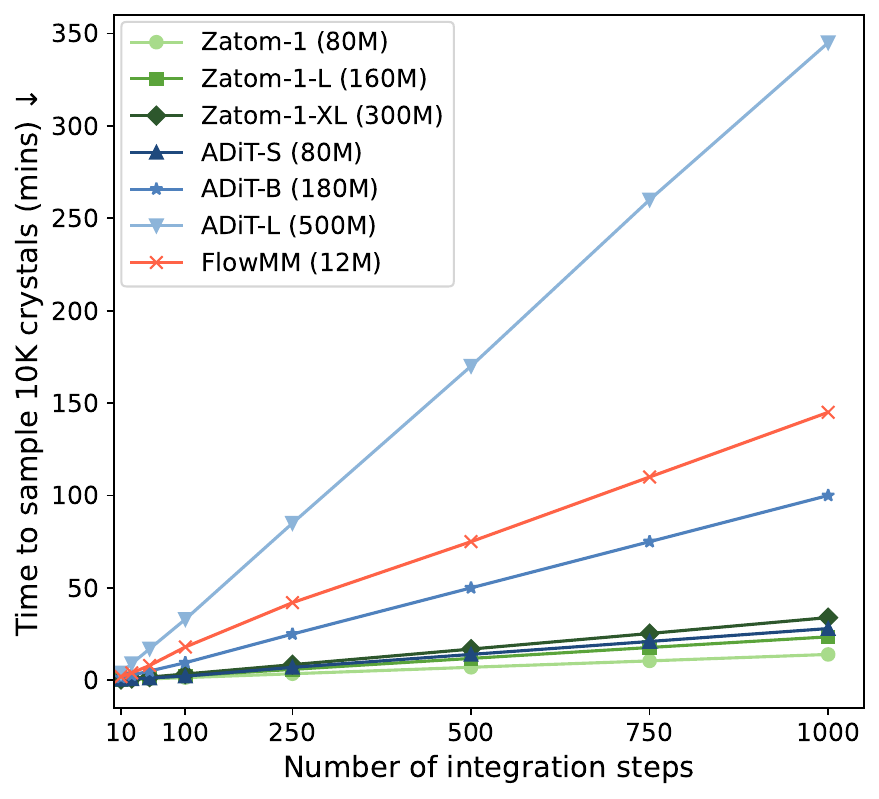}
        \caption{Crystals – MP20}
        \label{figure:mp20_model_speed_results}
    \end{subfigure}
    \hfill
    \begin{subfigure}[b]{0.32\textwidth}
        \centering
        \includegraphics[width=\linewidth]{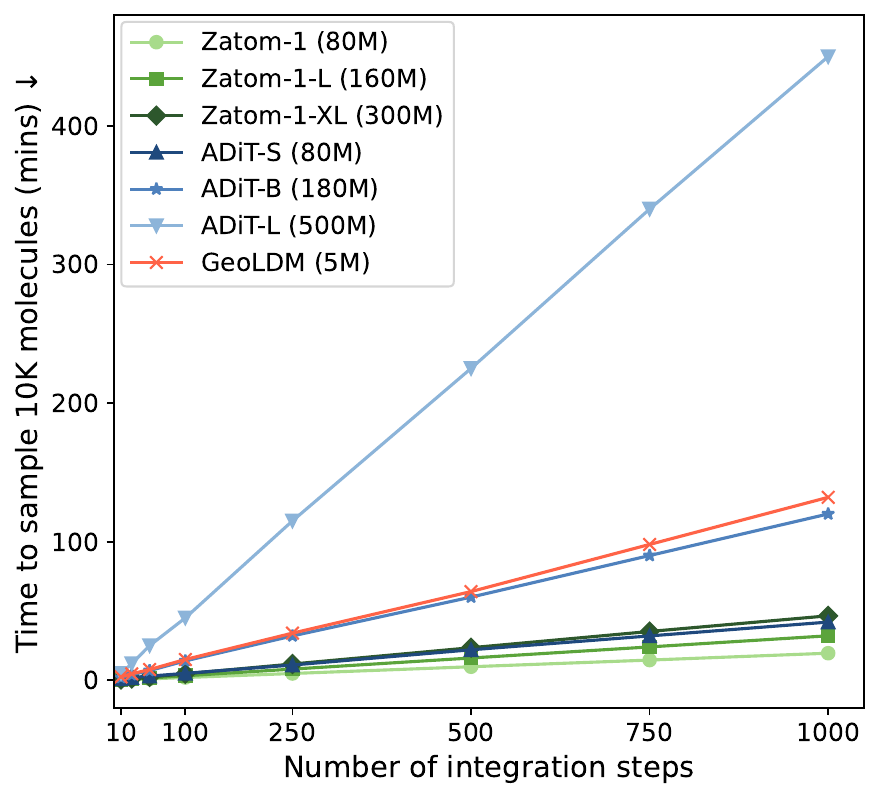}
        \caption{Molecules – QM9}
        \label{figure:qm9_model_speed_results}
    \end{subfigure}
    \hfill
    \begin{subfigure}[b]{0.32\textwidth}
        \centering
        \includegraphics[width=\linewidth]{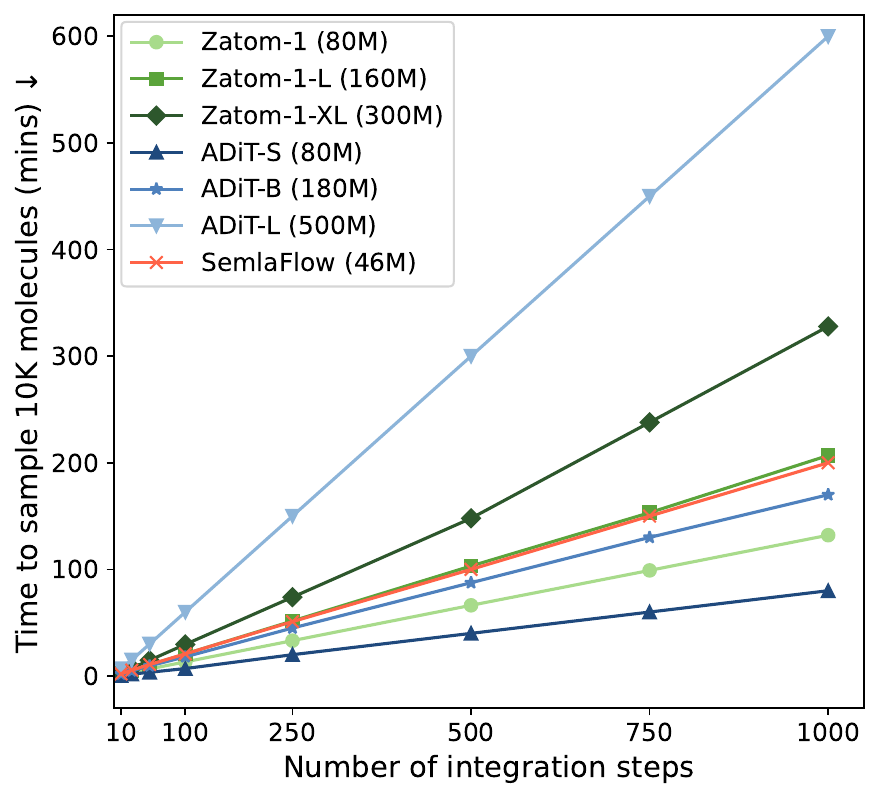}
        \caption{Molecules – GEOM-Drugs}
        \label{figure:geom_drugs_model_speed_results}
    \end{subfigure}
    
    \caption{
        \textbf{\textsc{Zatom-1} is significantly faster than equivariant and latent diffusion models.} 
        We plot the number of integration steps for \textsc{Zatom-1}, equivariant diffusion FlowMM/SemlaFlow, and latent diffusion ADiT/GeoLDM vs. time to generate 10,000 samples on a single NVIDIA GPU. \textsc{Zatom-1} scales better with the number of integration steps compared to equivariant and latent diffusion models.
    }
    \label{figure:model_speed_results}

    \vspace{4ex} 

    \begin{subfigure}[b]{0.32\textwidth}
        \centering
        \includegraphics[width=\linewidth]{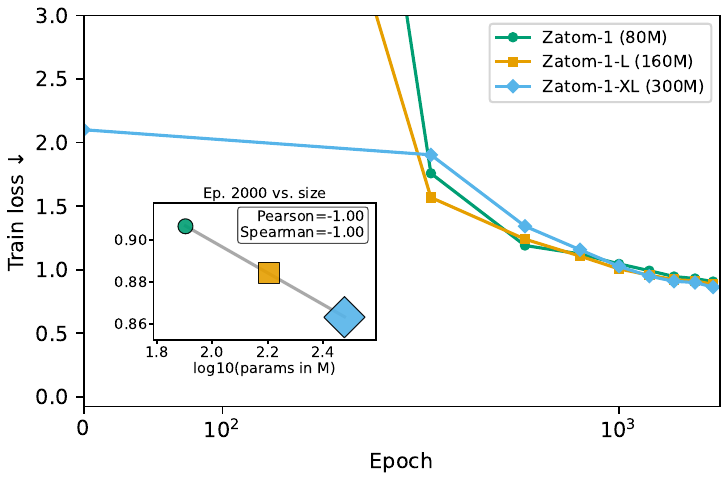}
        \label{figure:model_scaling_train_loss}
    \end{subfigure}
    \hfill
    \begin{subfigure}[b]{0.32\textwidth}
        \centering
        \includegraphics[width=\linewidth]{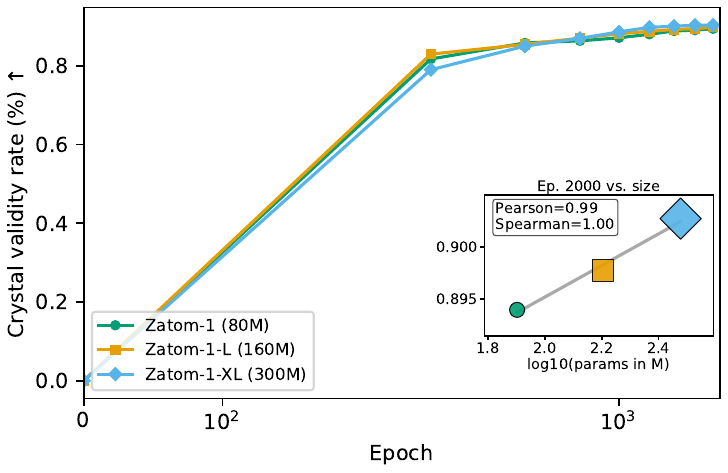}
        \label{figure:model_scaling_crystal_validity}
    \end{subfigure}
    \hfill
    \begin{subfigure}[b]{0.32\textwidth}
        \centering
        \includegraphics[width=\linewidth]{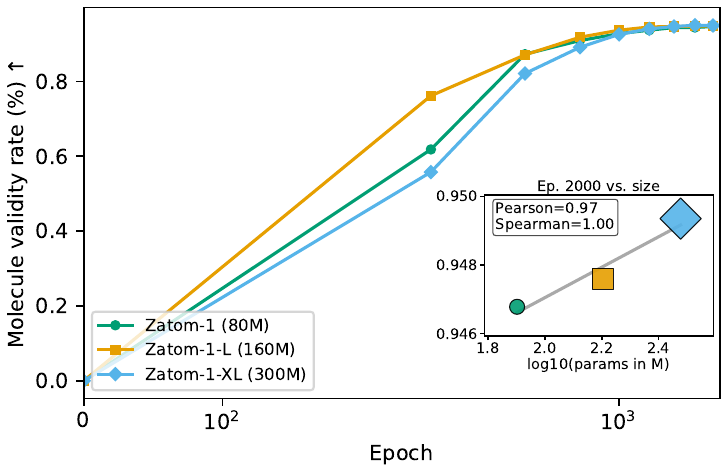}
        \label{figure:model_scaling_molecule_validity}
    \end{subfigure}
    
    \caption{
        \textbf{Scaling up \textsc{Zatom-1} predictably improves performance.} We show the effect of increasing the number of \textsc{Zatom-1} model parameters on training loss and generation validity rates over 2,000 epochs. Correlations for training loss and validity rates at epoch 2,000 vs. \textsc{Zatom-1} parameters (in Millions) demonstrate \textsc{Zatom-1}'s strong scaling potential (Pearson's $r>0.96$).
    }
    \label{figure:model_scaling_results}
\end{figure}

\section{Results}
\label{section:results}

In this section, we discuss \textsc{Zatom-1}'s performance in generative modeling and representation learning tasks for 3D molecules and materials, demonstrating its potential as a scalable proof-of-principle architecture for future atomistic foundation models. Additionally, we discuss potential future directions arising from this work.

\textbf{State-of-the-art crystal and molecule generation.} \cref{table:mp20_dng,table:qm9_dng} show that \textsc{Zatom-1} quantitatively benefits from positive transfer learning between chemical domains compared to MP20-only or QM9-only training (see Appendix \ref{appendix:characterizing_chemical_transfer_learning_in_shared_latent_space} for additional, data-controlled evidence). Specifically, \cref{table:mp20_dng}'s results demonstrate for respective (relaxed/unrelaxed) categories that \textsc{Zatom-1}, namely with \textsc{Zatom-1-WD}'s aggressive regularization and early stopping for the relatively small MP20 dataset, generates the highest percentage of (meta)stable, unique, and novel (i.e., (M)SUN) materials that are chemically valid, advancing the state-of-the-art (relaxed) SUN rate by $\sim$300\% with structural relaxation and the state-of-the-art (unrelaxed) MSUN rate by $\sim$250\% without relaxation. Similarly, \cref{table:qm9_dng}'s QM9 results highlight that $\sim$99\% of \textsc{Zatom-1}'s generated molecules pass all PoseBusters sanity checks, while $\sim$95\% of ADiT's molecules satisfy each check, establishing a new state-of-the-art. In line with \cref{table:qm9_dng}'s results, \cref{table:geom_dng} shows that \textsc{Zatom-1} is the first generative model to reach a maximum-possible PoseBusters validity rate of 94\% for larger, GEOM-Drugs-like molecules \citep{vonessen2026tabasco}, demonstrating the strength of \textsc{Zatom-1}'s molecule \& material pretraining for generating large, \textit{drug-like} molecules. See Appendix \ref{appendix:tfp_results} for \textsc{Platom-1}'s results and Table \ref{appendix:table:geom_dnmg_comprehensive_results} of Appendix \ref{appendix:additional_results} for additional GEOM-Drugs results.

\textbf{\textsc{Zatom-1} scales better than baselines.} For both QM9 and MP20 sample generation, \cref{figure:model_speed_results} illustrates that \textsc{Zatom-1} offers an order of magnitude faster inference speeds than ADiT, a state-of-the-art baseline, and also shows that \textsc{Zatom-1}'s speeds scale better for GEOM-Drugs sampling. Likewise, \cref{figure:model_scaling_results} highlights that \textsc{Zatom-1}'s training and validation metrics improve predictably as the model's size increases. This suggests that parallel data and model scaling (e.g., with OMol25 100M) may improve performance further, which, due to resource constraints, we defer to future work.

\textbf{Chemical transfer learning improves predictions.} \cref{table:qm9_matbench_property_prediction_results} demonstrates that \textsc{Zatom-1} achieves state-of-the-art \textit{multi-task} property prediction performance for QM9. For the majority of QM9's target properties, these results further mark the \textit{first} instance in SciML of positive transfer learning for prediction tasks and zero-shot generalization via joint generative pretraining, indicating that the later trunk layers of \textsc{Zatom-1} ($K=L$ and $K=3L/4$) excel in 3D molecular tasks, while its earlier trunk layers ($K=L/4$ and $K=L/2$) are better suited for (zero-shot) 3D material tasks. Importantly, while non-pretrained \textsc{Zatom-1} often outperforms the comparable AIM-MTL baselines, with joint generative pretraining ($K=L$), \textsc{Zatom-1} \textit{always} strongly surpasses or matches these baselines.

\textbf{Additional results.} Additional results for MOF generation, \textit{larger-scale} pretraining on OMol25 (4M), and property/energy and force prediction are provided in Appendix \ref{appendix:additional_results} on the path to developing unified generative/predictive machine learning-based interatomic potentials (MLIPs) \citep{wood2025uma}. Notably, \textsc{Zatom-1} achieves competitive performance in materials force prediction against Orb-v1 and successfully balances molecular and material property prediction tasks through additional (supervised) cross-domain finetuning. To our best knowledge, we are also the \textit{first} to enable \textit{non-equilibrium} 3D molecule generation via the chemically diverse OMol25 dataset (n.b., containing metal complexes, biomolecules, etc.), representing a 10x data scale-up over prior GEOM-Drugs benchmarking \citep{joshi2025allatom}.


\begin{table}[!t]
\centering
\caption{\textbf{Molecule generation results on GEOM-DRUGS.}
\label{table:geom_dng}
Validity, uniqueness, and \% pass rates on PoseBusters for 10,000 sampled molecules.
\textsc{Zatom-1} matches or exceeds state-of-the-art equivariant/latent diffusion baselines and generates the largest number of PoseBusters-valid samples.
}
\resizebox{0.75\textwidth}{!}{%
\begin{tabular}{lccccc|c}
\toprule
\textbf{Metric (\% pass) $\uparrow$} & \textbf{EQGAT-diff} & \textbf{SemlaFlow} & \textbf{ADiT} & \textbf{TABASCO} & \textbf{\textsc{Zatom-1}} & \textbf{GEOM-Drugs} \\
\midrule
Validity & 94.6 & 93.9 & \underline{95.3} & \textbf{97.6} & 93.6 & - \\
Uniqueness & \textbf{100.0} & \textbf{100.0} & \textbf{100.0} & 99.03 & \underline{99.93} & - \\
Atoms connected & 84.4 & 92.3 & 93.0 & \textbf{99.9} & \underline{99.6} & - \\
Bond angles & 86.9 & 94.8 & 92.3 & \underline{99.2} & \textbf{99.6} & - \\
Bond lengths & 87.0 & 94.6 & 92.5 & \underline{99.4} & \textbf{99.7} & - \\
Aromatic ring flat & 87.0 & 94.9 & \underline{95.4} & \textbf{100.0} & \textbf{100.0} & - \\
Double bond flat & 87.0 & 94.2 & 95.3 & \underline{99.8} & \textbf{100.0} & - \\
Internal energy & 86.8 & 94.8 & 91.3 & \textbf{99.4} & \underline{99.3} & - \\
No steric clash & 82.9 & 92.0 & 91.8 & \underline{94.3} & \textbf{96.4} & - \\
\midrule
PoseBusters valid & 59.7 & 87.5 & 85.3 & \underline{91.6} & \textbf{94.1} & 94.0 \\
\bottomrule
\end{tabular}
}
\end{table}


\begin{table}[!t]
\centering
\caption{\textbf{Property prediction for 3D molecules (QM9) / materials (Matbench), with zero-shot multi-task material predictions.} Results are reported as \textit{test} mean absolute errors (for \textsc{Zatom-1}, with standard deviations over three runs with different random seeds). $\dagger$ denotes using different data partitions. / represents a task for which both (QM9) molecule and (Matbench) material property labels are available. NN denotes pretraining using a Noisy Nodes self-supervised objective \citep{zaidi2023pretraining}. The best (or second-best) results for each optimized (i.e., hyperparameter-tuned) or unoptimized (i.e., non-hyperparameter-tuned) category are in \textbf{bold} (or \underline{underlined}). Notably, \textsc{Zatom-1} (80M \faSnowflake\ + 20M \faFire) and AIM-MTL are the only multi-task learning (MTL) methods represented in these results, which they achieved with only $\sim$100 GPU hours of additional finetuning for molecule property prediction, signifying their computational efficiency and expressivity. In contrast, single-task learning (STL) baselines such as EquiformerV2 (11M) traditionally require $\sim$150 GPU training hours \emph{for each task}, not including their extensive hyperparameter optimization per task. Further, for the majority of QM9's properties, jointly pretrained \textsc{Zatom-1} offers the \textit{first} example of positive transfer learning between molecules and materials for prediction tasks driven by cross-domain generative pretraining.}
\label{table:qm9_matbench_property_prediction_results}
\resizebox{\textwidth}{!}{
\begin{tabular}{llllllllllllll}
\toprule
\textbf{Model / Metrics} $\downarrow$ & 
\# Params & 
\textbf{\makecell{$\alpha$\\ ($a_0^3$)}} &
\textbf{\makecell{$\Delta\varepsilon$\\ (meV)}} &
\textbf{\makecell{$\varepsilon_{\mathrm{HOMO}}$\\ (meV)}} &
\textbf{\makecell{$\varepsilon_{\mathrm{LUMO}}$\\ (meV)}} &
\textbf{\makecell{$\mu$\\ (D)}} &
\textbf{\makecell{$C_\nu$\\ (cal/mol K)}} &
\textbf{\makecell{$G^{\mathrm{ATOM}}$\\ (meV)}} &
\textbf{\makecell{$H^{\mathrm{ATOM}}$\\ (meV)}} &
\textbf{\makecell{$R^2$\\ ($a_0^2$)}} &
\textbf{\makecell{$U^{\mathrm{ATOM}}$\\ (meV)}} &
\textbf{\makecell{$U_0^{\mathrm{ATOM}}$\\ (meV)}} &
\textbf{\makecell{$ZPVE$\\ (meV)}} \\
\midrule
\textbf{Optimized single-task learning} \\
DimeNet++$^\dagger$        & 5M & .044 & 33.0 / \textbf{199} & 25 & 20 & .030 & .023 & 8.00    & 7.00    & .331 & \underline{6.00}   & \underline{6.00}   & \underline{1.21} \\
EGNN$^\dagger$    & 1M & .071 & 48.0 / ---- & 29 & 25 & .029 & .031 & 12.0   & 12.0   & .106 & 12.0  & 11.0  & 1.55 \\
PaiNN               & 1M & .045 & 46.0 / ---- & 28 & 20 & .012 & .024 & \textbf{7.35} & \textbf{5.98} & \underline{.066} & \textbf{5.83} & \textbf{5.85} & 1.28 \\
TorchMD-NET   & 7M & .059 & 36.0 / ---- & 20 (16 w/ NN) & 18 (13 w/ NN) & .011 & .026 & 7.62 & 6.16 & \textbf{.033} & 6.38 & 6.15 & 1.84 \\
SphereNet              & 2M & .046 & 32.0 / ---- & 23 & 18 & .026 & \textbf{.021} & 8.00    & \underline{6.00}    & .292 & 7.00   & 6.00   & \textbf{1.12} \\
SEGNN$^\dagger$ & 1M & .060 & 42.0 / ---- & 24 & 21 & .023 & .031 & 15.0   & 16.0   & .660 & 13.0  & 15.0  & 1.62 \\
EQGAT                   & - & .053 & 32.0 / ---- & 20 & 16 & .011 & .024 & 23.0   & 24.0   & .382 & 25.0  & 25.0  & 2.00 \\
Equiformer          & 4M & \underline{.046} & \underline{30.0} / ---- & \underline{15} & \underline{14} & \underline{.011} & \underline{.023} & 7.63 & 6.63 & .251 & 6.74 & 6.59 & 1.26 \\
EquiformerV2          & 11M & .050 & \textbf{29.0} / ---- & \textbf{14} & \textbf{13} & .010 & .023 & \underline{7.57} & 6.22 & .186 & 6.49 & 6.17 & 1.47 \\
P$\Theta$NITA   & - & \textbf{.038} & 30.4 / ---- & 16 & 15 & .012 & .024 & 8.63 & 8.04 & .235 & 8.67 & 8.31 & 1.29 \\
Platonic Transformer         & - & .049 & 37.4 / ---- & 22 & 17 & \textbf{.010} & .024 & 12.0 & 12.0 & .222 & 11.9 & 13.0 & 1.30 \\
\midrule
\textbf{Unoptimized single-task learning} \\
AIM-STL$^\dagger$         & - & .181 & ---- / ---- & 61 & 54 & \textbf{.067} & .072 & 66.2 & 63.9 & \textbf{.503} & 64.2 & 58.8 & 4.54 \\
\textbf{Unoptimized multi-task learning} \\
AIM-MTL-Scalar$^\dagger$         & - & .268 & ---- / ---- & 59 & 71 & .089 & .103 & 113 & 112 & 4.18 & 112 & 112 & 9.82 \\
AIM-MTL-Matrix$^\dagger$         & - & .251 & ---- / ---- & 61 & 72 & \underline{.088} & .103 & 112 & 118 & 4.07 & 116 & 116 & 12.2 \\
Non-pretrained \textsc{Zatom-1} (\textsc{Ours})         & 80M \faSnowflake\ + 20M \faFire & .140$\pm$.001 & 72.3$\pm$0.1 / 3358$\pm$6 & 54$\pm$0.0 & 49$\pm$0.0 & .157$\pm$.001 & .064$\pm$.000 & 45.5$\pm$0.2 & 44.7$\pm$0.2 & 4.72$\pm$0.02 & 45.2$\pm$0.2 & 45.8$\pm$0.2 & 2.89$\pm$0.01 \\
QM9-pretrained \textsc{Zatom-1} (\textsc{Ours})         & 80M \faSnowflake\ + 20M \faFire & .095$\pm$.001 & 49.5$\pm$0.3 / 5297$\pm$29 & \underline{36$\pm$0.2} & 34$\pm$0.1 & .100$\pm$.000 & .044$\pm$.000 & 23.1$\pm$0.2 & 23.1$\pm$0.2 & 3.63$\pm$0.01 & 23.5$\pm$0.2 & 22.6$\pm$0.2 & 2.04$\pm$0.01 \\
Jointly pretrained \textsc{Zatom-1}, $K=L/4$ (\textsc{Ours})         & 80M \faSnowflake\ + 20M \faFire & \underline{.093$\pm$.000} & 49.1$\pm$0.1 / \textbf{2522}$\pm$\textbf{11} & 36$\pm$0.0 & 34$\pm$0.1 & .111$\pm$.000 & \underline{.042$\pm$.000} & \underline{22.8$\pm$0.2} & \underline{22.6$\pm$0.1} & 3.62$\pm$0.01 & \textbf{22.6}$\pm$\textbf{0.2} & \textbf{22.1}$\pm$\textbf{0.2} & \textbf{1.97}$\pm$\textbf{0.00} \\
Jointly pretrained \textsc{Zatom-1}, $K=L/2$ (\textsc{Ours})         & 80M \faSnowflake\ + 20M \faFire & .099$\pm$.001 & 51.2$\pm$0.3 / \underline{2640$\pm$8} & 37$\pm$0.1 & 34$\pm$0.2 & .121$\pm$.001 & .044$\pm$.000 & 24.1$\pm$0.1 & 23.3$\pm$0.2 & 3.96$\pm$0.01 & 23.5$\pm$0.2 & 23.4$\pm$0.1 & 2.01$\pm$0.01 \\
Jointly pretrained \textsc{Zatom-1}, $K=3L/4$ (\textsc{Ours})         & 80M \faSnowflake\ + 20M \faFire & .096$\pm$.000 & \underline{49.1$\pm$0.2} / 4327$\pm$14 & \underline{36$\pm$0.2} & \underline{33$\pm$0.1} & .125$\pm$.000 & .044$\pm$.000 & \textbf{22.0}$\pm$\textbf{0.2} & \textbf{22.3}$\pm$\textbf{0.2} & 4.00$\pm$0.01 & \underline{22.7$\pm$0.2} & \underline{22.4$\pm$0.2} & 1.99$\pm$0.00 \\
\rowcolor{orange!30} Jointly pretrained \textsc{Zatom-1} (\textsc{Ours})         & 80M \faSnowflake\ + 20M \faFire & \textbf{.091}$\pm$\textbf{.001} & \textbf{46.2}$\pm$\textbf{0.2} / 3891$\pm$20 & \textbf{34}$\pm$\textbf{0.1} & \textbf{32}$\pm$\textbf{0.2} & .090$\pm$.001 & \textbf{.041}$\pm$\textbf{.000} & 23.3$\pm$0.4 & 23.0$\pm$0.4 & \underline{3.29$\pm$0.01} & 24.0$\pm$0.4 & 23.3$\pm$0.4 & \underline{1.98$\pm$0.01} \\
\bottomrule
\end{tabular}}
\end{table}

\section{Conclusion}
\label{section:dicussions}

In this work, we introduced \textsc{Zatom-1}, a general-purpose model architecture for 3D molecules and materials, as a step towards a true foundation model architecture for 3D chemistry. Empirical results show that \textsc{Zatom-1} achieves state-of-the-art 3D molecule and materials generation, with fast training and inference speeds. Additionally, it demonstrates the advantages of multi-domain generative pretraining for downstream prediction tasks, providing state-of-the-art multi-task molecular property predictions for QM9, which marks SciML's first instance of positive inter-domain transfer for prediction tasks through cross-domain generative pretraining. Supplementary experiments, including those with the QMOF and OMol25 datasets, confirm that \textsc{Zatom-1} can also learn effective representations of complex atomistic systems, in spite of its current limitations (see Appendix \ref{appendix:additional_model_details}). Notably, given \textsc{Zatom-1}'s all-atom architectural scalability and versatility, future work performing parallel model and data scaling, in addition to integrating new data domains, tasks, and modalities (e.g., proteins or RNA), may unlock new foundational capabilities within and beyond the chemical sciences. As a generalized model for 3D chemistry, \textsc{Zatom-1}'s code and weights are freely available.

\begin{ack}
This research used resources of the National Energy Research Scientific Computing Center, a DOE Office of Science User Facility supported by the Office of Science of the U.S. Department of Energy under Contract No. DE-AC02-05CH11231 using AI4Sci@NERSC award NERSC DDR-ERCAP0036206 awarded to AM. NBE would like to acknowledge support from the U.S. Department of Energy, Office of Science, Office of Advanced Scientific Computing Research, EXPRESS: 2025 Exploratory Research for Extreme-Scale Science program, and the Scientific Discovery through Advanced Computing (SciDAC) program, under Contract Number DE-AC02-05CH11231 at Berkeley Lab. Lastly, we would also like to thank Luis Pinto and Ali Ramlaoui for their insightful feedback on the \textsc{Zatom-1} model architecture.
\end{ack}

\clearpage
\bibliography{zatom_1}
\bibliographystyle{abbrvnat}


\appendix

\appendixtoc
\clearpage

\section{Related Work}
\label{appendix:related_work}

\subsection{Scientific foundation models}
The concept of scientific foundation models, which are large, self-supervised networks pretrained on broad scientific data, has recently gained momentum in chemistry and materials science \citep{PyzerKnapp2025, Choi2025FMchem, edamadaka2025universally}. Recent reviews emphasize the value of training on diverse chemical datasets and using generative or multi-task objectives to build broadly useful chemical representations \citep{PyzerKnapp2025, Choi2025FMchem, wadell2025foundation}. However, most existing works remain domain-specific: "foundation” efforts in molecules often focus on SMILES or 2D graph pretraining, while material science approaches largely target property prediction or synthesis planning rather than joint 3D modeling.

\subsection{Generative modeling of 3D molecules}
Several recent works propose equivariant diffusion or flow models for de novo 3D molecule design. \citet{hoogeboom2022equivariant} introduce an E(3)-equivariant diffusion model (EDM) that denoises continuous atom coordinates and types together. Autoregressive graph models like Symphony \citep{daigavane2024symphony} build molecules fragment-by-fragment with spherical harmonics features. \citet{le2024navigating} explore E(3)-equivariant diffusion losses (EQGAT-diff) to improve validity on QM9 and GEOM-Drugs. Latent diffusion approaches like GeoLDM \citep{xu2023geometric} encode 3D structures into continuous latent spaces. More recent work explores the importance of equivariance: for instance, SemlaFlow uses flow matching with an equivariant graph neural network backbone \citep{irwin2025semlaflow}; FlowMol3 employs hybrid continuous-discrete flow matching with chirality-aware message passing \citep{Dunn2025FlowMol3}; and TABASCO shows that a non-equivariant transformer can match the validity of equivariant models while generating 3D drug-like molecules \citep{vonessen2026tabasco}. In summary, these models each focus on molecule generation but do not generalize to materials.

\subsection{Generative modeling of 3D materials}
Likewise, generative models for periodic crystals have been developed separately. Score-based models (diffusion) and flow matching have been applied to crystal generation: e.g., \citet{xie2022crystal} propose CDVAE, a score-based model for generating periodic unit cells, and \citet{jiao2023crystal} introduce DiffCSP, an E(3)-equivariant diffusion model for predicting stable crystal structures. Recent work often uses variants of normalizing flows: \citet{miller2024flowmm}’s FlowMM applies Riemannian flow matching over atom types and lattice geometry, and \citet{yang2024scalable} show that standard (non-equivariant) diffusion on a learnable crystal representation (UniMat) can scale to large collections of structures. Another parallel thread is generative models for inorganic materials like those of \citet{zeni2025generative}, \citet{hollmer2025open}, \citet{park2025guiding}, \citet{okhotin2025miad}, \citet{yi2026crystaldit}, \citet{seong2026multimodal}, and \citet{veljkovic2026crystalite}. FlowLLM combines a fine-tuned language model base with flow matching for crystals \citep{sriram2024flowllm}, while the All-Atom Diffusion Transformer (ADiT) unifies molecule and material generation by learning a shared latent autoencoding of both domains and diffusing in that latent space \citep{joshi2025allatom}. Notably, none of these approaches simultaneously addresses both molecules and materials in one model without resorting to complex two-stage training approaches, such as latent diffusion or LLM-based data generation, nor do they leverage generative pretraining for downstream tasks.

\subsection{Multimodal flow/diffusion architectures}
Our work joins a growing class of multimodal generative models that handle discrete atom types and continuous 3D geometries jointly. For example, \citet{miller2024flowmm} explicitly matches flows over categorical atom labels, continuous coordinates, and lattice parameters in a single framework. ADiT \citep{joshi2025allatom} analogously embeds all-atom molecular and crystal representations into a shared latent space for latent diffusion. These approaches stand in contrast to earlier graph or point-based generative models (e.g., EDM \citep{hoogeboom2022equivariant}, EQGAT-diff \citep{le2024navigating}) that typically model only a small number of data modalities simultaneously. By jointly training on atom types, positions, and unit cell parameters, our approach builds on the multimodal flow concept of FlowMM while using a standard Transformer backbone for expressiveness, scalability, and speed.

\subsection{Representation learning and pretraining in chemistry}
Beyond full generative models, chemical representation learning has employed self-supervised pretraining on molecular graphs. For instance, GraphMVP \citep{liu2022graphmvp} uses contrastive learning between 2D molecular topology and 3D geometry. Masked autoencoder approaches and property prediction pretraining have also been explored. However, most existing pretraining methods do not support generative modeling for sample generation \citep{zhou2023unimol, zaidi2023pretraining}. Even though image diffusion models have been shown to yield useful feature embeddings \citep{li2023image, velez2025video}, few works have investigated cross-domain generative pretraining as a widely applicable paradigm in 3D chemistry or AI for Science more broadly. Accordingly, our work frames conditional multimodal flow matching (over atom types and geometry) as a self-supervised “pretraining” task for 3D chemical systems. By pretraining on joint molecule \& material data, we obtain unified embeddings transferable to property, energy, and force prediction. This idea is similar to \citet{Chang2024SPMM}'s structure-property molecular model, which learns a bidirectional generative model of molecular graphs and properties, but here we apply it broadly across both molecules and crystals.

\subsection{Overview}
Overall, prior work in molecular/material generation has focused either on discrete (graph/SMILES) models or on separate generative flows/diffusions for molecules or crystals \citep{hoogeboom2022equivariant, xie2022crystal}. Our approach differs by unifying generative and predictive modeling for all-atom 3D chemistry. We draw on advances in flow matching and multimodal architecture design \citep{song2023equivariant, miller2024flowmm, vonessen2026tabasco}, and we treat generative modeling itself as a pretraining strategy. This differs from previous scientific foundation modeling efforts that focus on language or 2D graph representations \citep{PyzerKnapp2025, Choi2025FMchem, wadell2025foundation}; that do not support downstream prediction tasks with learned representations \citep{zhang2025unigenx}; that are not trained across multiple data domains \textit{end-to-end} \citep{lu2026unified}; or that autoregressively assume a canonical token ordering for unordered sets of atoms \citep{zhang2025unigenx, lu2026unified}. By explicitly modeling the ambient 3D space of atoms and elements together with a standardized Transformer architecture, we extend the flow-based generative paradigm to create and evaluate a single, general-purpose model architecture across 3D molecular and material domains.

\section{Evaluation Metrics}
\label{appendix:evaluation_metrics}

\subsection{Model development metrics for assessing crystal generation quality}
To evaluate the quality of generated crystals during (iterative) model \textit{development}, we adopt the methodology proposed by \citet{xie2022crystal}, \citet{miller2024flowmm}, and \citet{joshi2025allatom}. This process involves generating a sample of 1,000 crystals and then assessing them based on their validity and uniqueness. These metrics are described below:
\begin{itemize}
    \item \textbf{Structural Validity}: This metric represents the percentage of crystals where all pairwise atomic distances are at least 0.5 and the crystal volume is 0.1 or greater.
    \item \textbf{Compositional Validity}: In accordance with the SMACT framework \citep{davies2019smact}, this is the percentage of crystal compositions that successfully maintain both charge neutrality and electronegativity balance.
    \item \textbf{Overall Validity}: This is the percentage of crystals that satisfy both the structural and the compositional validity criteria mentioned above. Note that this metric is adopted in \cref{figure:model_scaling_results,appendix:figure:crystals_molecules_mp20_qm9_inference_sweep}.
    \item \textbf{Unique}: This measures the percentage of crystals that are structurally unique. Uniqueness is determined via an all-to-all comparison using the Structure Matcher tool within PyMatGen \citep{ong2013python}.
\end{itemize}

\subsection{Model evaluation metrics for assessing crystal generation quality}
To assess the quality of generated crystals during (extensive) model \textit{evaluation}, we adopt the methodology proposed by \citet{duval2025lemat}. This process involves generating a sample of 2,500 crystals and then evaluating them based on their validity, uniqueness, novelty, energy, and stability. These metrics are detailed below:
\begin{itemize}
    \item \textbf{Charge Validity}: This metric ensures structures are charge-balanced using oxidation state analysis and bond valence calculations.
    \item \textbf{Distance Validity}: This validates that atomic distances exceed minimum thresholds based on atomic radii.
    \item \textbf{Coordination Validity}: This checks if coordination numbers match expected values for each element.
    \item \textbf{Physical Validity}: This validates density, lattice parameters, crystallographic format, and symmetry.
    \item \textbf{Overall Validity}: This represents the percentage of crystals meeting each of the evaluation validity criteria above.
    \item \textbf{Unique}: This measures the percentage of \textit{valid} crystals that are also structurally unique. Uniqueness is determined via an all-to-all comparison using PyMatGen's Structure Matcher tool.
    \item \textbf{Novel}: This assesses the percentage of \textit{valid}, \textit{unique} crystals that are also structurally novel. Novelty is determined via an all-to-all comparison between the generated crystals and the LeMat-Bulk reference set \citep{siron2025lemat} using the Structure Matcher tool of PyMatGen.
    \item \textbf{Formation Energy}: This denotes the average formation energy across multiple MLIPs, namely Orb-v3 \citep{rhodes2025orb}, MACE-MP \citep{Batatia2025Foundation}, and UMA \citep{wood2025uma}.
    \item \textbf{Mean Energy Above Hull}: This represents the average energy above hull across multiple MLIPs (i.e., Orb-v3, MACE-MP, UMA).
    \item \textbf{Relaxation Stability}: This signifies the root mean square deviation (RMSD) between original and relaxed atomic positions.
    \item \textbf{Stability Ratio}: This denotes the fraction of structures with energy above hull $\leq$ 0 eV/atom (thermodynamically stable).
    \item \textbf{Stable, Unique, \& Novel (SUN)}: This represents the fraction of structures that are simultaneously valid, stable (energy above hull $\leq$ 0 eV/atom), unique, and novel, with the number of structures that meet each of these criteria shown in proceeding parentheses.
    \item \textbf{Metastability Ratio}: This signifies the fraction of structures with 0 $<$ energy above hull $\leq$ 0.1 eV/atom (metastable).
    \item \textbf{Metastable, Unique, \& Novel (MSUN)}: This denotes the fraction of structures that are simultaneously valid, metastable (0 $<$ energy above hull $\leq$ 0.1 eV/atom), unique, and novel, with the number of structures that meet each of these criteria shown in proceeding parentheses.
\end{itemize}

\subsection{Model development/evaluation metrics for assessing molecule generation quality}
For the evaluation of generated molecules, we utilize the protocol developed by \citet{hoogeboom2022equivariant}, \citet{daigavane2024symphony}, and \citet{joshi2025allatom}. From a sample of 10,000 molecules, we calculate rates for validity and uniqueness, and also determine success rates for seven distinct sanity checks provided by the PoseBusters toolkit \citep{buttenschoen2024posebusters}:
\begin{itemize}
    \item \textbf{Validity}: This measures the percentage of generated molecules for which a canonical SMILES representation can be successfully identified using RDKit \citep{landrum2013rdkit}.
    \item \textbf{Uniqueness}: Among the molecules deemed valid, this metric calculates the percentage that possess a unique SMILES string.
    \item \textbf{All-atoms Connected}: This check quantifies the percentage of molecules in which every atom is part of a single connected component, meaning a continuous path of bonds links all atoms.
    \item \textbf{Reasonable Bond Angles/Lengths}: This is the fraction of molecules where all bond angles and lengths fall within a tolerance range, specifically between 0.75 and 1.25 times the standard values derived from distance geometry.
    \item \textbf{Aromatic Rings Flatness}: This metric counts the percentage of molecules where all atoms in 5- or 6-membered aromatic rings are positioned within 0.25\AA\ of their nearest shared plane.
    \item \textbf{Double Bonds Flatness}: This is the percentage of molecules where the atoms of aliphatic carbon-carbon double bonds, along with their four neighboring atoms, lie within 0.25\AA\ of the closest shared plane.
    \item \textbf{Reasonable Internal Energy}: This check determines the percentage of molecules whose calculated internal energy does not exceed 100 times the average energy observed in an ensemble of 50 conformations of the input molecule \citep{harris2023posecheck}.
    \item \textbf{No Internal Steric Clash}: This metric represents the percentage of molecules where the distance between any pair of non-covalently bound atoms is greater than 0.8 times the lower bound established by distance geometry.
\end{itemize}
In summary, the validity and uniqueness metrics primarily confirm the chemical processability of the generated molecules by RDKit. In contrast, the PoseBusters sanity checks scrutinize the physical realism of the generated 3D structures, applying a range of criteria from geometric constraints to energetic plausibility.

\subsection{Model evaluation metrics for assessing cross-domain transfer learning}
\label{appendix:model_evaluation_metrics_for_assessing_cross_domain_transfer_learning}
To assess whether a model is positively leveraging cross-domain transfer learning between molecules and materials during pretraining, we extend the embedding visualization protocol developed by \citet{joshi2025allatom}. From a random sample of 100 molecules and materials, each generated by \textsc{Zatom-1}, as well as 100 molecules and materials drawn from QM9 and MP20, respectively, we calculate the Clustering \& Transfer Learning Score (higher is better) as follows:
\begin{itemize}
    \item \textbf{Clustering \& Transfer Learning Score ($\uparrow$)}: A summary score for 2D (e.g., PCA-derived) model embeddings. It is defined as Inter-Element Separation / (Intra-Element Spread + Dataset-Generated Shift + Crystal-Molecule Shift), so it is larger when atom types are well separated, while the same element remains aligned across source (e.g., dataset, generated) and system (e.g., molecule, material) categories.
    \item \textbf{Inter-Element Separation ($\uparrow$)}: This is the mean pairwise Euclidean distance between the 2D centroids of element clusters (C, N, O, F). Larger values indicate stronger separation of different atom types.
    \item \textbf{Intra-Element Spread ($\downarrow$)}: This is the mean Euclidean distance from each point to the centroid of its own element cluster. Smaller values indicate that atoms of the same type form tighter clusters.
    \item \textbf{Dataset-Generated Shift ($\downarrow$)}: This is the mean Euclidean distance between centroids of dataset-derived and generated samples for the same element and system type. Smaller values indicate that generated atoms are embedded similarly to dataset atoms within molecules and within crystals, reducing distribution shifts between generated and real data.
    \item \textbf{Crystal-Molecule Shift ($\downarrow$)}: This is the mean Euclidean distance between crystal and molecule centroids for the same element and source type. Smaller values indicate that, e.g., oxygen atoms in molecules and crystals occupy similar regions of the embedding space, suggesting the model has learned to leverage positive transfer learning between molecules and materials.
\end{itemize}

\clearpage

\section{Visualization}
\label{appendix:visualization} 

\subsection{Characterizing chemical transfer learning in shared latent space}
\label{appendix:characterizing_chemical_transfer_learning_in_shared_latent_space}
\begin{figure}[p] 
    \centering
    
    \begin{subfigure}{0.48\textwidth}
        \centering
        \includegraphics[width=\linewidth]{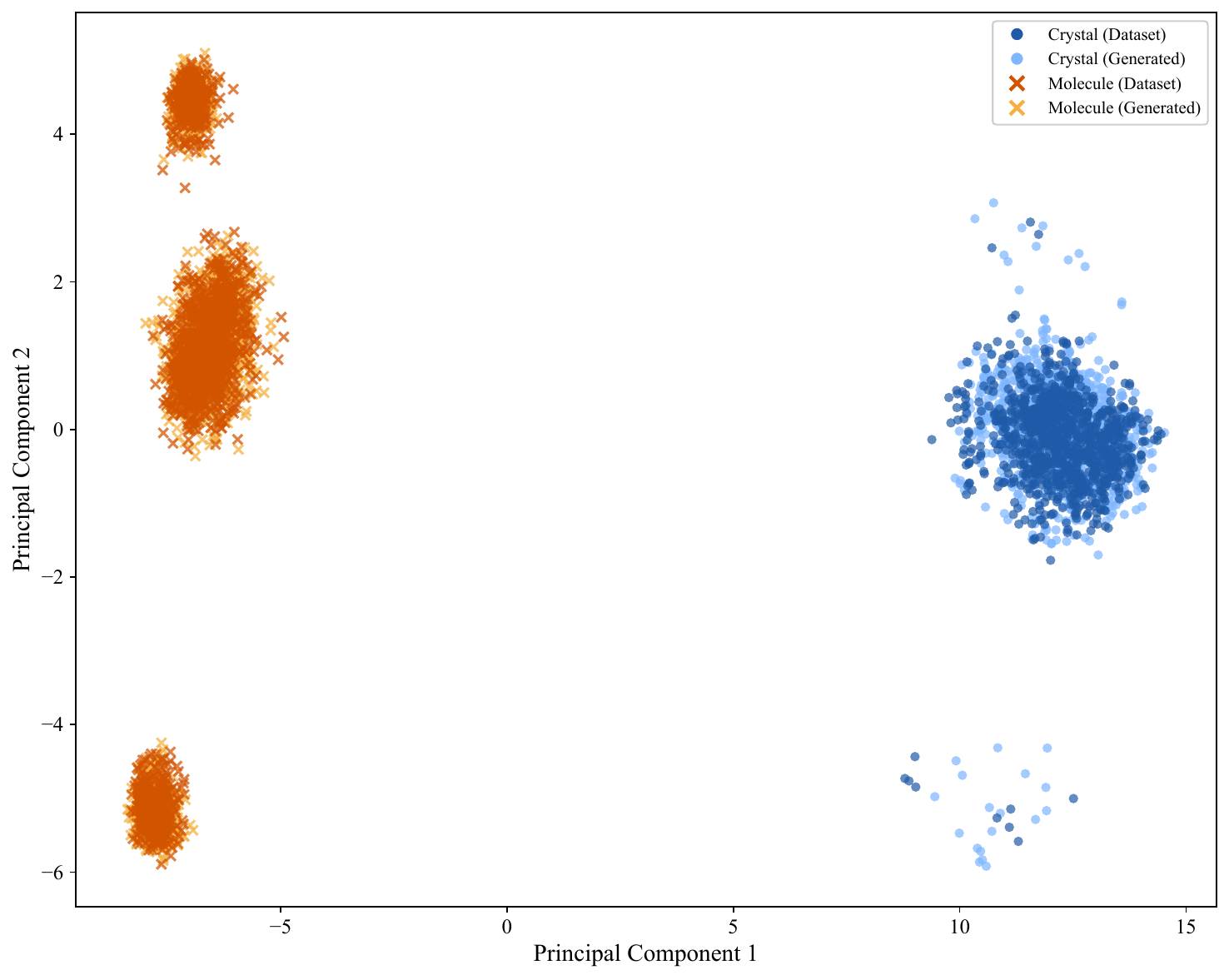}
        \caption{Untrained model}
        \label{appendix:figure:pca_untrained_latent_space}
    \end{subfigure}
    \hfill 
    \begin{subfigure}{0.48\textwidth}
        \centering
        \includegraphics[width=\linewidth]{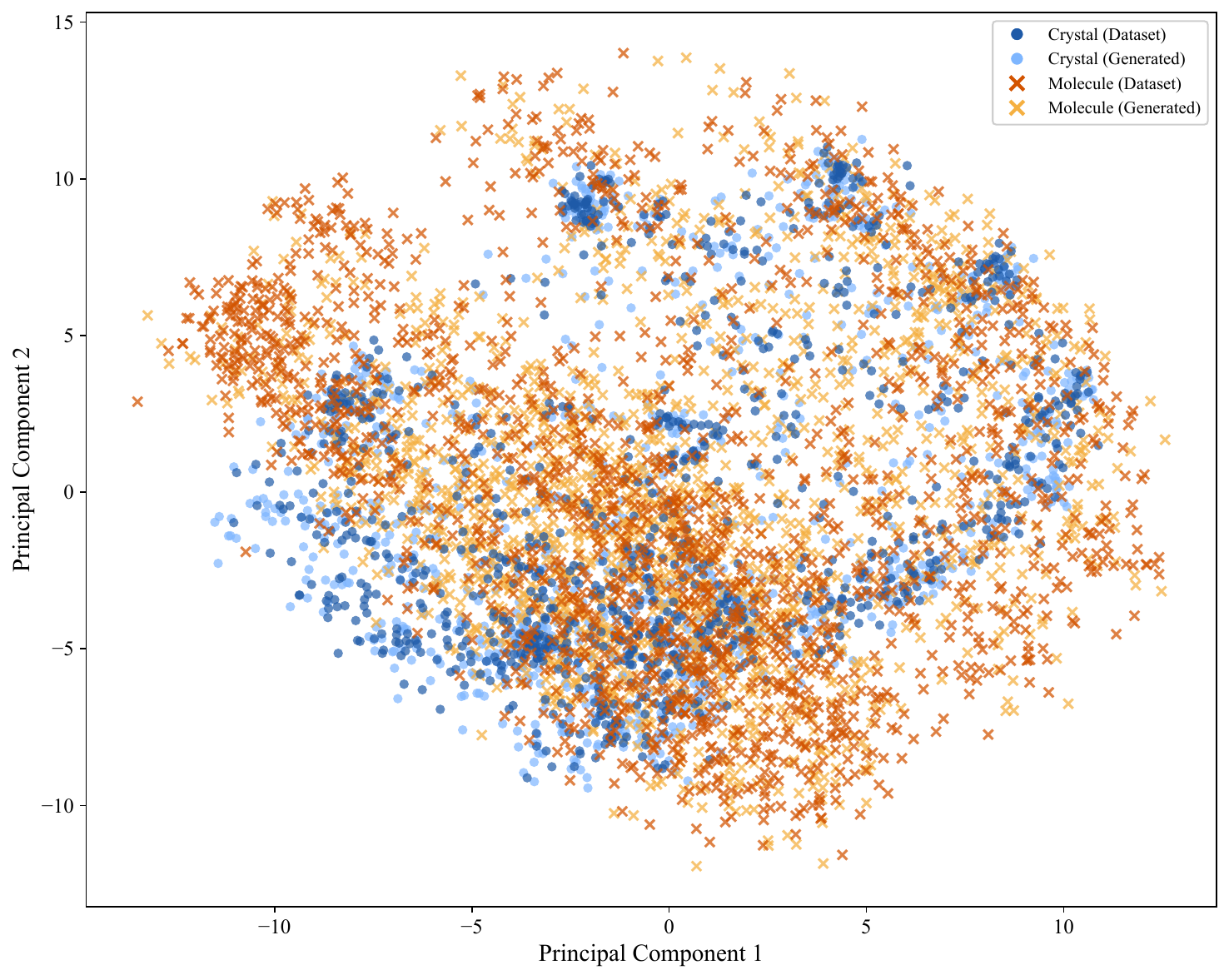}
        \caption{QM9-only pretrained model}
        \label{appendix:figure:pca_qm9_only_latent_space}
    \end{subfigure}

    \vspace{1cm} 

    \begin{subfigure}{0.48\textwidth}
        \centering
        \includegraphics[width=\linewidth]{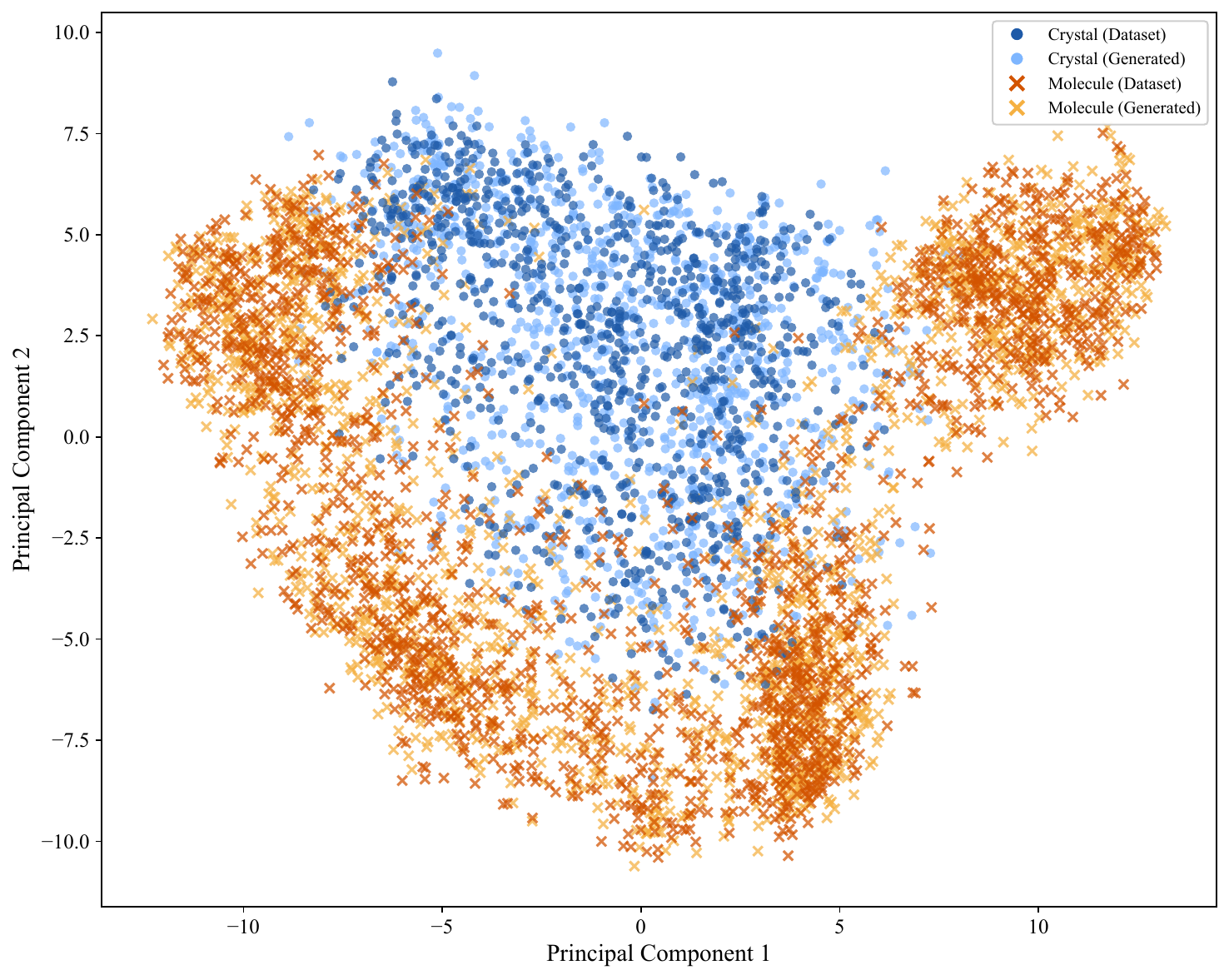}
        \caption{MP20-only pretrained model}
        \label{appendix:figure:pca_mp20_only_latent_space}
    \end{subfigure}
    \hfill
    \begin{subfigure}{0.48\textwidth}
        \centering
        \includegraphics[width=\linewidth]{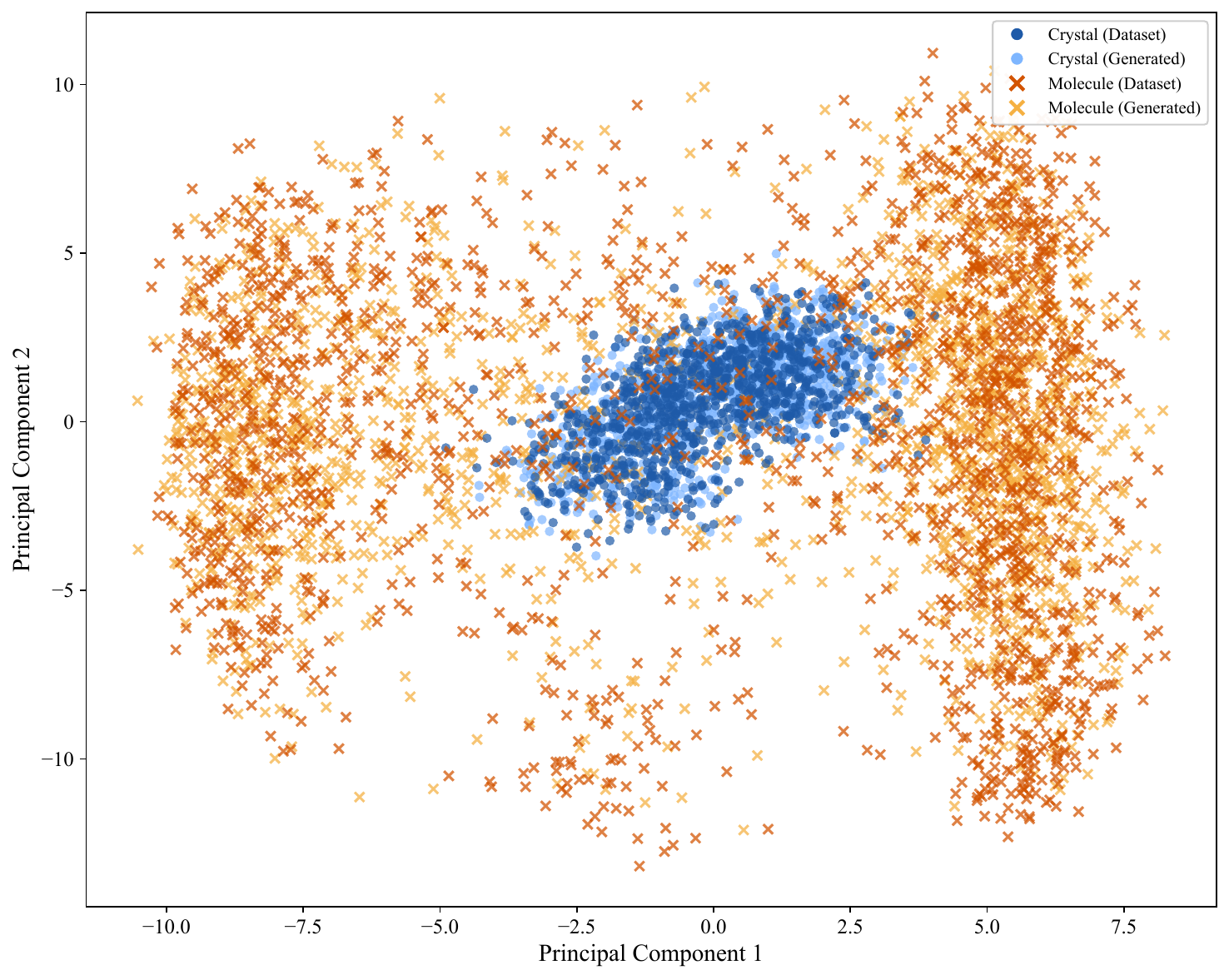}
        \caption{Jointly pretrained model}
        \label{appendix:figure:pca_jointly_pretrained_latent_space}
    \end{subfigure}

    \caption{PCA plots of latent embeddings from \textsc{Zatom-1}’s Transformer trunk for 100 data points from the MP20 and QM9 datasets, as well as 100 molecules/materials each generated by \textsc{Zatom-1}. Each point denotes an atom (including hydrogens for molecules), with its shape representing its chemical system type and its primary color indicating whether it comes from real or generated data. \textbf{(a)} \textbf{Embeddings taken from an untrained trunk are distributed in an overly clustered and chemically-segmented manner.} \textbf{(b-c)} \textbf{Embeddings taken from a trunk pretrained on either QM9 or MP20 form less well-structured molecule and material clusters.} \textbf{(d)} \textbf{Embeddings taken from a jointly pretrained trunk, however, show structured yet overlapping clusters, suggesting positive representation transfer between molecules and materials has occurred}.}
    \label{appendix:figure:PCA_latent_space_comparison}
\end{figure}

In Figure \ref{appendix:figure:PCA_latent_space_comparison}, following \citep{joshi2025allatom}, we plot the first two PCA principal components of \textsc{Zatom-1}'s final-layer ($K=L$) Transformer trunk embeddings for 100 random samples each drawn from the MP20 and QM9 validation datasets, as well as 100 generated crystals and 100 generated molecules sampled from jointly pretrained \textsc{Zatom-1}. Overall, we observe that a jointly pretrained \textsc{Zatom-1} Transformer trunk, compared to one left untrained or pretrained using only QM9 or MP20, produces a latent space with well-structured (yet overlapping) clusters between molecule and crystal atoms (including hydrogens for molecules), suggesting positive representation transfer has occurred between these two chemical domains.

\begin{figure}[p] 
    \centering

    \begin{subfigure}{0.48\textwidth}
        \centering
        \includegraphics[width=\linewidth]{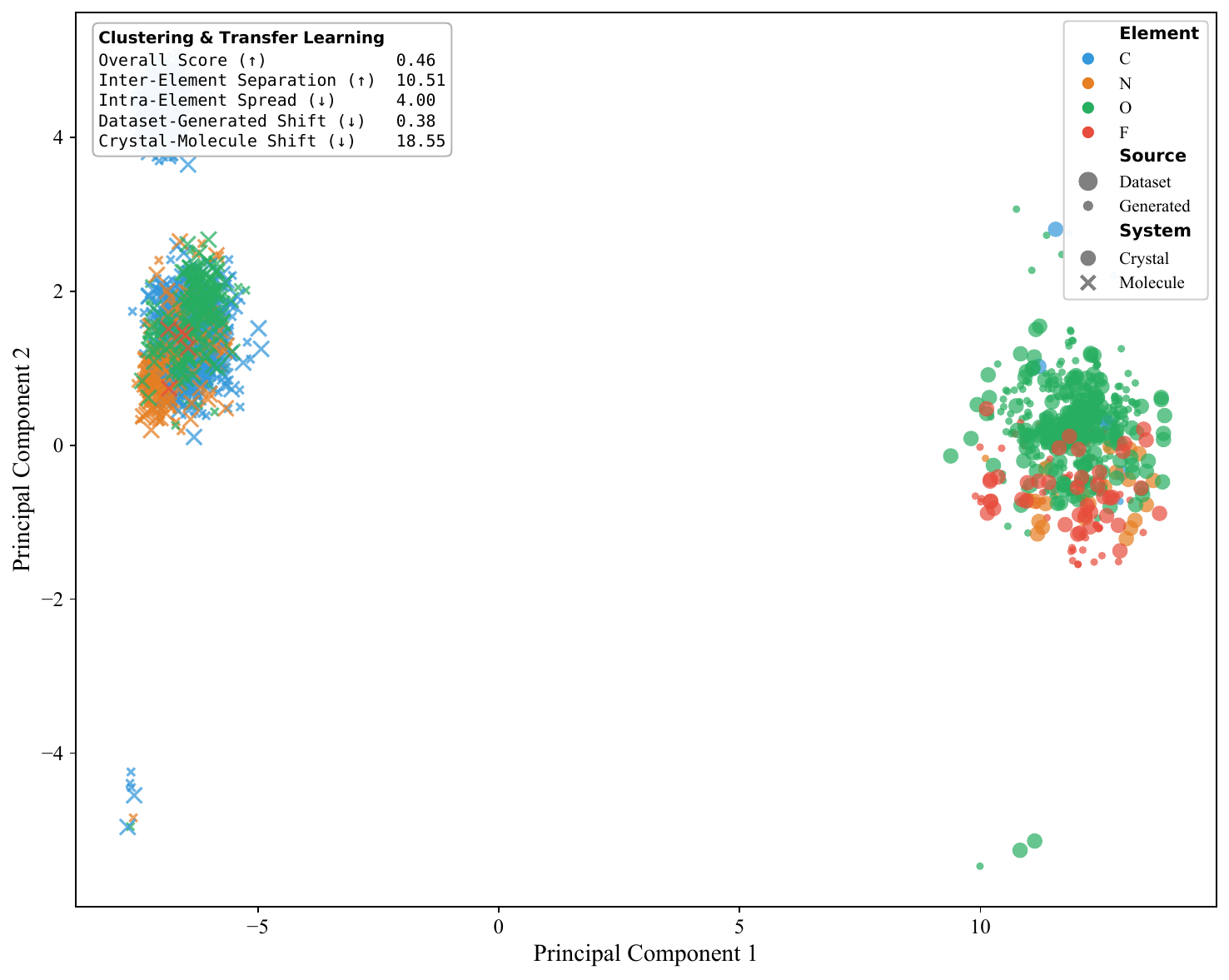}
        \caption{Untrained model}
        \label{appendix:figure:pca_untrained_element_embeddings}
    \end{subfigure}
    \hfill 
    \begin{subfigure}{0.48\textwidth}
        \centering
        \includegraphics[width=\linewidth]{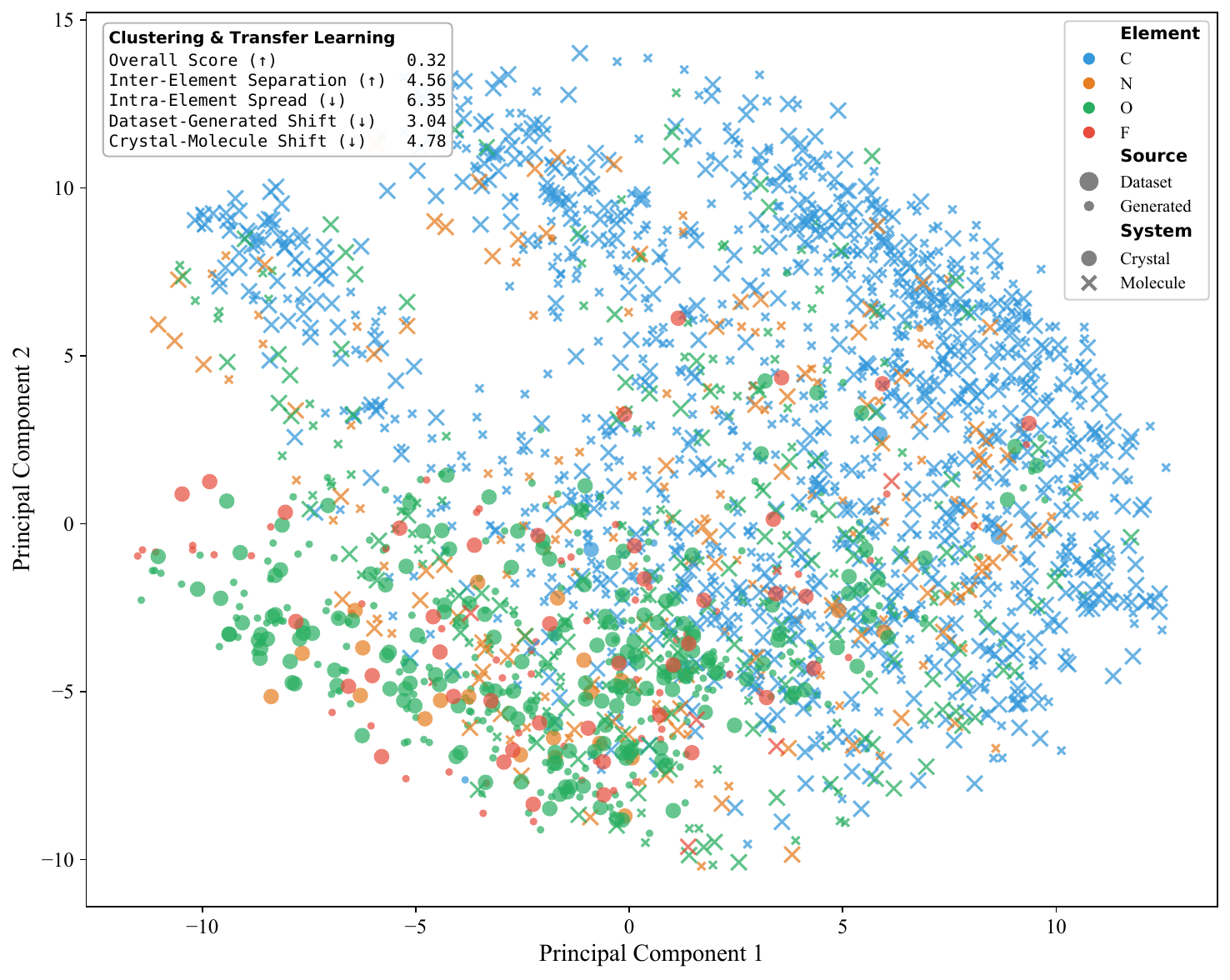}
        \caption{QM9-only pretrained model}
        \label{appendix:figure:pca_qm9_only_element_embeddings}
    \end{subfigure}

    \vspace{1cm} 

    \begin{subfigure}{0.48\textwidth}
        \centering
        \includegraphics[width=\linewidth]{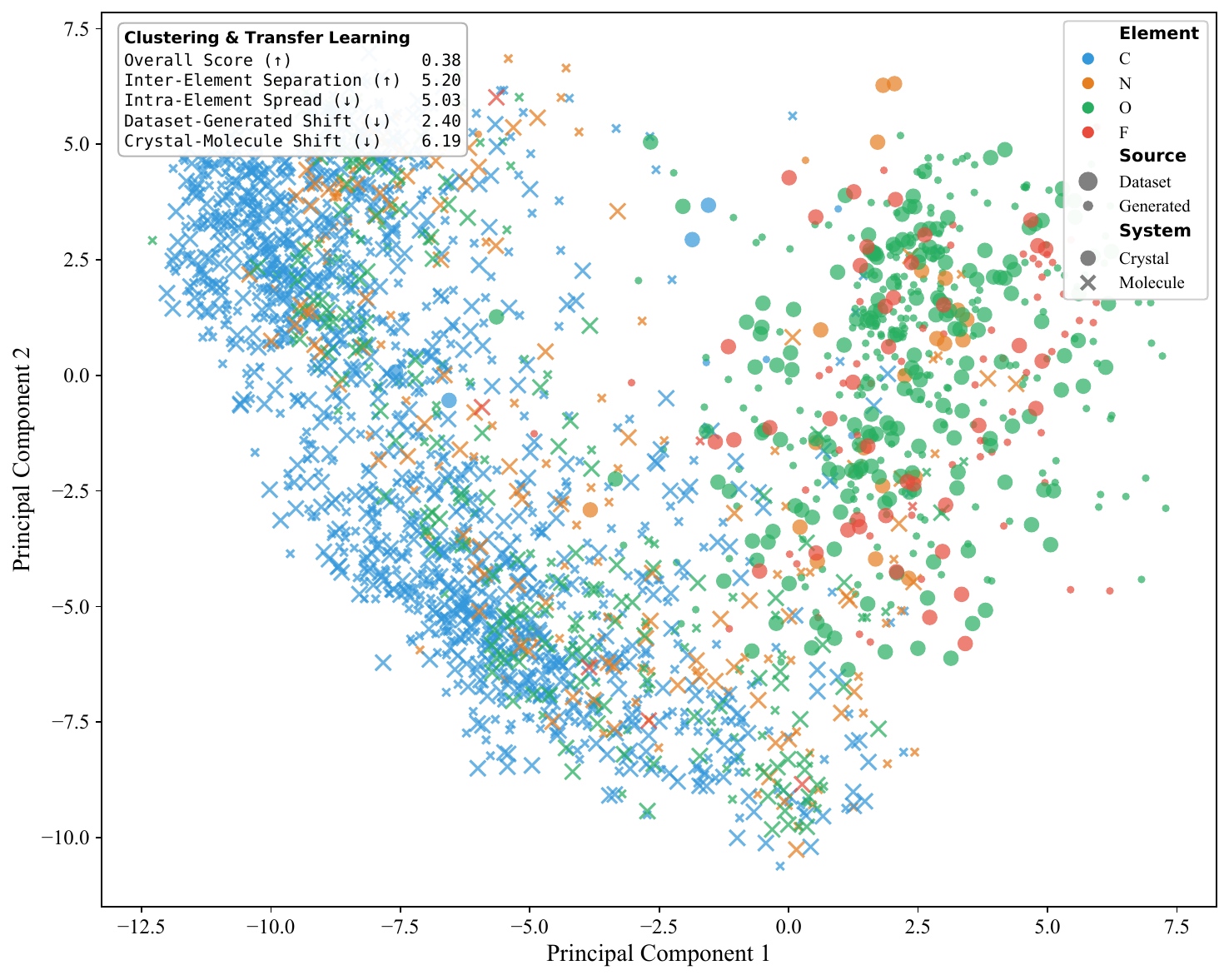}
        \caption{MP20-only pretrained model}
        \label{appendix:figure:pca_mp20_only_element_embeddings}
    \end{subfigure}
    \hfill
    \begin{subfigure}{0.48\textwidth}
        \centering
        \includegraphics[width=\linewidth]{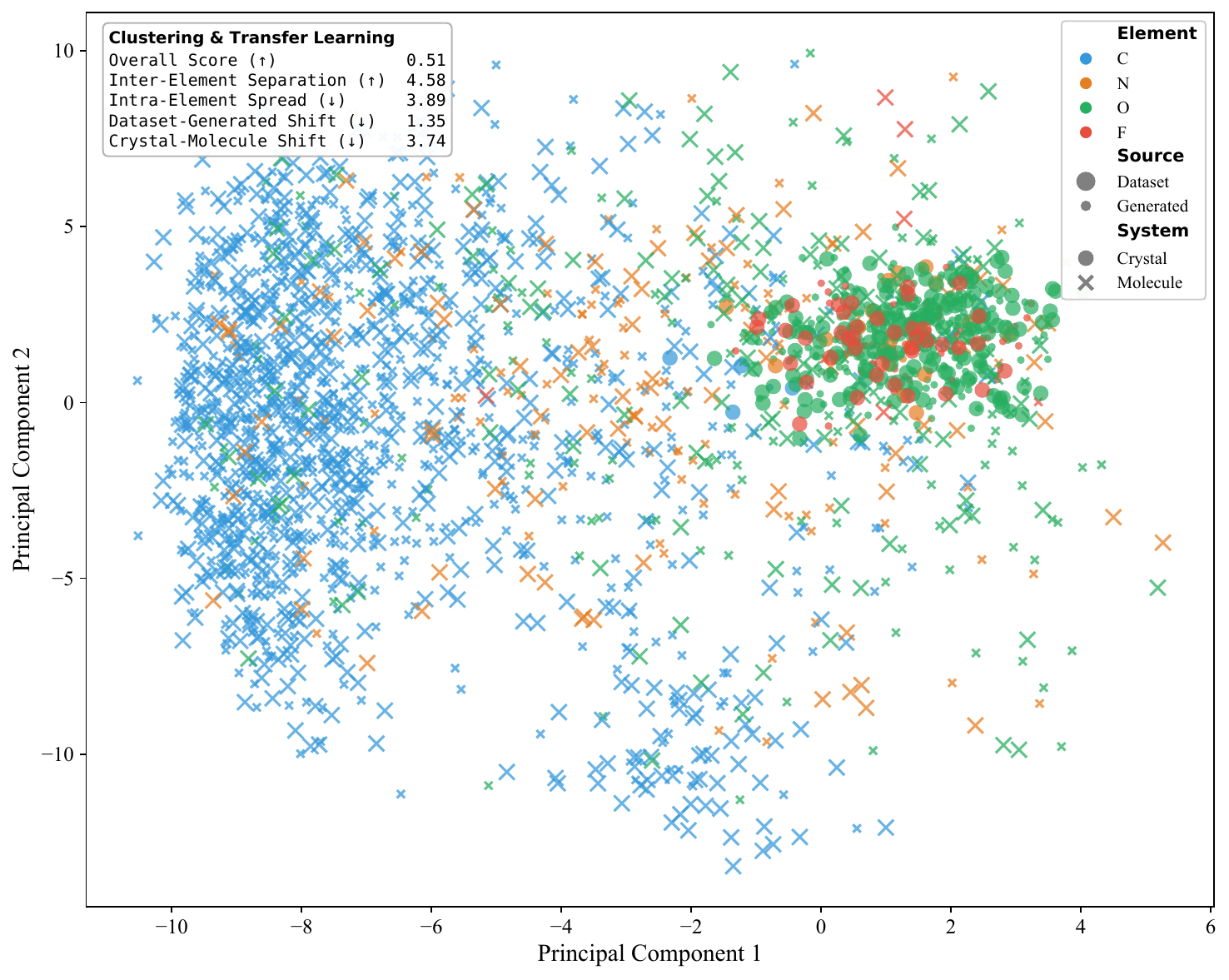}
        \caption{Jointly pretrained model}
        \label{appendix:figure:pca_jointly_pretrained_element_embeddings}
    \end{subfigure}

    \caption{PCA plots of latent embeddings for carbon, nitrogen, oxygen, and fluorine atoms from \textsc{Zatom-1}’s Transformer trunk for 100 data points from the MP20 and QM9 datasets, as well as 100 molecules/materials each generated by \textsc{Zatom-1}. Each point denotes an atom, with its shape representing its chemical system type, its size defined by whether it comes from real or generated data, and its color indicating its element type. \textbf{(a)} Embeddings taken from an untrained trunk are distributed in an overly clustered and chemically-segmented manner, with a quantitatively high crystal-molecule shift between atom embeddings of the same element and source type. \textbf{(b-c)} Embeddings taken from a trunk pretrained on either QM9 or MP20 form less well-structured molecule and material clusters, with a quantitatively higher intra-element spread and crystal-molecule shift. \textbf{(d)} A jointly pretrained model, however, produces trunk embeddings that achieve the best quantitative balance across all clustering and transfer learning metrics. Moreover, in such a model, principal component 1 largely relates to whether a system is a crystal (within range -2 – 4) or a molecule, while principal component 2, generally speaking, relates to an atom's type. \textbf{Joint pretraining induces a latent space that shows distinct clusters for different atom types, with oxygen (and, to some extent, fluorine) atoms having similar representations in both molecules and crystals. Notably, the overlap in oxygen atom representations suggests that \textsc{Zatom-1}’s latent space captures shared chemical properties across non-periodic and periodic systems, enabling effective knowledge transfer during joint pretraining.}}
    \label{appendix:figure:PCA_element_embeddings_comparison}
\end{figure}

To quantify and further characterize this observation, in Figure \ref{appendix:figure:PCA_element_embeddings_comparison}, we plot the same PCA of \textsc{Zatom-1}'s trunk embeddings but now only retain carbon, nitrogen, oxygen, and fluorine atoms, as these atoms appear in both QM9 molecules and MP20 crystals, allowing us to analyze how their representations compare across non-periodic and periodic systems. For this experiment, we propose a new metric called the \textbf{Clustering \& Transfer Learning Score} (defined in Appendix \ref{appendix:model_evaluation_metrics_for_assessing_cross_domain_transfer_learning}), which assesses how well a model's embeddings produce well-separated clusters of different elements while ensuring the same element remains aligned across source (e.g., dataset, generated) and system (e.g., molecule, material) categories. Accordingly, Figure \ref{appendix:figure:PCA_element_embeddings_comparison} demonstrates clearly the following two patterns. (\textbf{1}) Embeddings taken from a trunk pretrained on either QM9 or MP20 form \textit{less} well-structured molecule and material clusters, with a quantitatively higher intra-element spread and crystal-molecule shift, while a jointly pretrained trunk produces embeddings that achieve the \textit{best} Clustering \& Transfer Learning Score (n.b., which remains the \textit{same best score} of 0.51 when jointly pretraining on an equal amount of molecule + material data (13.5k + 13.5k) as the MP20-only or QM9-only baseline models, suggesting that the \textit{cross-domain diversity} of \textsc{Zatom-1}'s pretraining data \citep{xie2023doremi, zhang2025harnessing} is the driving factor behind its strong clustering and transfer learning results). (\textbf{2}) For a jointly pretrained model, principal component 1 largely distinguishes between
crystals (clustered between -2 and 4) and molecules, while principal component 2, generally speaking, relates to an atom's type. Most notably, for a jointly pretrained trunk, oxygen (and, to some extent, fluorine) atoms show similar latent representations whether they appear in molecules or crystals, suggesting that \textsc{Zatom-1}'s latent space captures fundamental chemical properties that transfer across both chemical domains. This shared representation of oxygen, a key atom in both datasets, may partially explain \textsc{Zatom-1}’s successful joint pretraining with periodic and non-periodic systems.

\subsection{Generated crystals, molecules, and MOFs}
\begin{figure}[h!]
    \centering
    \begin{subfigure}{0.3\linewidth}
        \centering
        \includegraphics[width=.6\linewidth]{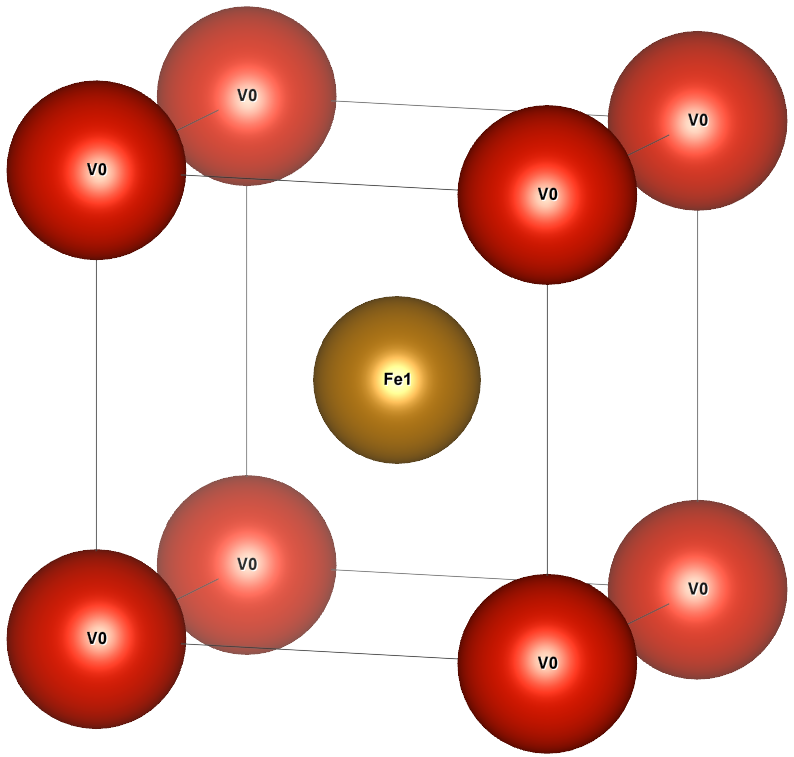}
        \caption{V1 Fe1 (Pm-3m)}
    \end{subfigure}%
    \begin{subfigure}{0.3\linewidth}
        \centering
        \includegraphics[width=.6\linewidth]{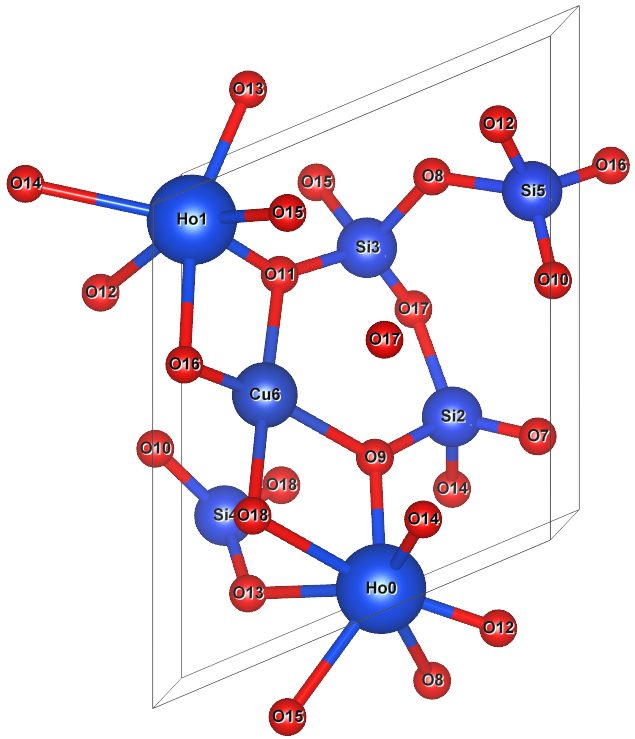}
        \caption{Ho2 Cu1 Si4 O12 (P1)}
    \end{subfigure}%
    \begin{subfigure}{0.3\linewidth}
        \centering
        \includegraphics[width=.6\linewidth]{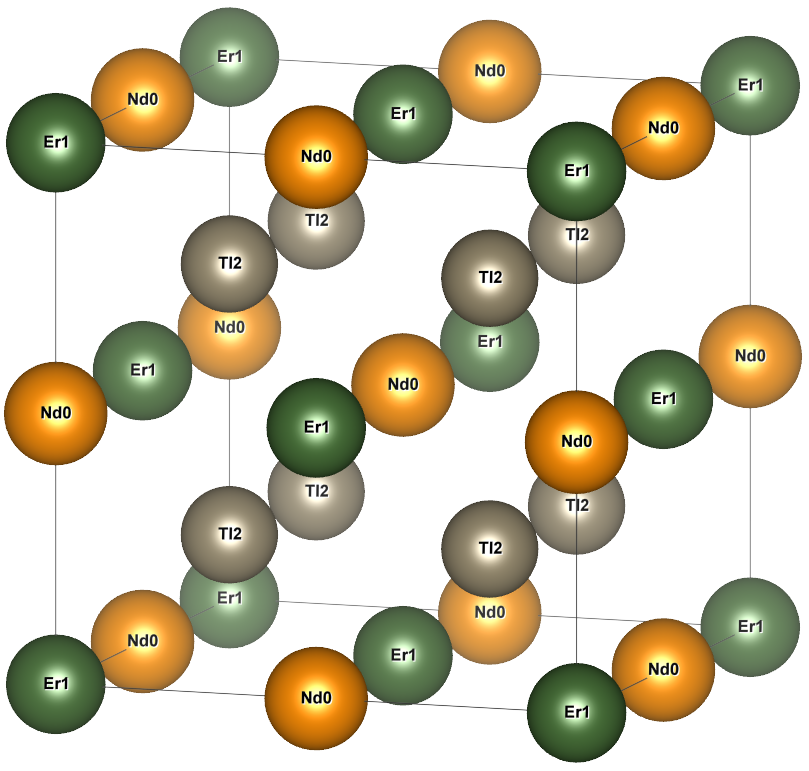}
        \caption{Nd4 Er4 Tl8 (Fm-3m)}
    \end{subfigure}
    \begin{subfigure}{0.3\linewidth}
        \centering
        \includegraphics[width=.6\linewidth]{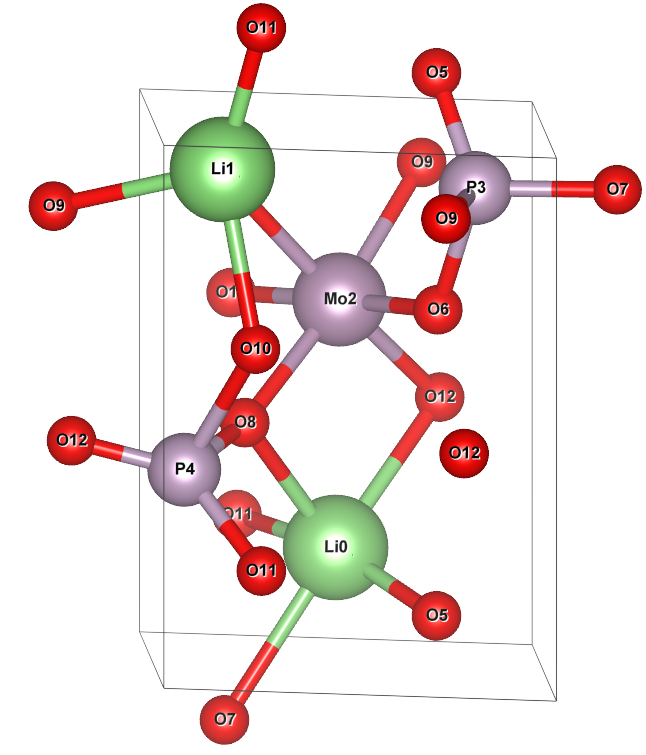}
        \caption{Li2 Mo1 P2 O8 (P1)}
    \end{subfigure}%
    \begin{subfigure}{0.3\linewidth}
        \centering
        \includegraphics[width=.6\linewidth]{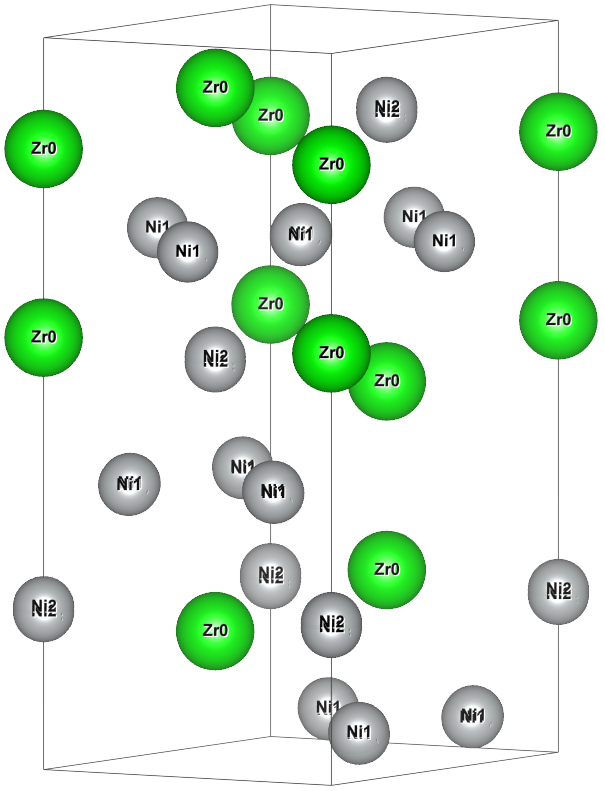}
        \caption{Zr6 Ni12 (R3m)}
    \end{subfigure}%
    \begin{subfigure}{0.3\linewidth}
        \centering
        \includegraphics[width=.6\linewidth]{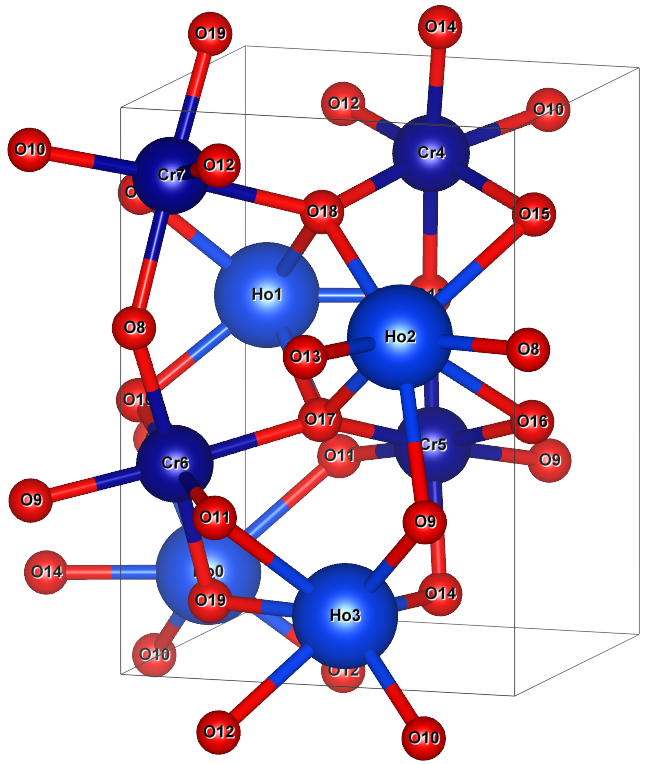}
        \caption{Ho4 Cr4 O12 (P1)}
    \end{subfigure}
    \begin{subfigure}{0.3\linewidth}
        \centering
        \includegraphics[width=.6\linewidth]{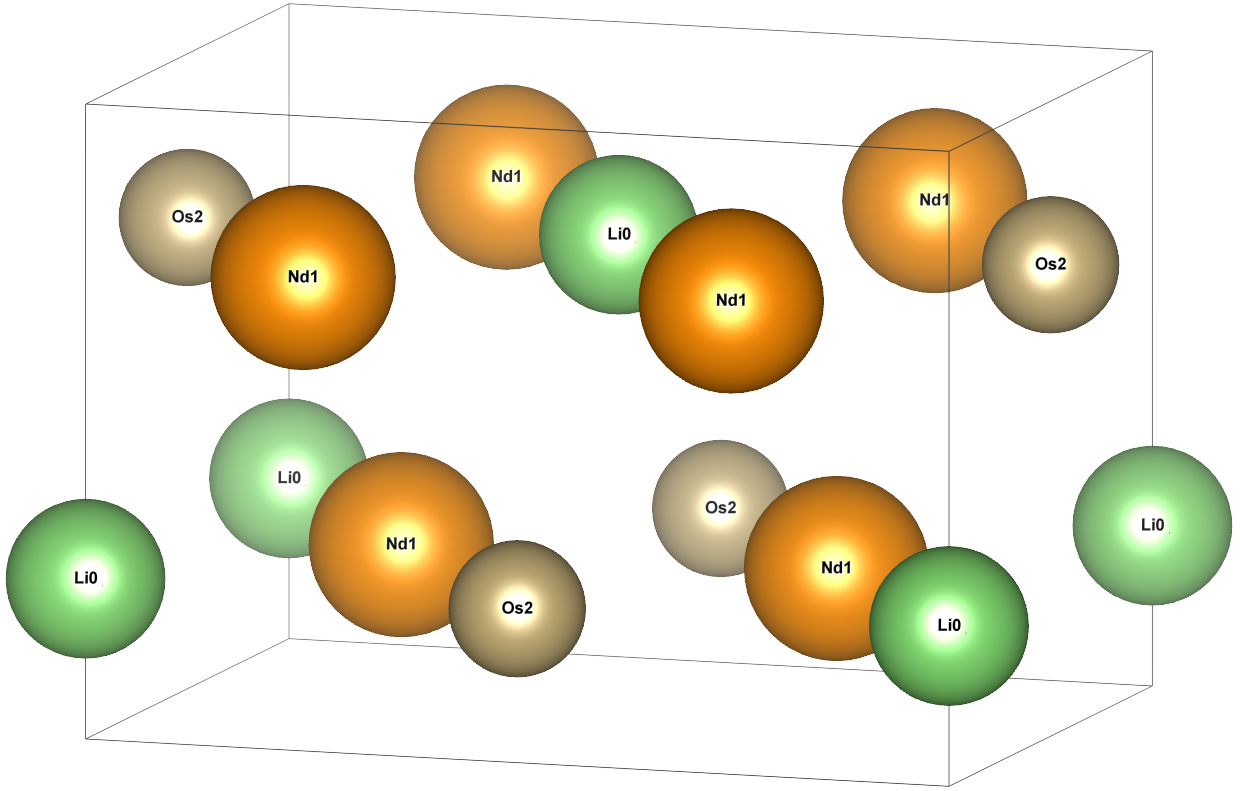}
        \caption{Li2 Nd4 Os2 (Imm2)}
    \end{subfigure}%
    \begin{subfigure}{0.3\linewidth}
        \centering
        \includegraphics[width=.6\linewidth]{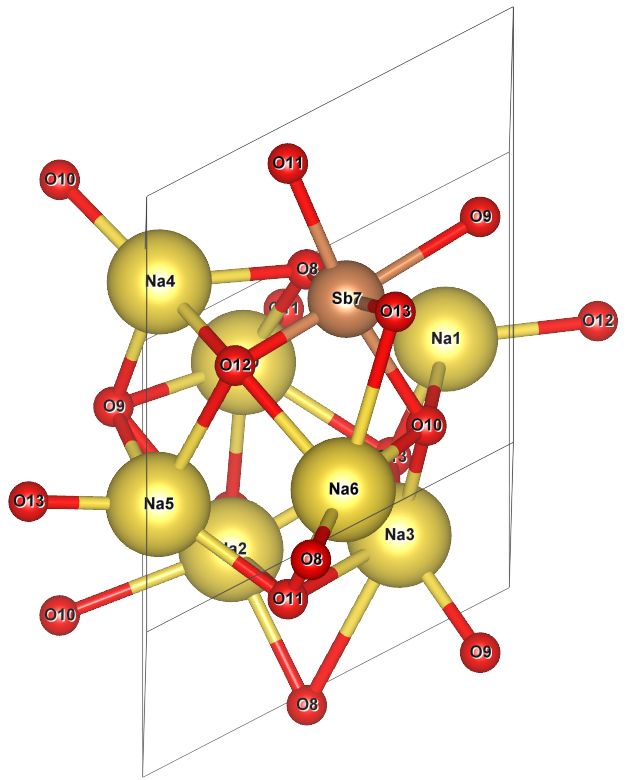}
        \caption{Na7 Sb1 O6 (P1)}
    \end{subfigure}%
    \begin{subfigure}{0.3\linewidth}
        \centering
        \includegraphics[width=.6\linewidth]{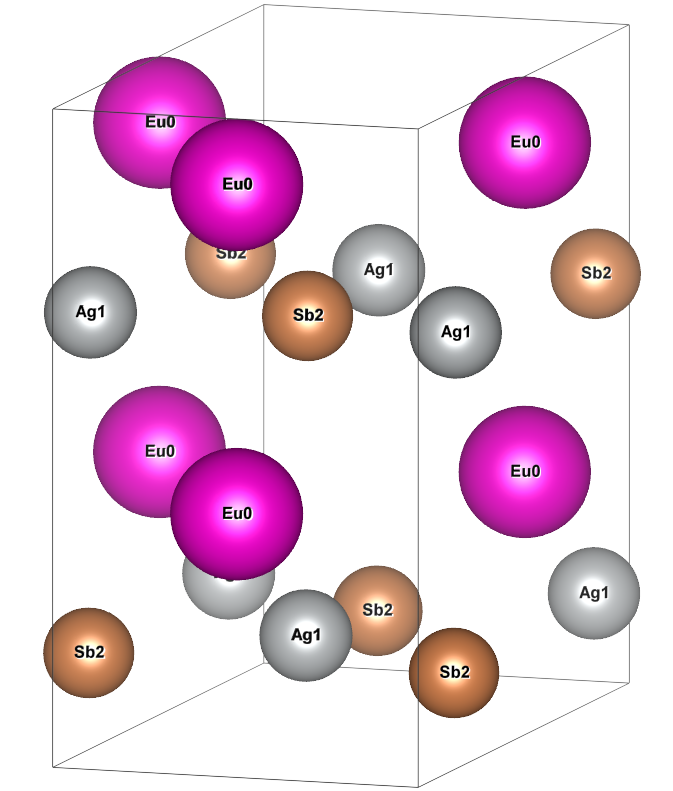}
        \caption{Eu4 Ag4 Sb4 (Cm)}
    \end{subfigure}

    \caption{Generated crystals from \textsc{Zatom-1} trained jointly on MP20 materials and QM9 molecules.}
    \label{appendix:figure:generated_crystals}
\end{figure}

\begin{figure}[t]
    \centering
    \includegraphics[width=0.9\textwidth]{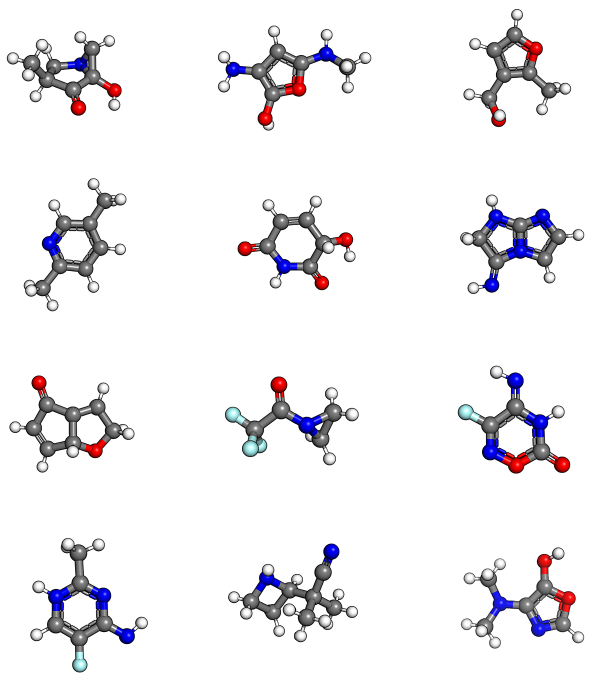}
    \caption{Generated molecules from \textsc{Zatom-1} trained jointly on MP20 materials and QM9 molecules.}
    \label{appendix:figure:generated_molecules}
\end{figure}

\begin{figure}[h!]
    \centering
    \begin{subfigure}{0.3\linewidth}
        \centering
        \includegraphics[width=.75\linewidth]
        {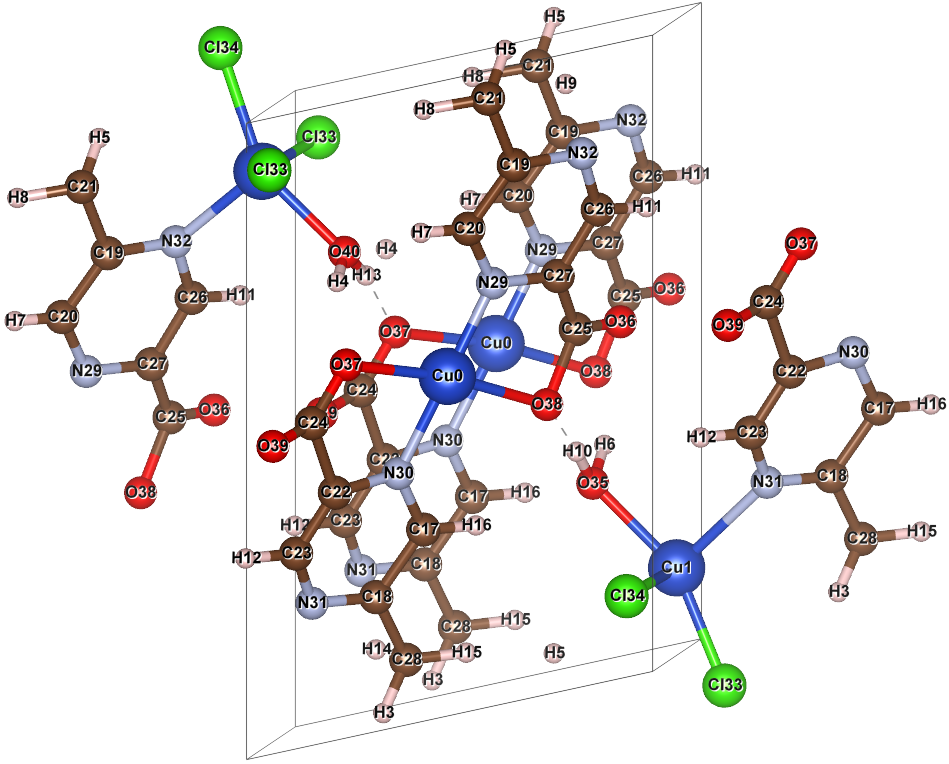}
        \caption{Cu3 H14 C12 N4 Cl2 O6 (P1)}
    \end{subfigure}%
    \begin{subfigure}{0.3\linewidth}
        \centering
        \includegraphics[width=.75\linewidth]{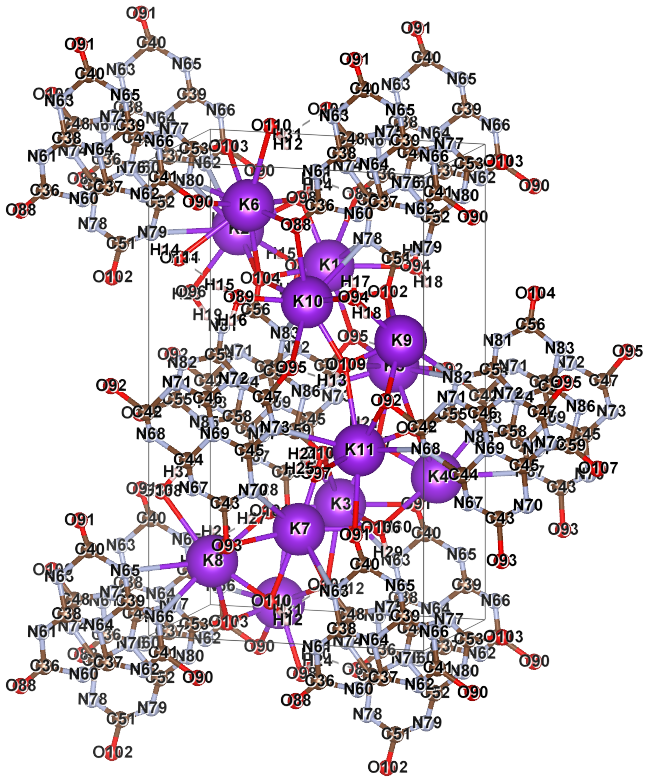}
        \caption{K12 H24 C24 N28 O24 (P1)}
    \end{subfigure}%
    \begin{subfigure}{0.3\linewidth}
        \centering
        \includegraphics[width=.75\linewidth]{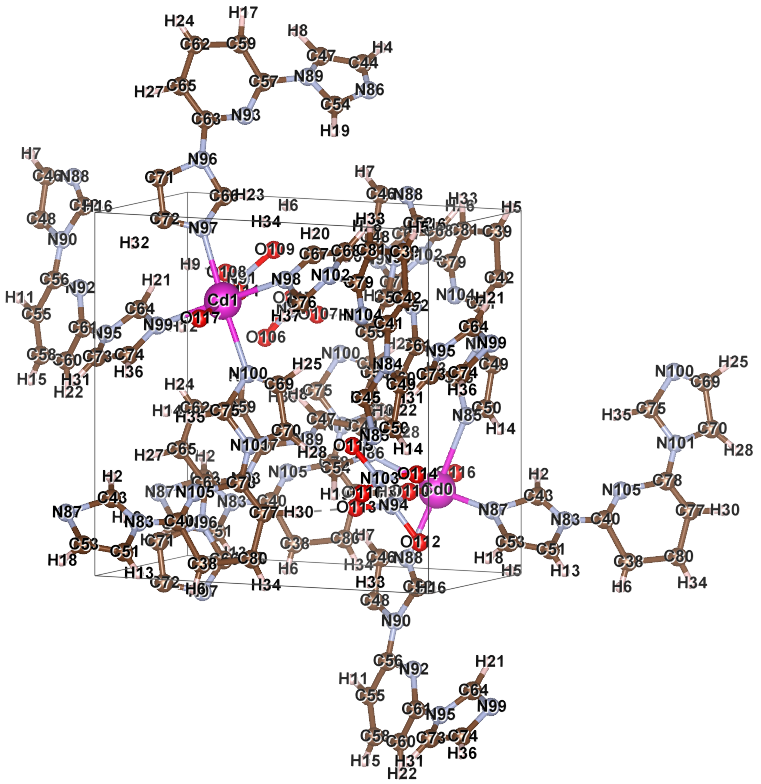}
        \caption{Cd2 H36 C44 N24 O12 (P1)}
    \end{subfigure}

    \begin{subfigure}{0.3\linewidth}
        \centering
        \includegraphics[width=.75\linewidth]{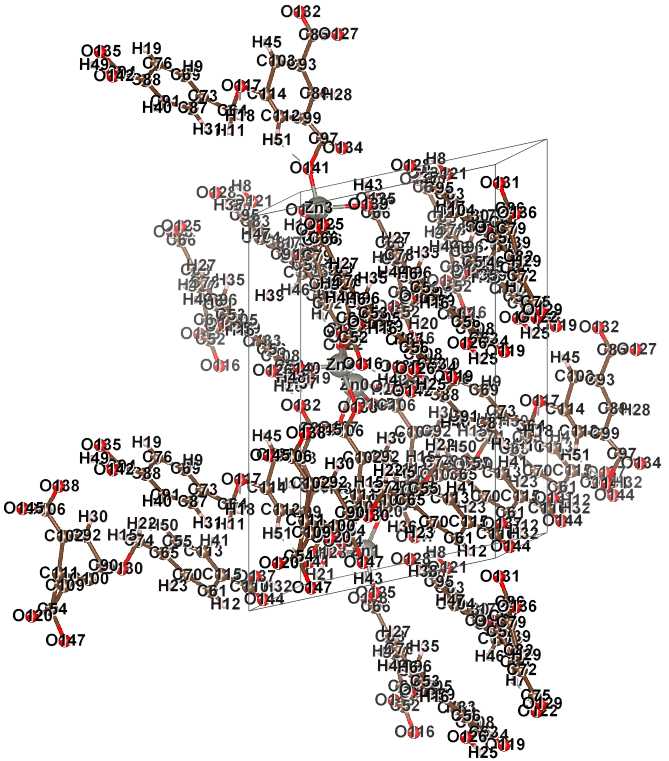}
        \caption{Zn4 H48 C64 O32 (P1)}
    \end{subfigure}%
    \begin{subfigure}{0.3\linewidth}
        \centering
        \includegraphics[width=.75\linewidth]{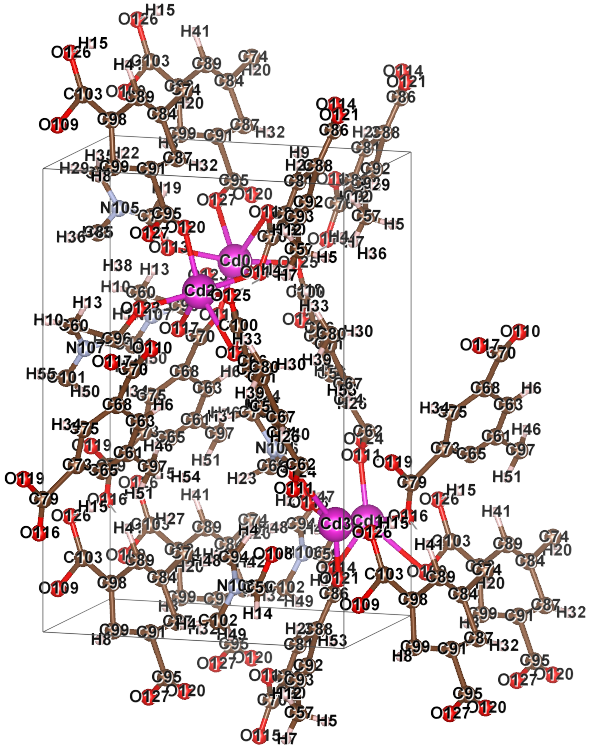}
        \caption{Cd4 H52 C48 N4 O20 (P1)}
    \end{subfigure}%
    \begin{subfigure}{0.3\linewidth}
        \centering
        \includegraphics[width=.75\linewidth]{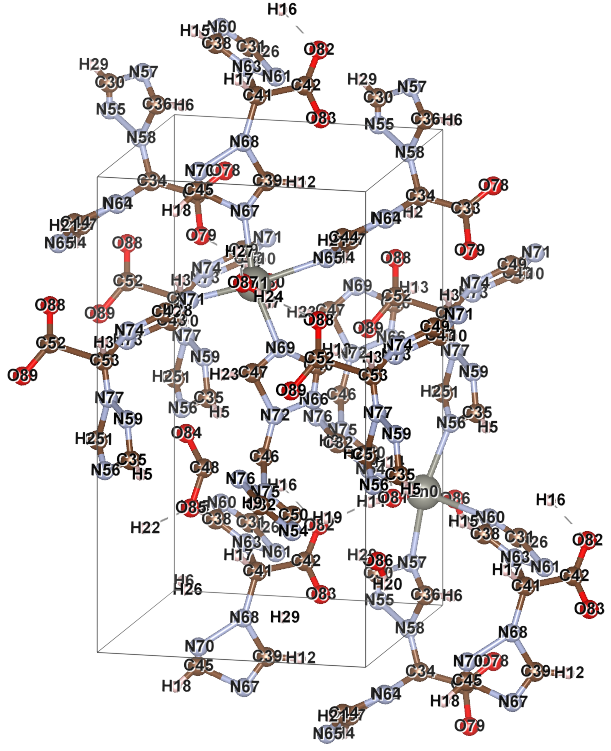}
        \caption{Zn2 H28 C24 N24 O12 (P1)}
    \end{subfigure}

    \begin{subfigure}{0.3\linewidth}
        \centering
        \includegraphics[width=.75\linewidth]{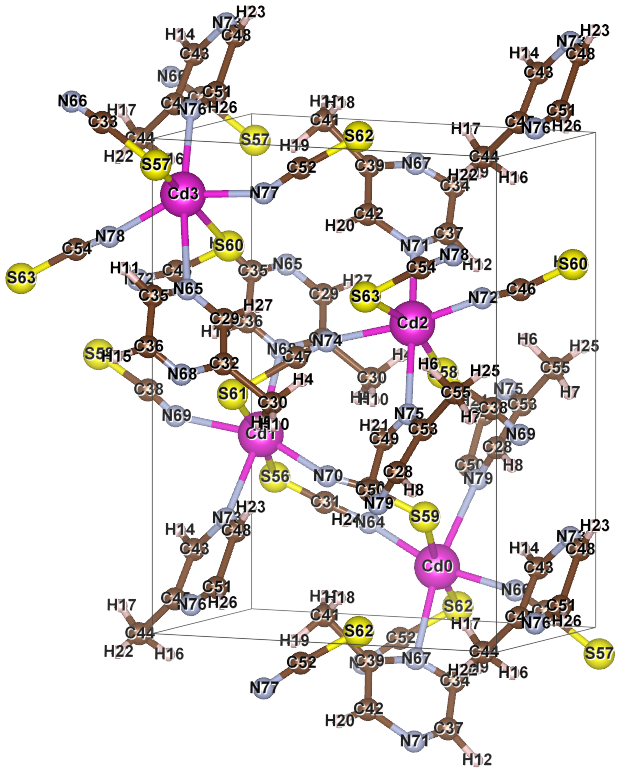}
        \caption{Cd4 H24 C28 S8 N16 (P1)}
    \end{subfigure}%
    \begin{subfigure}{0.3\linewidth}
        \centering
        \includegraphics[width=.75\linewidth]{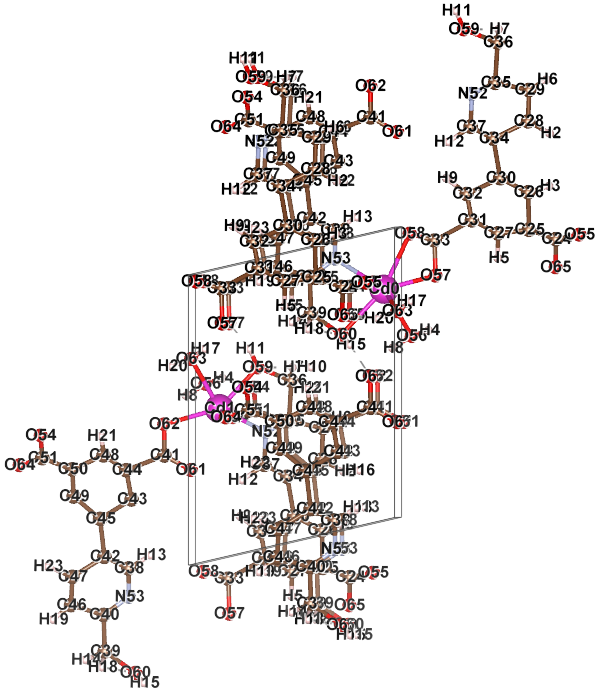}
        \caption{Cd2 H22 C28 N2 O12 (P1)}
    \end{subfigure}%
    \begin{subfigure}{0.3\linewidth}
        \centering
        \includegraphics[width=.75\linewidth]{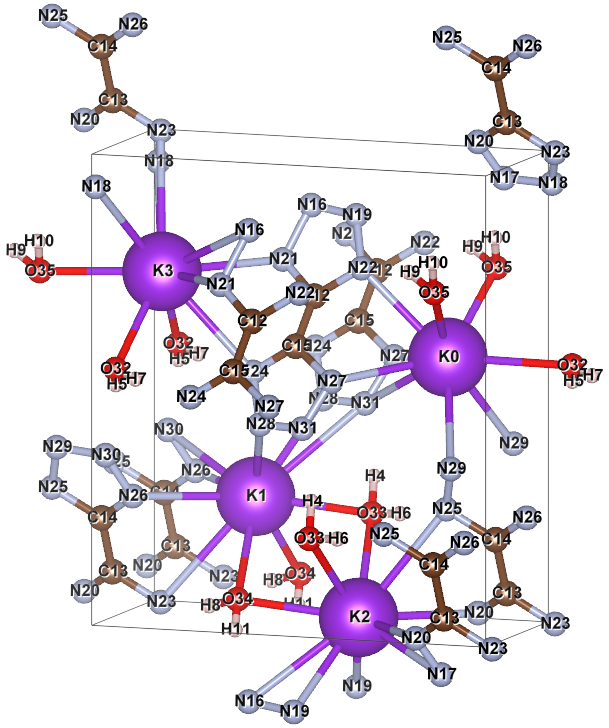}
        \caption{K4 H8 C4 N16 O4 (P1)}
    \end{subfigure}

    \caption{Generated metal-organic frameworks from \textsc{Zatom-1} trained on QMOF.}
    \label{appendix:figure:generated_mofs}
\end{figure}

\cref{appendix:figure:generated_crystals,appendix:figure:generated_molecules} show representative crystals and small molecules, respectively, sampled from \textsc{Zatom-1} pretrained jointly on the QM9 and MP20 datasets. Additionally, \cref{appendix:figure:generated_mofs} visualizes MOFs generated by \textsc{Zatom-1} pretrained on the QMOF dataset. Overall, \textsc{Zatom-1} generates compositionally diverse crystals across multiple spacegroups and chemically valid, structurally diverse molecules, indicating that it learns a rich generative latent space rather than a redundant set of representations.

\clearpage

\section{Additional Model Details}
\label{appendix:additional_model_details}

\begin{figure}[ht]
    \centering
    \begin{subfigure}[b]{0.48\textwidth}
        \includegraphics[width=\textwidth]{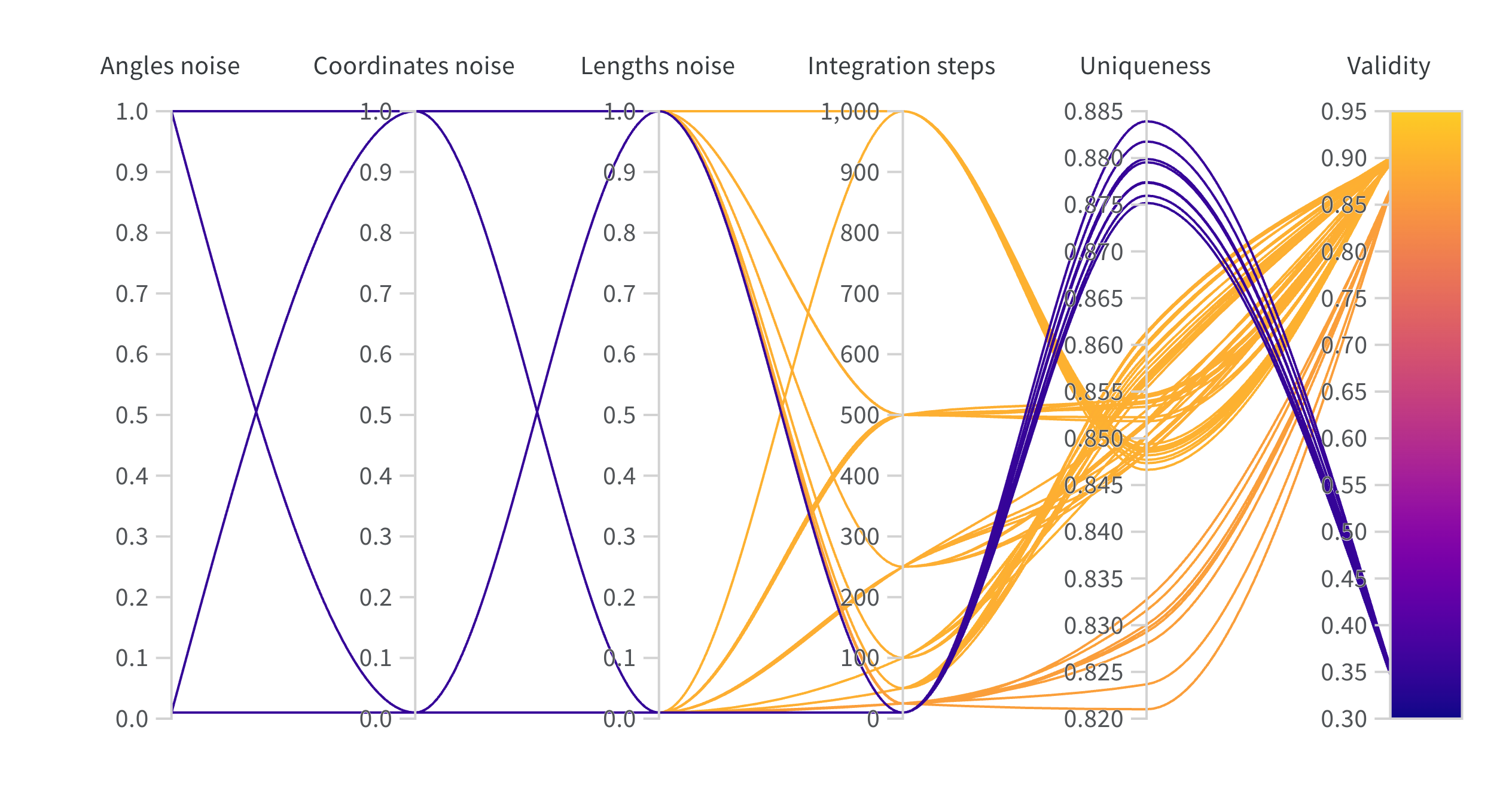}
        \caption{Crystals -- MP20}
        \label{appendix:figure:crystals_mp20_inference_sweep}
    \end{subfigure}
    \hfill
    \begin{subfigure}[b]{0.48\textwidth}
        \includegraphics[width=\textwidth]{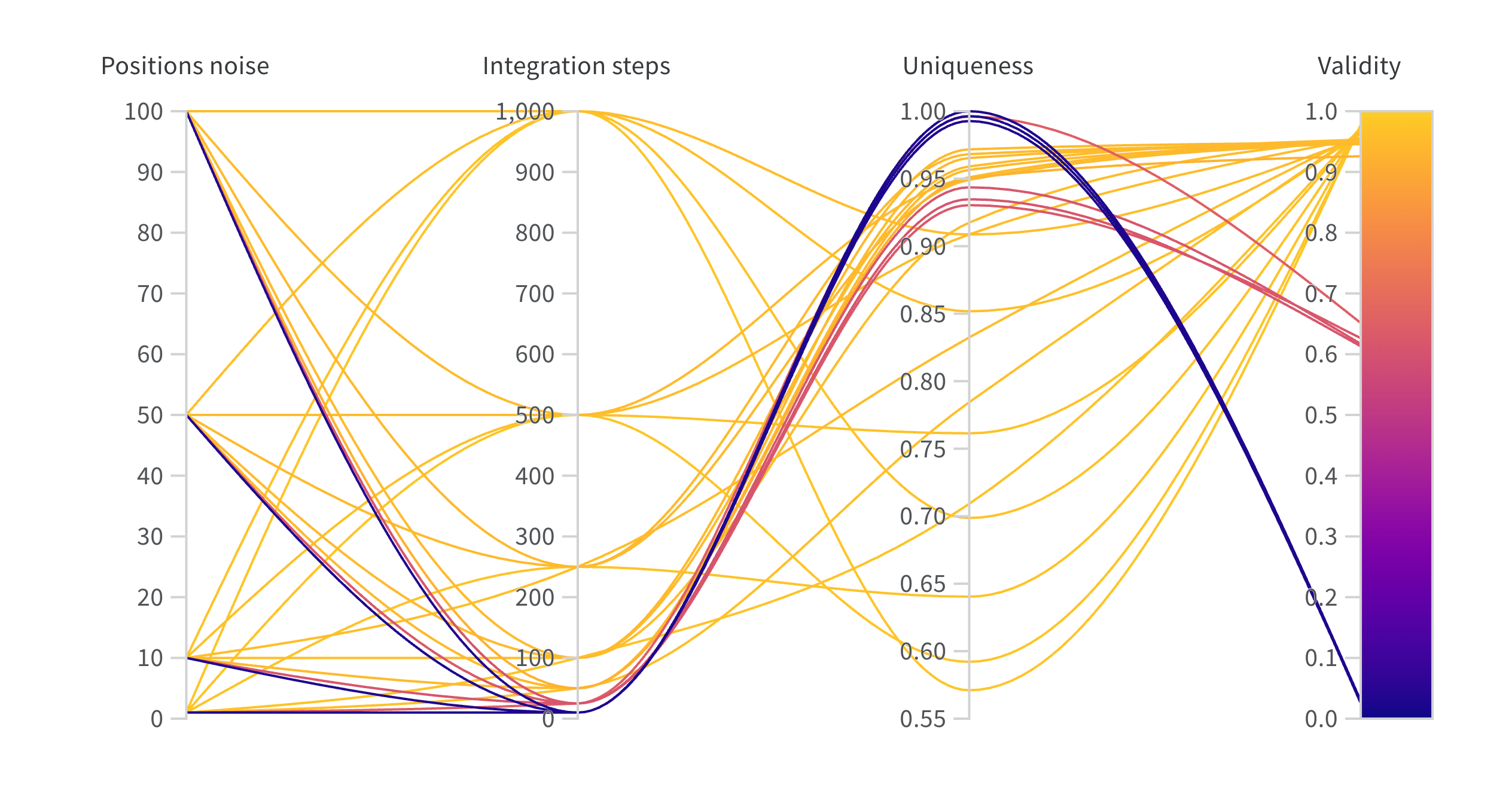}
        \caption{Molecules -- QM9}
        \label{appendix:figure:molecules_qm9_inference_sweep}
    \end{subfigure}
    \caption{
        \textbf{Tuning inference hyperparameters for best performance.} Best generative modeling results for crystals and molecules, balancing sample validity and uniqueness, are generally achieved with a limited number of integration steps (e.g., $T = 100$), a high white noise scale $\gamma_g$ (e.g., 50) for continuous, non-periodic atom positions, and a low noise scale (e.g., 0.01) for continuous, periodic modalities such as fractional coordinates.
    }
    \label{appendix:figure:crystals_molecules_mp20_qm9_inference_sweep}
\end{figure}

\paragraph{Hyperparameters.} \cref{appendix:table:mp20_qm9_pretraining_hyperparams} reports the hyperparameters used to pretrain different versions of \textsc{Zatom-1} on MP20 and QM9, and \cref{appendix:table:geom_qmof_omol25_pretraining_hyperparams} shows which hyperparameters were used to pretrain \textsc{Zatom-1} on QMOF or OMol25 as well as on GEOM-Drugs jointly with other datasets. Likewise, \cref{appendix:table:finetuning_hyperparams} displays the hyperparameters used to finetune \textsc{Zatom-1} (80M) for the downstream tasks explored in this work. For generative inference, \cref{appendix:figure:crystals_molecules_mp20_qm9_inference_sweep} shows the results of a hyperparameter sweep, where we find that using a limited number of integration steps and adding (varying levels of) white noise to each continuous modality during sampling generally improves sample validity and uniqueness. Notably, we find that for larger chemical systems, such as the molecules present in GEOM-Drugs, a smaller noise scale for atom positions (e.g., 0.01) is beneficial. Additionally, in all generative contexts, we unexpectedly find that classifier-free guidance worsens the quality of generated samples considerably (e.g., degrading 90\% materials validity rates to 40\%), motivating us to disable such inference guidance by default. We believe this may be addressed with alternative class label conditioning methods, which we defer to future work.

\paragraph{Compute resources.} All experiments used an internal SLURM cluster containing GPU nodes with 60 GBs of CPU memory and 80 GBs of (NVIDIA A100) GPU memory associated with each of the four GPUs attached to each node. In total, all experiments required $\sim$20,000 GPU hours for training and inference and 1 TB of local storage for datasets, model checkpoints, and experiment artifacts, with $\sim$3,000 GPU hours spent on preliminary experiments that did not make it into this manuscript.

\paragraph{Limitations.} We discuss our method's lack of (e.g., 100M) dataset size scaling results and preliminary precision for energy and force modeling in Section \ref{section:results} and Appendix \ref{appendix:extension_to_energies_and_forces}, respectively. Meaningful additional experiments in these two areas require a significantly larger compute budget than is available for this work. We also note that our method's molecular and material property prediction results show room for improvement compared to single-task learning baselines, which could be addressed with either dedicated single-task learning experiments for each molecular/material property (n.b., out of scope for this methodologically-focused work, due to compute constraints) or with more expressive variations of flow matching-based pretraining (n.b., which we aim to explore in future work). Importantly, our current computational experiments in Section \ref{section:results} and Appendix \ref{appendix:additional_results} already verify the effectiveness and computational efficiency of our proposed method in a fair (i.e., data, task, and compute-controlled) setting, with future wet-lab experiments in preparation.

\newpage
\paragraph{Broader Impacts} This paper presents work intended to advance the field of Scientific Machine Learning. There are many potential societal consequences of our work, including the acceleration of scientific discovery in the molecular and materials sciences. We acknowledge the risk that, in the hands of "bad actors", extensions of such technologies may be used with harmful ends in mind. However, with strong research community guidelines in place, such as those of \url{https://responsiblebiodesign.ai/}, it is our hope that efforts in developing scientific foundation models for the chemical sciences will disproportionately influence the positive societal outcomes of such research, such as improved medicines and energy technologies, as opposed to possible negative consequences, such as the development of new bioweapons.

\paragraph{Licensing.} \textsc{Zatom-1} is available under a permissive BSD license. The license of each publicly available dataset that \textsc{Zatom-1} uses for pretraining or finetuning is listed below.

\begin{enumerate}
    \item QM9, found at \url{https://pytorch-geometric.readthedocs.io/en/stable/generated/torch_geometric.datasets.QM9.html}, has a Creative Commons Attribution 4.0 (CC-BY-4.0) license.
    \item MP20, found at \url{https://huggingface.co/datasets/chaitjo/MP20_ADiT}, has an MIT license.
    \item GEOM-Drugs, found at \url{https://bits.csb.pitt.edu/files/geom_raw/}, has a Creative Commons Attribution 4.0 (CC-BY-4.0) license.
    \item QMOF, found at \url{https://huggingface.co/datasets/chaitjo/QMOF150_ADiT}, has an MIT license.
    \item Matbench, found at \url{https://huggingface.co/datasets/Ty-Perez/matbench_properties}, has an MIT license.
    \item OMol25, found at \url{https://huggingface.co/facebook/OMol25}, has a Creative Commons Attribution 4.0 (CC-BY-4.0) license.
    \item MPtrj, found at \url{https://huggingface.co/datasets/nimashoghi/mptrj}, has an MIT license.
\end{enumerate}

\vspace*{0.5cm}
\begin{table}[ht]
\centering
\caption{Hyperparameters across different model versions when jointly pretraining on QM9 \& MP20.}
\label{appendix:table:mp20_qm9_pretraining_hyperparams}
\resizebox{\textwidth}{!}{%
\begin{tabular}{lcccc}
\toprule
\textbf{Hyperparam.}    & \textbf{\textsc{Zatom-1-WD} (80M)}      & \textbf{\textsc{Zatom-1} (80M)} & \textbf{\textsc{Zatom-1-L} (160M)} & \textbf{\textsc{Zatom-1-XL} (300M)} \\ \midrule
\# Params.                & 77,350,264  & 77,350,264             & 162,615,544           & 293,887,096             \\
Hidden size               & 512 & 512                   & 768                  & 1024                    \\
\# Transformer blocks     & 16  & 16                    & 16                   & 16                     \\
\# Attn. heads            & 8   & 8                     & 8                    & 8                      \\
Train $t$ distribution    & Beta(1.8,1) & Beta(1.8,1)           & Beta(1.8,1)          & Beta(1.8,1)            \\
$\lambda_{\text{discrete}}$ & 0.1   & 0.1                   & 0.1                  & 0.1                    \\
Learning rate             & 0.001   & 0.001                 & 0.0005                & 0.00025                 \\
Learning rate schedule             & N/A    & N/A                 & N/A                & N/A                 \\
Optimizer                 & AdamW   & AdamW                  & AdamW                 & AdamW                   \\
Weight decay                 & 0.01  & 0.0                  & 0.0                 & 0.0                   \\
EMA weight                & 0.999   & 0.999                 & 0.999                & 0.999                  \\ \midrule
Batch size                & 64 & 256                   & 128                  & 64                    \\
\# Grad. accum. steps                & 1    & 1                   & 2                  & 4                    \\
\# Rotation augs.         & 8   & 8                     & 8                    & 8                      \\
Effective batch size      & 2,048  & 32,768                  & 32,768                 & 32,768                   \\ \midrule
\# GPUs                   & 4 (80GB)   & 16 (80GB)                     & 16 (80GB)                    & 16 (80GB)                      \\
Training duration         & 10h & 24h                   & 48h                  & 96h                    \\ \midrule
\# Sampling steps         & 100 & 100                   & 100                  & 100                    \\
$g(t)$                    & $\frac{1}{t+0.01}$  & $\frac{1}{t+0.01}$    & $\frac{1}{t+0.01}$   & $\frac{1}{t+0.01}$     \\
$\gamma_g$ ($\rightarrow$ for small molecule coordinates)                  & 0.01 ($\rightarrow$ 50.0)  & 0.01 ($\rightarrow$ 50.0)                  & 0.01 ($\rightarrow$ 50.0)                 & 0.01 ($\rightarrow$ 50.0)                   \\ \bottomrule
\end{tabular}
}%
\end{table}

\begin{table}[ht]
\centering
\caption{Hyperparameters when jointly pretraining \textsc{Zatom-1} on GEOM-Drugs, QMOF, or OMol25.}
\label{appendix:table:geom_qmof_omol25_pretraining_hyperparams}
\resizebox{0.6\textwidth}{!}{%
\begin{tabular}{lccc}
\toprule
\textbf{Hyperparam.}      & \textbf{GEOM-Drugs \& MP20} & \textbf{QMOF} & \textbf{OMol25}    \\ \midrule
\# Params.                & 77,350,264            & 77,350,264            & 77,350,264              \\
Hidden size               & 512                   & 512                   & 512                    \\
\# Transformer blocks     & 16                    & 16                    & 16                     \\
\# Attn. heads            & 8                     & 8                     & 8                      \\
Train $t$ distribution    & Beta(1.8,1)           & Beta(1.8,1)           & Beta(1.8,1)            \\
$\lambda_{\text{discrete}}$ & 0.1                 & 0.1                 & 0.1                    \\
Learning rate             & 0.001                 & 0.001                 & 0.001                  \\
Learning rate schedule    & N/A                   & N/A                   & N/A                    \\
Optimizer                 & AdamW                 & AdamW                 & AdamW                  \\
Weight decay              & 0.0                   & 0.0                   & 0.0                    \\
EMA weight                & 0.999                 & 0.999                 & 0.999                  \\ \midrule
Batch size                & 32                    & 32                    & 16                     \\
\# Grad. accum. steps     & 1                     & 1                     & 1                      \\
\# Rotation augs.         & 8                     & 8                     & 8                      \\
Effective batch size      & 4,096                 & 4,096                 & 2,048                  \\ \midrule
\# GPUs                   & 16 (80GB)             & 16 (80GB)             & 16 (80GB)               \\
Training duration         & 168h                  & 672h                  & 168h                   \\ \midrule
\# Sampling steps         & 100                   & 100                   & 100                    \\
$g(t)$                    & $\frac{1}{t+0.01}$    & $\frac{1}{t+0.01}$    & $\frac{1}{t+0.01}$     \\
$\gamma_g$                & 0.01                  & 0.01                  & 0.01                   \\ \bottomrule
\end{tabular}
}%
\end{table}

\begin{table}[ht]
\centering
\caption{Finetuning hyperparameters across different downstream prediction tasks for \textsc{Zatom-1} (80M), optionally trained in sequential order.}
\label{appendix:table:finetuning_hyperparams}
\resizebox{\textwidth}{!}{%
\begin{tabular}{lccc}
\toprule
\textbf{Hyperparam.}      & \textbf{Mol. Properties (80M \faSnowflake\ + 20M \faFire) $\rightarrow$} & \textbf{Mol. / Mat. Properties (80M \faSnowflake\ + 20M \faFire) $\rightarrow$} & \textbf{Mol. / Mat. Energy \& Forces (80M \faSnowflake\ + 220M \faFire)} \\ \midrule
\# Dedicated params.                & 18,104,595             & 18,104,595           & 217,255,140             \\
Aux. hidden size               & 512                   & 512                  & 1024                    \\
\# Aux. transformer blocks     & 4                    & 4                   & 8                     \\
\# Attn. heads            & 8                     & 8                    & 8                      \\
Train $t$ distribution    & max(Beta(1.8,1), 0.98)           & max(Beta(1.8,1), 0.98)          & max(Beta(1.8,1), 1.0)            \\
$\lambda_{\text{forces}}$ & N/A                   & N/A                  & 5.0                    \\
Learning rate             & 0.001                 & 0.001                & 0.0003                 \\
Learning rate schedule    & N/A                   & N/A                  & N/A                 \\
Optimizer                 & AdamW                  & AdamW                 & AdamW                   \\
Weight decay              & 0.0                  & 0.0                 & 0.0                   \\
EMA weight                & N/A                 & N/A                & N/A                  \\ \midrule
Batch size                & 128                   & 32                  & 72                    \\
\# Grad. accum. steps                & 2                   & 4                  & 1                    \\
\# Rotation augs.         & 8                     & 8                    & 1                      \\
Effective batch size      & 4,096                  & 2,048                 & 1,152                   \\ \midrule
\# GPUs                   & 2 (40GB)                     & 2 (80GB)                    & 16 (80GB)                      \\
Training duration         & 48h                   & 192h                  & 120h                    \\ \bottomrule
\end{tabular}
}%
\end{table}

\clearpage

\section{\ \textsc{Platom-1} \ -- \ Explicitly Equivariant Transformer Architecture}
\label{appendix:tfp_main}

A long-standing discussion in the molecular modeling community revolves around whether symmetries in the data should be learned or hard-coded into the model architecture.
Given the considerable success of entirely non-equivariant models like AlphaFold 3 \citep{abramson2024accurate}, the community's consensus has shifted toward the conclusion that equivariant models would be slower, harder to optimize, and generally inferior compared to architectures without such inductive biases. This appendix provides empirical evidence challenging this common belief, showing that -- \emph{when done right} -- equivariance can unlock significant performance gains.

Section~\ref{appendix:tfp_model} introduces the Trunk-based Flow Platoformer (TFP),
a rotation equivariant counterpart of the Trunk-based Flow Transformer (TFT) architecture underlying \textsc{Zatom-1}. Empirical results, showing faster convergence and higher validation scores of TFP, are presented in Section~\ref{appendix:tfp_results} and Figure~\ref{appendix:figure:tfp_qm9_equivariance_experiment}. For now, we focus exclusively on molecular tasks, since the current fractional coordinate-based invariant embedding of materials used in \textsc{Zatom-1} is at odds with the equivariant inputs expected by TFP.

\subsection{Trunk-based Flow Platoformer architecture (TFP)}
\label{appendix:tfp_model}

The Trunk-based Flow Platoformer (TFP) is a rotation equivariant counterpart of our TFT architecture from Section~\ref{subsection:network_architecture},
replacing its standard transformer layers with \emph{Platonic Transformer} layers introduced by \citet{islam2025platonic}. These layers are equivariant to discrete subgroups of roto-reflections $G<\mathrm{O}(3)$ which are symmetries of Platonic solids like the tetrahedron ($|G|=12$) or cube ($|G|=24$) \citep{cesa2021ENsteerable}.

Rather than relying on $\mathrm{O}(3)$-irrep features (i.e., $\mathrm{O}(3)$-irreducible representations), as most equivariant point cloud message passing neural networks do, the features of TFP are associated with group elements, essentially encoding molecular geometry as viewed from $|G|$ different frames of reference. Mathematically, this is known as a \emph{regular $G$-representation} \citep{weiler2025EquivCoordIndepCNNsWS}. In code, a point cloud of $N$ features with $C$ channels is represented by a tensor of shape $(N,|G|,C)$ rather than the usual shape $(N,C)$. Such regular representation-valued features are more commonly used in vision models, most notably group convolutional networks \citep{Cohen2016-GCNN,Weiler2018SFCNN,bekkers2018roto}.

\paragraph{TFP architecture.}
Before applying Platonic layers, inputs need to be lifted to regular $G$-representations.
Scalars (e.g., atom types) are lifted to features that are invariant over the group axis.
Vector-valued inputs, like Euclidean coordinates, lift to non-trivial features encoding directional information. Platonic linear layers satisfy a symmetry constraint
which reduces their parameter count via $|G|$-fold weight sharing. The original linear embedding layers of TFT are replaced by such Platonic linear layers, applied after $G$-lifts.

Platonic attention layers utilize common scaled dot-product attention \citep{vaswani2017attention}, and are thus compatible with IO-efficient Flash Attention kernels \citep{dao2024flashattention}. The group axis of regular $G$-features is thereby reshaped into the attention heads axis rather than channels, i.e., ${(H\!\cdot\!|G|,\, N,\, C)}$ instead of ${(H,\, N,\, |G|\!\cdot\!C)}$ -- this has the advantage that attention scores remain regular representation-valued rather than becoming $G$-invariant due to contracting over $G$ \citep{islam2025platonic}. We use full softmax attention instead of the linear attention variants explored in the original Platonic Transformer paper, as these were required mainly to address training instabilities when encoding rotary positions in molecular modeling tasks.

A core component of the Platonic Transformer is a $G$-equivariant generalization of Rotary Position Embeddings (RoPE) \citep{su2024roformer}, which guarantees both translational and $G$-equivariance, i.e., approximate $\mathrm{E}(3)$-equivariance. As we aim to stick as close as possible to the original TFT design without \emph{relative} RoPE attention, we omit this feature, relying merely on the \emph{absolute} embedding of Euclidean coordinates via lifting and Platonic linear layers discussed above. Note that this design choice breaks translational symmetries but preserves $G$-equivariance. Since we zero-center non-periodic 3D atom coordinates, this is not an issue in practice.

We reimplement and extend many other operations from the original Platonic Transformer code, including cross-attention layers, new normalization modules, embedding layers, and optimized Platonic linear layers. Given these modules, we implement TFP in analogy to the TFT design summarized in Algorithm~\ref{algorithm:tft_model_pseudocode}. Final predictions are projected back from $G$-valued features to scalar and vector outputs. Our implementation is available as part of the \textsc{Zatom-1} code repository.

\paragraph{Compute and parameter cost.}
As TFP operates on tensors of shape $(N,|G|,C)$ with an additional group axis, one may wonder how its computational, memory, and parameter complexities compare to those of the TFT baseline. Given that regular features have $|G|C$ effective dimensions, it is common practice to reduce their channel count $C$ relative to the~$D$ channels of TFT's shape $(N,D)$ features. The implied complexities for TFT's and TFP's linear and attention layers are summarized in Table~\ref{appendix:table:TFP_complexity}. From these complexities, the following two heuristics to choose $C$ to build ``matching'' architectures arise \citep{Weiler2019_E2CNN}:

\begin{itemize}[leftmargin=1.cm, align=left]
\setlength{\itemindent}{-.1cm}
\item[\emph{Compute-matching}:]
    To match the compute and memory cost of TFP and TFT, one sets $C = D/|G|$,
    essentially sticking with the same effective number of channels but splitting them into group and channel axes. However, this implies $G$ times fewer trainable parameters in TFP.
\item[\emph{Parameter-matching}:]
    Alternatively, one may match the number of parameters of TFP and TFT by setting $C = D/\sqrt{|G|}$. This leads to a $|G|$ fold increase in memory and Floating-point Operations Per Second (FLOPS).
\end{itemize}

As a middle ground between these two extremes, we choose $C = D/|G|^{2/3}$,
resulting in $\sqrt{|G|}$ fewer parameters and $\sqrt{|G|}$ increased computational cost.
For the tetrahedral group, $|G|=12$ and $\sqrt{|G|}\approx 3.5$, such that a TFP analog to \textsc{Zatom-1} with 80M parameters has 23M parameters. Note that this model happens to be approximately compute-matched with \textsc{Zatom-1-XL}, which has 300M parameters, i.e., $3.75$ times the number of \textsc{Zatom-1}'s parameters.

\begin{table}[t]
    \vspace*{1ex}
    \centering
    \small
    \renewcommand{\arraystretch}{1.45}
    \setlength{\tabcolsep}{6pt}
    \begin{tabular}{r@{\hskip40pt}c c@{\hskip40pt}c c}
        \toprule
                      & Linear              & $G$-linear               & Attention           & $G$-attention \\
        \midrule
        Compute FLOPS & $\mathcal{O}(ND^2)$ & $\mathcal{O}(N|G|^2C^2)$ & $\mathcal{O}(N^2D)$ & $\mathcal{O}(N^2|G|C)$ \\
        Memory        & $\mathcal{O}(ND)$   & $\mathcal{O}(N|G|C)$     & $\mathcal{O}(ND)$   & $\mathcal{O}(N|G|C)$ \\
        Parameters    & $D^2$               & $|G| \!\cdot\! C^2$      & ---                 & --- \\
        \bottomrule
    \end{tabular}
    \vspace*{2ex}
    \caption{
        \textbf{Compute, memory, and parameter complexity of TFT and TFP.}
        TFT and TFP operate on tensors of shapes $(N,|G|,C)$ and $(N,D)$, respectively. Memory costs follow immediately from these shapes. Linear layers in TFT come with the usual quadratic number of FLOPS and parameters in their $D$ channels. TFP is internally leveraging \texttt{nn.Linear} layers as well, however, after flattening features to shape $(N,|G|\!\cdot\!C)$, implying a computational cost quadratic in $|G|$ as well.
        Equivariance imposes a symmetry constraint on the matrices (weight sharing), resulting in a $|G|$ times-enhanced parameter efficiency. TFT's attention layers have the usual computational complexity linear in channels $D$, while those of TFP are linear in $|G|C$. Neither introduces additional parameters.
    }
    \label{appendix:table:TFP_complexity}
    \vspace{0.5em}
\end{table}

\subsection{Empirical results}
\label{appendix:tfp_results}

Based on TFP, we implement \textsc{Platom-1}, the Platonic group equivariant counterpart of \textsc{Zatom-1}. Due to limited computational resources, we specifically evaluate the (orientation-preserving) tetrahedral group with $|G|=12$ rotations (no reflections) and 23M parameters as discussed above. Since \textsc{Platom-1} is only equivariant w.r.t. discrete rotations, we use the same continuous $\mathrm{SO}(3)$ rotational data augmentation as used for \textsc{Zatom-1}.

\begin{figure}[!h]
    \vspace*{4ex}
    \centering
    \begin{subfigure}[b]{.44\textwidth}
        \includegraphics[width=\textwidth]{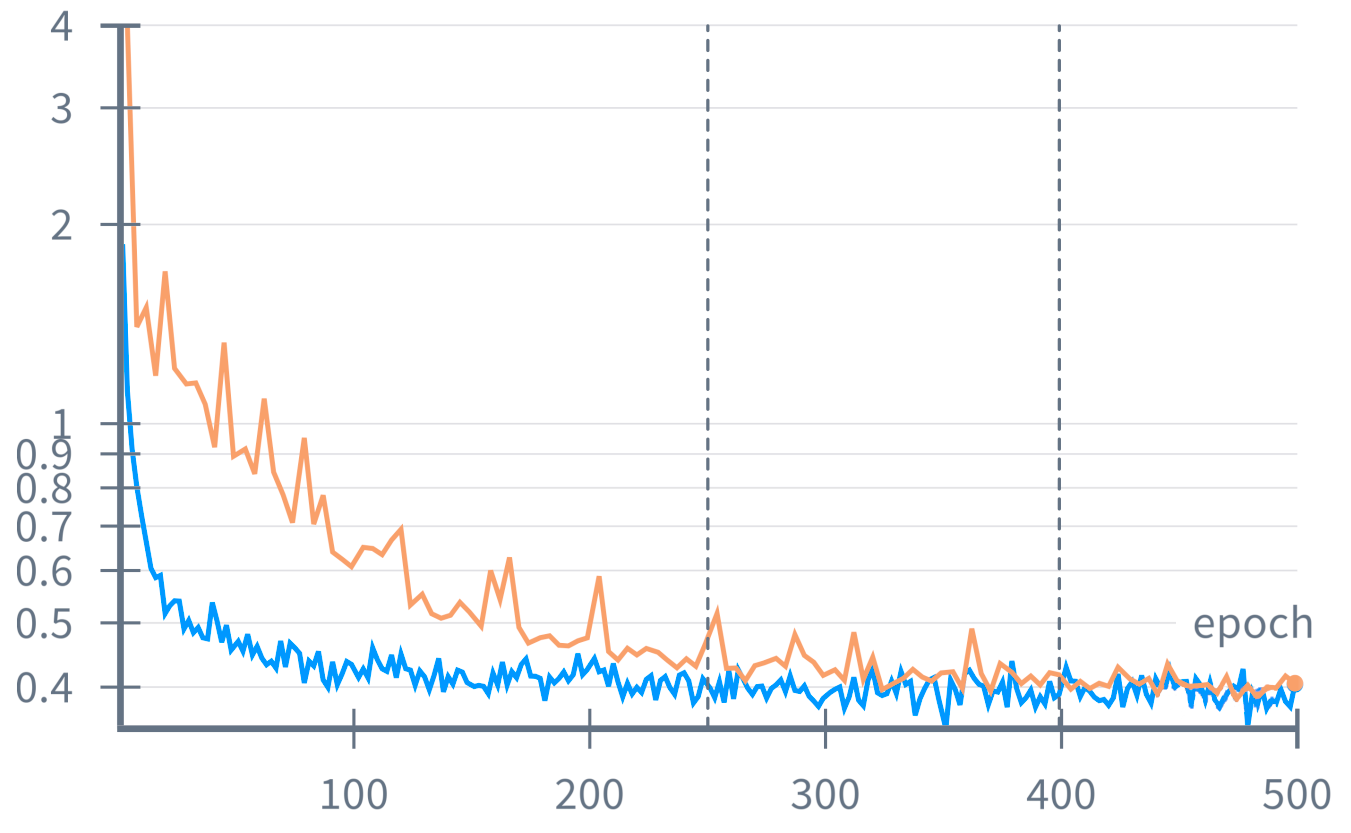}
        \vspace*{-3.25ex}
        \caption{Training loss}
        \label{appendix:figure:tfp_qm9_loss_train}
    \end{subfigure}
    \hspace{4ex}
    \begin{subfigure}[b]{.44\textwidth}
        \centering
        \includegraphics[width=.85\textwidth]{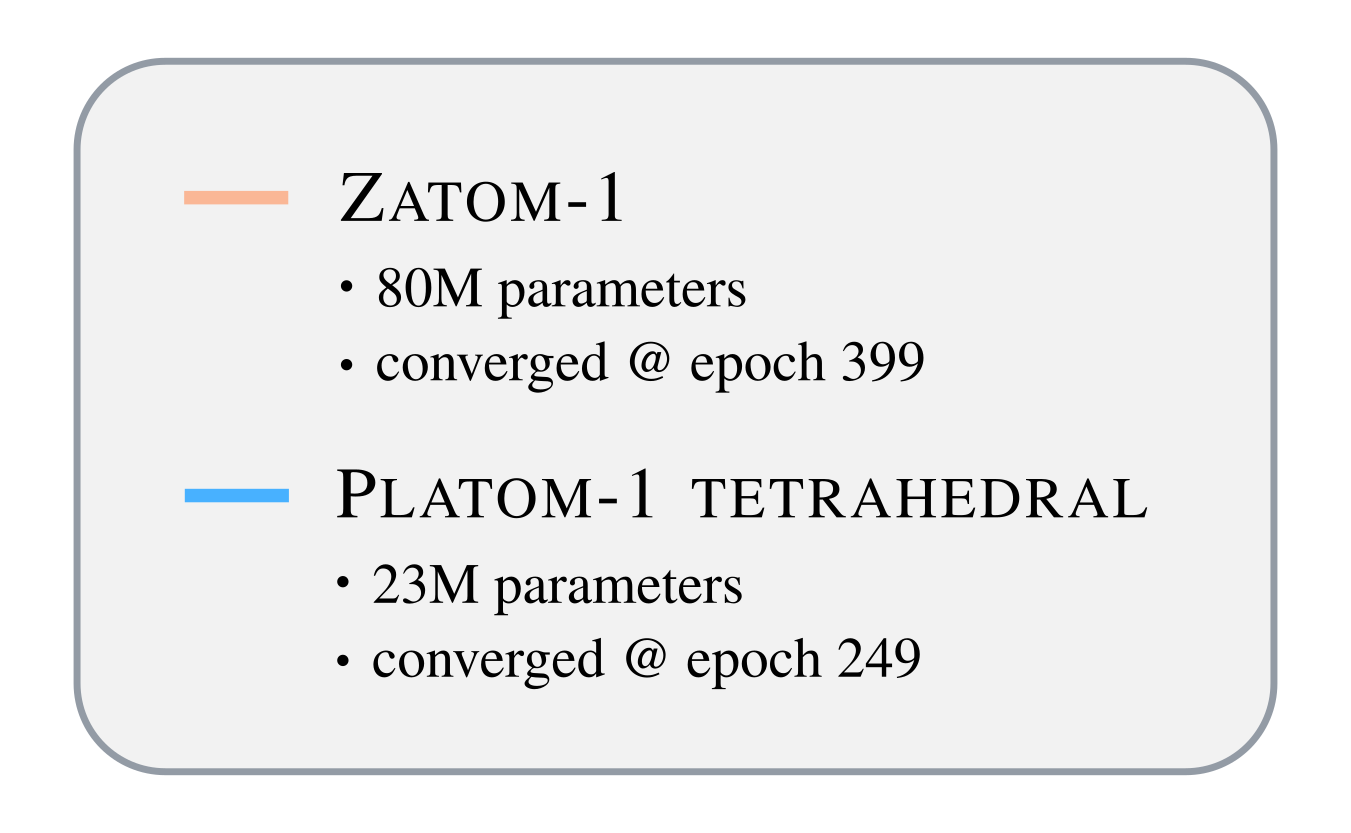}
        \vspace*{7ex}
    \end{subfigure}
    \vspace*{3.5ex}

    \begin{subfigure}[b]{.44\textwidth}
        \includegraphics[width=\textwidth]{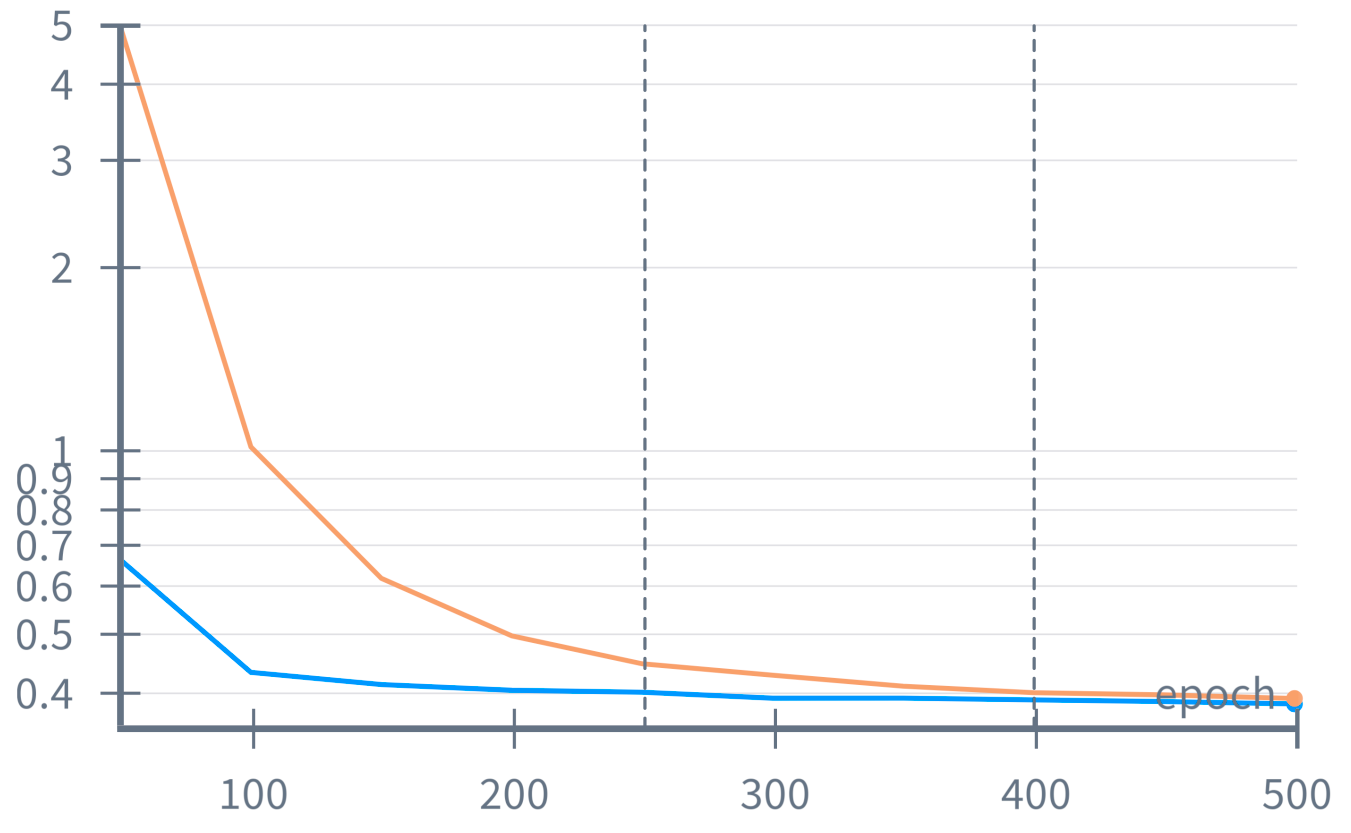}
        \vspace*{-3.25ex}
        \caption{Validation loss}
        \label{appendix:figure:tfp_qm9_loss_val}
    \end{subfigure}
    \hspace{4ex}
    \begin{subfigure}[b]{.44\textwidth}
        \includegraphics[width=\textwidth]{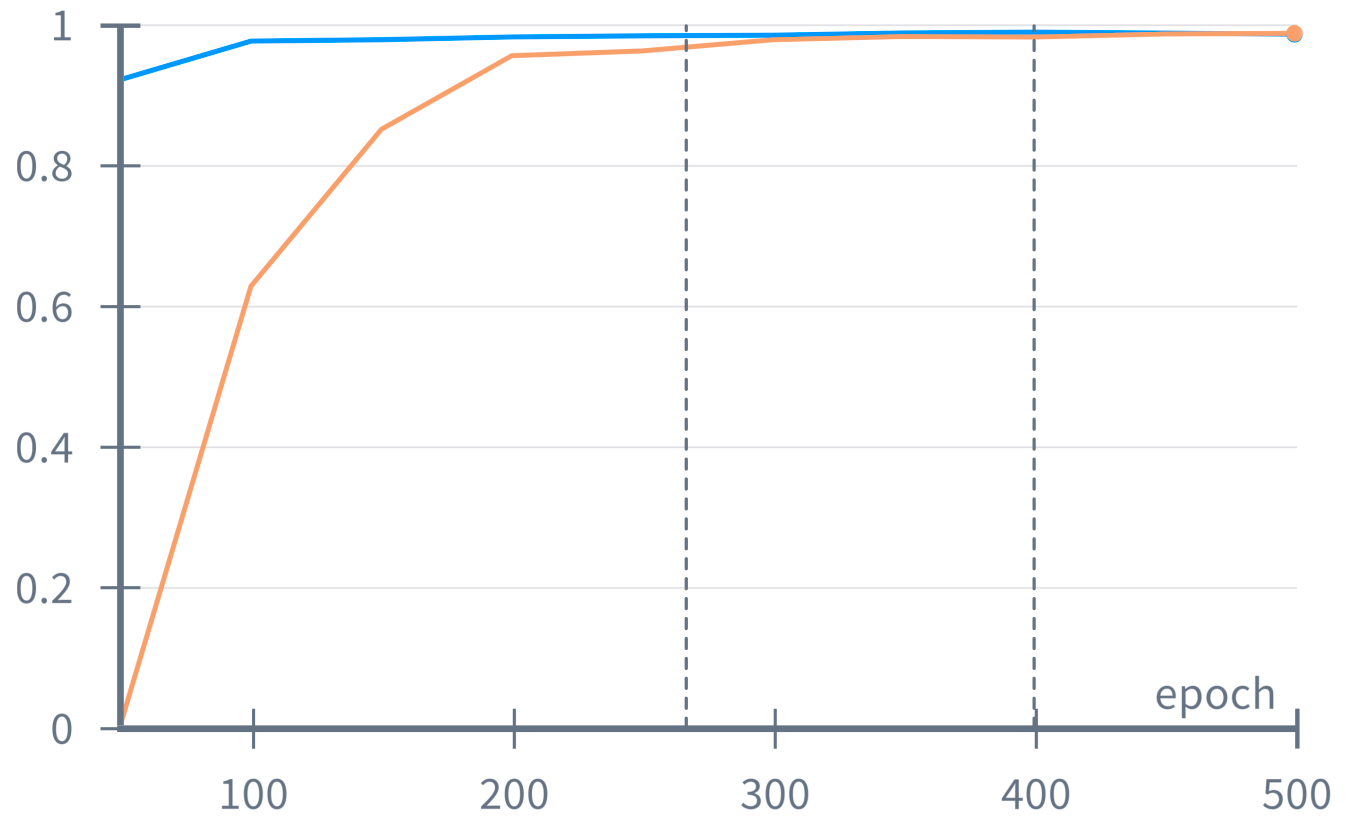}
        \vspace*{-3.25ex}
        \caption{Posebusters rate}
        \label{appendix:figure:tfp_qm9_loss_posebusters}
    \end{subfigure}
    \vspace*{3.5ex}

    \begin{subfigure}[b]{.44\textwidth}
        \includegraphics[width=\textwidth]{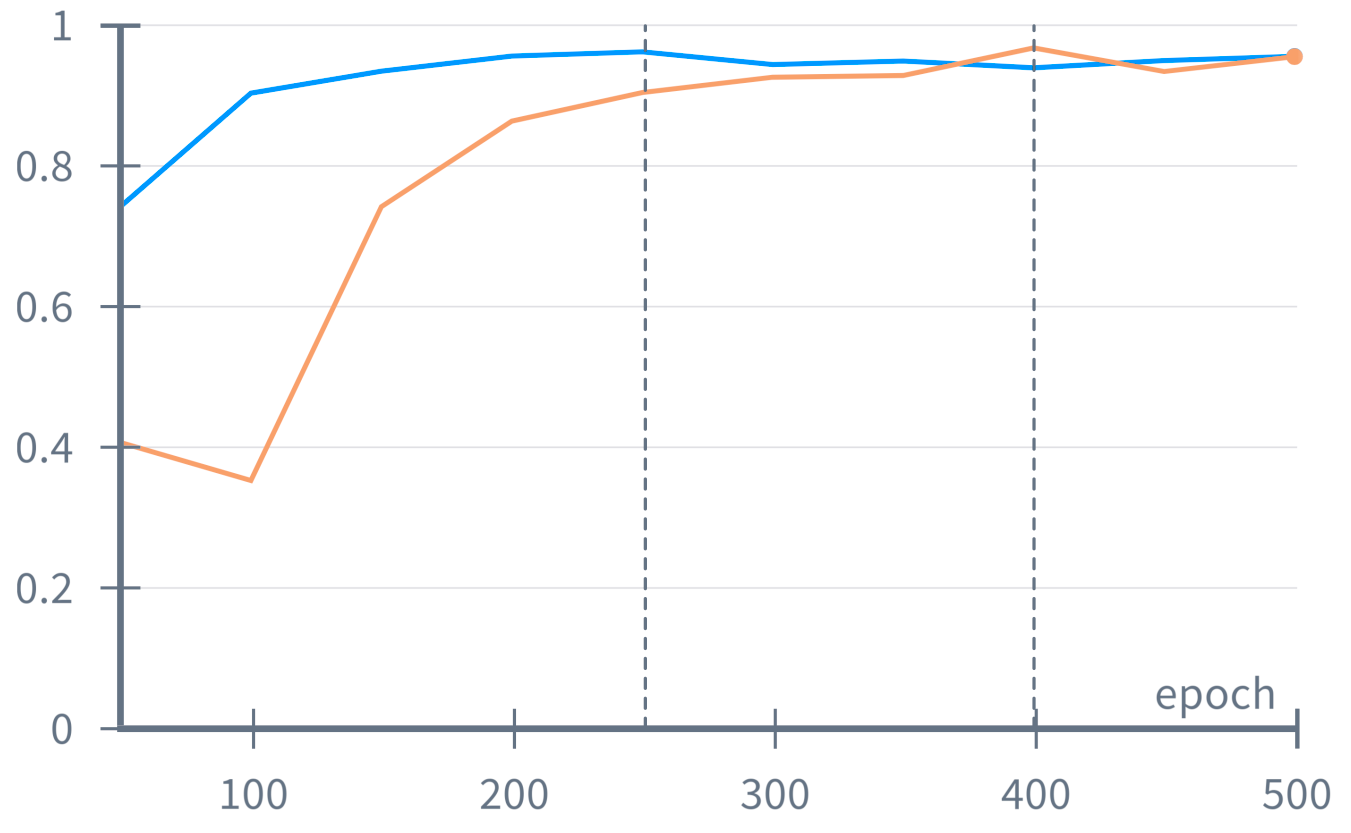}
        \vspace*{-3.25ex}
        \caption{Valid rate}
        \label{appendix:figure:tfp_qm9_loss_valid_rate}
    \end{subfigure}
    \hspace{4ex}
    \begin{subfigure}[b]{.44\textwidth}
        \includegraphics[width=\textwidth]{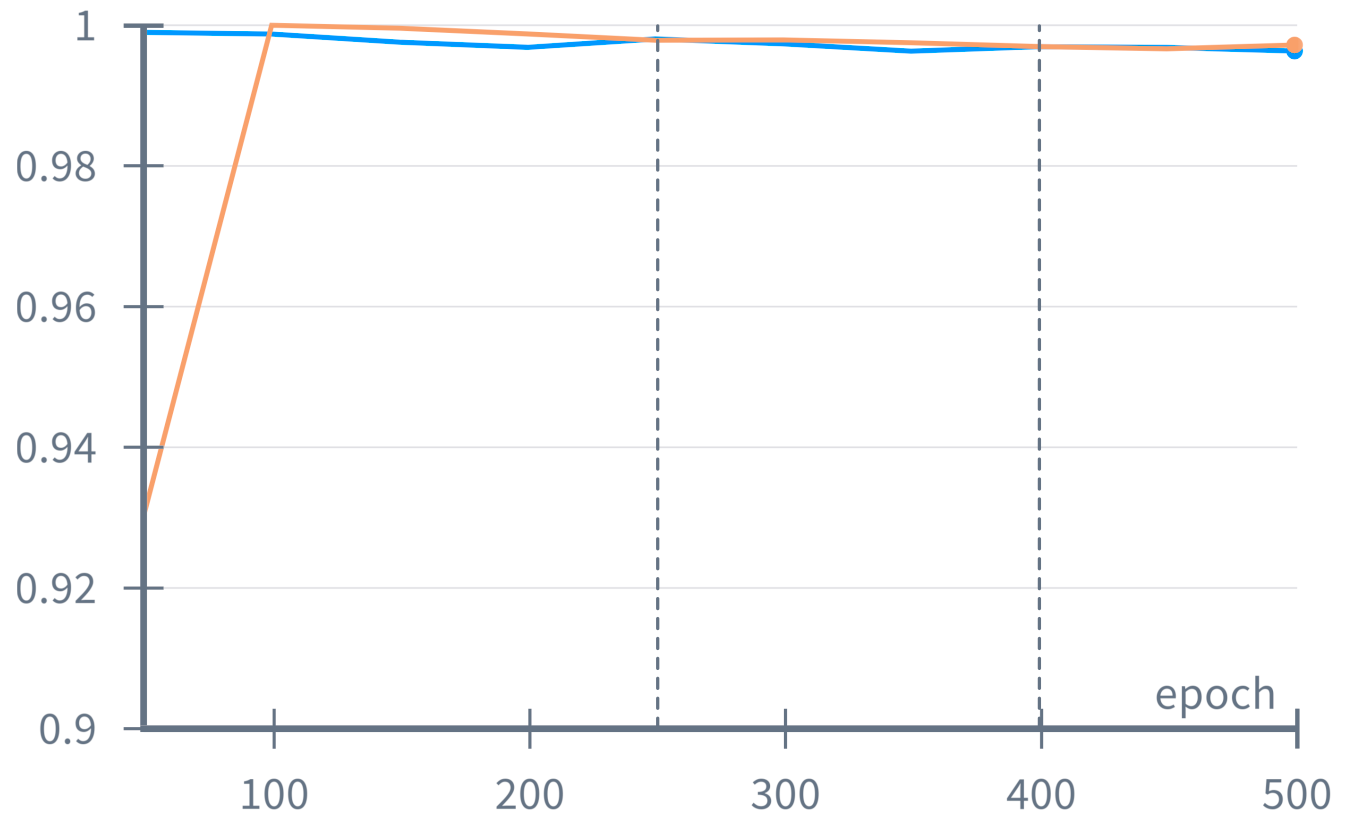}
        \vspace*{-3.25ex}
        \caption{Unique rate}
        \label{appendix:figure:tfp_qm9_loss_unique_rate}
    \end{subfigure}

    \vspace*{2ex}
    \caption{
        \textbf{QM9-only pretraining of \textsc{Zatom-1} vs. \textsc{Platom-1}.}
        These plots show the training dynamics of \textsc{Zatom-1} compared to the analogous tetrahedral group equivariant \textsc{Platom-1} model described in Section~\ref{appendix:tfp_model}.
        Due to its equivariance, the Platonic model does not explicitly need to learn symmetries, resulting in significantly faster convergence while achieving improved evaluation metrics.
        These findings translate to the test results in Table~\ref{appendix:table:qm9_dng_equivariance}.
    }
    \label{appendix:figure:tfp_qm9_equivariance_experiment}
\end{figure}

\paragraph{QM9 molecule generation.}
We first compare to the QM9-only pretrained \textsc{Zatom-1} baseline flow model in Figure~\ref{appendix:figure:tfp_qm9_equivariance_experiment}. It is evident from these plots that \textsc{Platom-1} converges much faster than \textsc{Zatom-1} in terms of training and validation losses, as well as validation metrics. Such improved convergence is common for equivariant models, as they do not need to implicitly learn symmetries from data \citep{weiler2025EquivCoordIndepCNNsWS,zhdanov2024CS-CNNs}. The best TFP checkpoint on the validation set is reached after 249 epochs, instead of 399 epochs for TFT, partially offsetting TFP's increased computational cost.

Table~\ref{appendix:table:qm9_dng_equivariance} confirms that these findings translate to all test metrics calculated for these checkpoints. In particular, \textsc{Platom-1} achieves better results than \textsc{Zatom-1}, prior equivariant QM9-only pretrained models Equivariant Diffusion \citep{hoogeboom2022equivariant} and Symphony \citep{daigavane2024symphony}, and the much larger and slower latent diffusion model ADiT \citep{joshi2025allatom}. It even outperforms our 300M parameter model \textsc{Zatom-1-XL},
despite the latter being jointly trained on molecules and materials.

\paragraph{Joint molecule and material pretraining.}

One of the main hypotheses of \textsc{Zatom-1} is the benefit of positive transfer learning when pretraining jointly on molecular and material modalities. A similar mutual benefit should, in principle, be observable for \textsc{Platom-1}. However, our current encoding of materials is inherently incompatible with \text{Platom-1}'s equivariant design, leading to training instabilities. Specifically, materials are not represented in terms of their (vector-valued) Euclidean coordinates, but rather in terms of fractional coordinates and lattice basis lengths and angles. All these quantities are scalars, and therefore lift to features that are invariant over the $G$-axis. Being blind to the original Euclidean geometry expected by \textsc{Platom-1}, the model fails to converge.

While \textsc{Zatom-1} can learn to leverage fractional coordinates inputs, we believe that this is not the ideal (equivariant) data representation for \textsc{Platom-1}, since it obfuscates the Euclidean geometry of the materials. As part of future work, we plan to experiment with TFP using a unified Euclidean coordinate representation of molecules and materials, for which we expect to see positive transfer learning results in a \textsc{Platom-1}-style model.

\begin{table}[!ht]
\vspace*{2ex}
\centering
\caption{
    \textbf{Molecule generation results on QM9 with explicit equivariance.}
    We report (a) validity and uniqueness rates as well as (b) \% pass rates on 7 sanity checks from PoseBusters for 10,000 sampled molecules.
    All models explicitly generate hydrogen atoms.
    Notably, QM9-only \textsc{Platom-1} achieves its results using 249 training epochs compared to QM9-only \textsc{Zatom-1} (80M), which requires 399 training epochs to converge.
    Despite its smaller size and not being jointly pretrained on molecules and materials, \textsc{Platom-1} is competitive with ADiT and \textsc{Zatom-1-XL}.
}
\label{appendix:table:qm9_dng_equivariance}

\begin{minipage}[t]{0.56\textwidth}
    \centering
    (a) \footnotesize \textbf{Validity results}
    \\[1ex]
    \resizebox{\linewidth}{!}{%
    \begin{tabular}{lccc}
    \toprule
    \textbf{Model} & \textbf{\# Params} & \textbf{Validity (\%) $\uparrow$} & \textbf{Unique (\%) $\uparrow$} \\
    \midrule
    \multicolumn{3}{l}{\textbf{One-stage training}} \\
    Equivariant Diffusion & 20M & 91.90 & \underline{98.69} \\
    Symphony & 2M & 83.50 & 97.98 \\
    \midrule
    \multicolumn{3}{l}{\textbf{Two-stage training}} \\
    GeoLDM & 20M & 93.80 & \textbf{98.82} \\
    QM9-only ADiT & 180M & 92.19 & 97.90 \\
    Jointly trained ADiT & 180M & 94.45 & 97.82 \\
    \midrule
    \multicolumn{3}{l}{\textbf{Ours: One-stage training}} \\
    QM9-only \textsc{Zatom-1}  & 80M & 92.88 & 97.71 \\
    \rowcolor{orange!30}
    QM9-only \textsc{Platom-1} & 23M & \underline{95.20} & 98.13 \\
    \midrule
    Jointly trained \textsc{Zatom-1} & 80M & 94.94 & 97.16 \\
    Jointly trained \textsc{Zatom-1-L} & 160M & \textbf{95.26} & 96.84 \\
    Jointly trained \textsc{Zatom-1-XL} & 300M & 95.19 & 97.10 \\
    \bottomrule
    \end{tabular}%
    }
\end{minipage}
\\[16pt]
\begin{minipage}[t]{0.85\textwidth}
    \centering
    (b) \footnotesize \textbf{PoseBusters results}
    \\[1ex]
    \resizebox{\linewidth}{!}{%
    \begin{tabular}{lcccccc}
    \toprule
    \makecell{\textbf{Test (\% pass) $\uparrow$} \\ ~} &
    \makecell{\textbf{Symphony}          \\[1pt] \footnotesize (QM9-only)} &
    \makecell{\textbf{Eq. Diff.}         \\[1pt] \footnotesize (QM9-only)} &
    \makecell{\textbf{ADiT}              \\[1pt] \footnotesize (jointly trained)} &
    \makecell{\textbf{\textsc{Platom-1}} \\[1pt] \footnotesize (QM9-only)} &
    \makecell{\textbf{\textsc{Zatom-1}}  \\[1pt] \footnotesize (QM9-only)} &
    \makecell{\textbf{\textsc{Zatom-1}}  \\[1pt] \footnotesize (jointly trained)} \\
    \midrule
    Atoms connected    &  99.92 &  99.88 &  99.70 &  \underline{99.97} &  99.95 &  \textbf{99.98} \\
    Bond angles        &  99.56 &  \textbf{99.98} &  99.85 &  99.94 &  99.90 &  \underline{99.95} \\
    Bond lengths       &  98.72 & \textbf{100.00} &  99.41 &  99.96 &  99.89 &  \underline{99.97} \\
    Aromatic ring flat & \textbf{100.00} & \textbf{100.00} & \textbf{100.00} & \textbf{100.00} & \textbf{100.00} & \textbf{100.00} \\
    Double bond flat   &  99.07 &  98.58 &  99.98 & \textbf{100.00} & \textbf{100.00} &  \underline{99.99} \\
    Internal energy    &  95.65 &  94.88 &  95.86 &  \underline{99.73} &  99.63 &  \textbf{99.78} \\
    No steric clash    &  98.16 &  99.79 & 99.79 &  \textbf{99.84} &  \underline{99.81} &  \underline{99.81} \\
    \bottomrule
    \end{tabular}%
    }
\end{minipage}
\end{table}

\clearpage

\section{Additional Results}
\label{appendix:additional_results}


\begin{table}[t]
\centering
\caption{
    \textbf{Metal-organic framework generation results on QMOF.} We report sanity checks from MOFChecker \citep{jin2025mofchecker} for 1,000 sampled MOFs, comparing methods trained only on QMOF for at least 10,000 epochs. ($\uparrow$/$\downarrow$ indicate higher/lower is better, respectively, and * denotes a model that has not converged after 50,000 training epochs)
}
\label{appendix:table:mof_generation_results}
\resizebox{0.5\textwidth}{!}{%
\begin{tabular}{l|rr}
\toprule
\textbf{Test\%} & \textbf{QMOF ADiT} & \textbf{QMOF \textsc{Zatom-1}}* \\
\midrule
Has carbon $\uparrow$ & 100.0 & 100.0 \\
Has hydrogen $\uparrow$ & 99.6 & 100.0 \\
Has atomic overlap $\downarrow$ & 8.3 & 8.8 \\
Has overcoord. C $\downarrow$ & 23.6 & 1.1 \\
Has overcoord. N $\downarrow$ & 1.5 & 0.0 \\
Has overcoord. H $\downarrow$ & 1.0 & 2.9 \\
Has undercoord. C $\downarrow$ & 60.0 & 65.4 \\
Has undercoord. N $\downarrow$ & 39.1 & 22.8 \\
Has undercoord. rare earth $\downarrow$ & 0.4 & 0.0 \\
Has metal $\uparrow$ & 100.0 & 100.0 \\
Has lone molecule $\downarrow$ & 72.9 & 80.5 \\
Has high charge $\downarrow$ & 0.9 & 0.5 \\
Has suspicious terminal oxo $\downarrow$ & 2.6 & 0.3 \\
Has undercoord. alkali $\downarrow$ & 1.0 & 0.3 \\
Has geom. exposed metal $\downarrow$ & 7.0 & 1.8 \\
\midrule
Validity rate (all passed) $\uparrow$ & 15.7 & 15.1
\\
\bottomrule
\end{tabular}
}%
\end{table}


\begin{table}[t]
\centering
\caption{\textbf{Molecule generation results on OMol25.}
\label{appendix:table:omol25_dng}
Validity, uniqueness, and \% pass rates on PoseBusters for 16,000 sampled molecules.
\textsc{Zatom-1}, pretrained on the (non-equilibrium) molecules of OMol25, takes a first step toward non-equilibrium 3D molecule generation.
}
\vspace{0.5em}
\begin{tabular}{lc}
\toprule
\textbf{Metric (\% pass) $\uparrow$} & \textbf{OMol25 \textsc{Zatom-1}} \\
\midrule
Validity & 30.4 \\
Uniqueness & 93.5 \\
Atoms connected & 17.0 \\
Bond angles & 78.9 \\
Bond lengths & 54.6 \\
Aromatic ring flat & 100.0 \\
Double bond flat & 100.0 \\
Internal energy & 73.5 \\
No steric clash & 60.2 \\
\midrule
PoseBusters valid & 15.1 \\
\bottomrule
\end{tabular}
\end{table}


\begin{table}[t]
\centering
\caption{\textbf{Property prediction for 3D molecules (QM9) / materials (Matbench) with joint molecule and materials finetuning.} Results are reported as \textit{test} mean absolute errors (for \textsc{Zatom-1}, with standard deviations over three runs with different random seeds). $\dagger$ denotes using different data partitions. / represents a task for which both (QM9) molecule and (Matbench) material property labels are available. NN denotes pretraining using a Noisy Nodes self-supervised objective \citep{zaidi2023pretraining}. The best (or second-best) results for each optimized (i.e., hyperparameter-tuned) or unoptimized (i.e., non-hyperparameter-tuned) category are in \textbf{bold} (or \underline{underlined}). Notably, finetuning jointly for multi-task molecule and ($\Delta\varepsilon$) materials property prediction improves or marginally trades off \textsc{Zatom-1}'s molecule property prediction performance for expanded applicability of its property predictions to both molecules and materials simultaneously.}
\label{appendix:table:qm9_matbench_finetuned_property_prediction_results}
\resizebox{\textwidth}{!}{
\begin{tabular}{llllllllllllll}
\toprule
\textbf{Model / Metrics} $\downarrow$ & 
\# Params & 
\textbf{\makecell{$\alpha$\\ ($a_0^3$)}} &
\textbf{\makecell{$\Delta\varepsilon$\\ (meV)}} &
\textbf{\makecell{$\varepsilon_{\mathrm{HOMO}}$\\ (meV)}} &
\textbf{\makecell{$\varepsilon_{\mathrm{LUMO}}$\\ (meV)}} &
\textbf{\makecell{$\mu$\\ (D)}} &
\textbf{\makecell{$C_\nu$\\ (cal/mol K)}} &
\textbf{\makecell{$G^{\mathrm{ATOM}}$\\ (meV)}} &
\textbf{\makecell{$H^{\mathrm{ATOM}}$\\ (meV)}} &
\textbf{\makecell{$R^2$\\ ($a_0^2$)}} &
\textbf{\makecell{$U^{\mathrm{ATOM}}$\\ (meV)}} &
\textbf{\makecell{$U_0^{\mathrm{ATOM}}$\\ (meV)}} &
\textbf{\makecell{$ZPVE$\\ (meV)}} \\
\midrule
\textbf{Optimized single-task learning} \\
DimeNet++$^\dagger$        & 5M & .044 & 33.0 / \textbf{199} & 25 & 20 & .030 & .023 & 8.00    & 7.00    & .331 & \underline{6.00}   & \underline{6.00}   & \underline{1.21} \\
EGNN$^\dagger$    & 1M & .071 & 48.0 / ---- & 29 & 25 & .029 & .031 & 12.0   & 12.0   & .106 & 12.0  & 11.0  & 1.55 \\
PaiNN               & 1M & .045 & 46.0 / ---- & 28 & 20 & .012 & .024 & \textbf{7.35} & \textbf{5.98} & \underline{.066} & \textbf{5.83} & \textbf{5.85} & 1.28 \\
TorchMD-NET   & 7M & .059 & 36.0 / ---- & 20 (16 w/ NN) & 18 (13 (w/ NN) & .011 & .026 & 7.62 & 6.16 & \textbf{.033} & 6.38 & 6.15 & 1.84 \\
SphereNet              & 2M & .046 & 32.0 / ---- & 23 & 18 & .026 & \textbf{.021} & 8.00    & \underline{6.00}    & .292 & 7.00   & 6.00   & \textbf{1.12} \\
SEGNN$^\dagger$ & 1M & .060 & 42.0 / ---- & 24 & 21 & .023 & .031 & 15.0   & 16.0   & .660 & 13.0  & 15.0  & 1.62 \\
EQGAT                   & - & .053 & 32.0 / ---- & 20 & 16 & .011 & .024 & 23.0   & 24.0   & .382 & 25.0  & 25.0  & 2.00 \\
Equiformer          & 4M & \underline{.046} & \underline{30.0} / ---- & \underline{15} & \underline{14} & \underline{.011} & \underline{.023} & 7.63 & 6.63 & .251 & 6.74 & 6.59 & 1.26 \\
EquiformerV2          & 11M & .050 & \textbf{29.0} / ---- & \textbf{14} & \textbf{13} & .010 & .023 & \underline{7.57} & 6.22 & .186 & 6.49 & 6.17 & 1.47 \\
P$\Theta$NITA   & - & \textbf{.038} & 30.4 / ---- & 16 & 15 & .012 & .024 & 8.63 & 8.04 & .235 & 8.67 & 8.31 & 1.29 \\
Platonic Transformer         & - & .049 & 37.4 / ---- & 22 & 17 & \textbf{.010} & .024 & 12.0 & 12.0 & .222 & 11.9 & 13.0 & 1.30 \\
\midrule
\textbf{Unoptimized single-task learning} \\
AIM-STL$^\dagger$         & - & .181 & ---- / ---- & 61 & 54 & \textbf{.067} & .072 & 66.2 & 63.9 & \textbf{.503} & 64.2 & 58.8 & \textbf{4.54} \\
\textbf{Unoptimized multi-task learning} \\
AIM-MTL-Scalar$^\dagger$         & - & .268 & ---- / ---- & 59 & 71 & .089 & .103 & 113 & 112 & 4.18 & 112 & 112 & 9.82 \\
AIM-MTL-Matrix$^\dagger$         & - & .251 & ---- / ---- & 61 & 72 & .088 & .103 & 112 & 118 & 4.07 & 116 & 116 & 12.2 \\
QM9-pretrained \textsc{Zatom-1} (\textsc{Ours})         & 80M \faSnowflake\ + 20M \faFire & \underline{.105$\pm$.000} & \underline{52.0$\pm$0.1} / \underline{362$\pm$1} & \underline{37$\pm$0.1} & \underline{36$\pm$0.1} & .102$\pm$0.001 & \underline{.054$\pm$0.000} & \underline{47.2$\pm$0.2}  & \underline{48.5$\pm$0.2} & 3.93$\pm$0.00 & \underline{48.6$\pm$0.2} & \underline{48.7$\pm$0.2} & 5.03$\pm$0.01 \\
\rowcolor{orange!30} Jointly pretrained \textsc{Zatom-1} (\textsc{Ours})         & 80M \faSnowflake\ + 20M \faFire & \textbf{.095}$\pm$\textbf{.000} & \textbf{46.2}$\pm$\textbf{0.1} / \textbf{354}$\pm$\textbf{1} & \textbf{34}$\pm$\textbf{0.1} & \textbf{32}$\pm$\textbf{0.1} & \underline{.087$\pm$.000} & \textbf{.047}$\pm$\textbf{.000} & \textbf{39.2}$\pm$\textbf{0.2} & \textbf{42.1}$\pm$\textbf{0.2} & \underline{3.33$\pm$0.01} & \textbf{41.6}$\pm$\textbf{0.2} & \textbf{41.6}$\pm$\textbf{0.2} & \underline{4.62$\pm$0.02} \\
\bottomrule
\end{tabular}}
\end{table}


\begin{table}[t]
\centering
\caption{\textbf{Energy and force prediction for 3D materials (MPtrj) / molecules (OMol25-4M).} The results are reported as \textit{validation} Energy (in meV or meV/Atom) and Force (in meV/\AA) mean absolute error (MAE). All models directly predict forces, rather than using the gradient of their predicted energies with respect to input 3D coordinates for conservative force prediction, to considerably increase their computational efficiency. * denotes a model that was pretrained on OMol25 and which has not converged after 80 training epochs.}
\label{appendix:table:mptrj_omol25_finetuned_mlip_results}
\resizebox{0.9\textwidth}{!}{
\renewcommand{\arraystretch}{1.15}
\begin{tabular}{lccccc}
\toprule
\textbf{Model} & \textbf{\# Params} & \textbf{Energy MAE/Atom (meV)} $\downarrow$ & \textbf{Energy MAE (meV)} $\downarrow$ & \textbf{Force MAE (meV/\AA)} $\downarrow$ \\
\midrule
\multicolumn{5}{c}{\textbf{Material Interatomic Potential Prediction on MPtrj}} \\
\midrule
 Non-pretrained Orb-v1 & $\sim$30M & - & 350.00 & 122.00 \\
 Denoising-pretrained Orb-v1 & $\sim$30M & - & 210.00 & 32.00 \\
 Jointly pretrained \textsc{Zatom-1} & 80M \faSnowflake\ + 220M \faFire & 112.81 & 2631.22 & 30.92 \\
\midrule
\multicolumn{5}{c}{\textbf{Molecule Interatomic Potential Prediction on OMol25-4M}} \\
\midrule
 eSEN-md-d. & $\sim$51M & 1.32 & 77.13 & 6.78 \\
  Molecule-only MLIP-pretrained \textsc{Zatom-1}* & $\sim$93M & 7.13 & 514.00 & 86.99 \\
  Molecule-only (generatively-)pretrained \textsc{Zatom-1}* & 80M \faSnowflake\ + 220M & 19.46 & 1307.23 & 72.40 \\
 Jointly pretrained \textsc{Zatom-1} & 80M \faSnowflake\ + 220M \faFire & 34.45 & 2512.78 & 62.99 \\
\bottomrule
\end{tabular}}
\end{table}

\begin{table}[h]
\centering
\small
\caption{QM9 molecule generation results after 20 epochs of QM9 multi-task property prediction finetuning.}
\label{appendix:table:qm9_dnmg_finetuning_results}
\begin{tabular}{lcccc}
\toprule
\textbf{Fine-Tuning Strategy} & \textbf{Validity} $\uparrow$ & \textbf{Uniqueness} $\uparrow$ & \textbf{Novelty} $\uparrow$ & \textbf{PoseBusters Valid} $\uparrow$ \\ 
\midrule
Trunk Unfreezing         & 0.17 & 0.33 & 0.33 & 0.00 \\
LoRA                     & 0.48 & 1.00 & 1.00 & 0.61 \\
\rowcolor{orange!30} Trunk Freezing (Baseline)& 0.95 & 0.97 & 0.96 & 0.99 \\ 
\bottomrule
\end{tabular}
\end{table}

\begin{table}[h]
\centering
\small
\caption{Comprehensive GEOM-Drugs molecule generation results. -- denotes unavailable results, and * indicates results were estimated from 1,000 samples per \citet{vonessen2026tabasco}.}
\label{appendix:table:geom_dnmg_comprehensive_results}
\begin{tabular}{lcccc}
\toprule
\textbf{Metric} & \textbf{ADiT} & \textbf{TABASCO} & \textbf{\textsc{Zatom-1} (Ours)} & \textbf{GEOM-Drugs} \\ 
\midrule
Validity $\uparrow$         & 0.95    & 0.98 & 0.94    & --   \\
Uniqueness $\uparrow$       & 1.00    & 0.99          & 1.00    & --   \\
Diversity $\uparrow$        & 0.91* & 0.89          & 0.89    & 0.90 \\
Novelty $\uparrow$          & 0.97* & 0.99 & 0.98    & 0.00 \\
QED $\uparrow$              & --      & 0.64          & 0.64    & 0.67 \\
Lipinski $\uparrow$         & --      & 4.90          & 4.80    & 5.00 \\
LogP $\uparrow$             & --      & 2.83          & 2.77    & 2.92 \\
Strain Energy $\downarrow$  & 46.36* & 17.50 & 22.58   & --   \\
PoseBusters Valid $\uparrow$      & 0.85    & 0.92          & 0.94    & 0.94 \\
\bottomrule
\end{tabular}
\end{table}

\subsection{Exploring metal-organic framework generation}
\cref{appendix:table:mof_generation_results} investigates \textsc{Zatom-1}'s ability to generate metal-organic frameworks (MOFs), a novel class of materials with several promising applications \citep{kitagawa2014metal}. With minimal hyperparameter tuning, we find that \textsc{Zatom-1} can match the MOF generation validity rate of ADiT despite the numerous chemical and geometric constraints of MOFs, such as requiring a balance of metal, carbon, and hydrogen atoms without steric clashes. Nonetheless, the model (as currently evaluated) has not converged, indicating that its proportion of successfully generated MOFs could likely be improved with further training.

\subsection{Stepping towards non-equilibrium 3D molecule generation} 
\cref{appendix:table:omol25_dng} studies \textsc{Zatom-1}'s ability to generate non-equilibrium molecules, like those contained in the OMol25 dataset. With minimal hyperparameter tuning and only 80 epochs of OMol25-based pretraining (due to compute resource constraints), we find that \textsc{Zatom-1} establishes a strong first step towards non-equilibrium 3D molecule generation. However, we note that it is currently unclear whether using the default settings for RDKit and the PoseBusters software suite is an appropriate way of assessing non-equilibrium molecule validity and realism, e.g., as PoseBusters' unit tests have likely been tuned using a different distribution of 3D molecules. Nonetheless, with additional training and hyperparameter/metric tuning, these results could likely be improved, which we leave for future work.

\subsection{Balancing multi-task molecule and material property prediction}
\cref{appendix:table:qm9_matbench_finetuned_property_prediction_results} presents the results of performing a second stage of finetuning for multi-task molecule and ($\Delta\varepsilon$) materials property prediction using QM9 and Matbench, starting from \textsc{Zatom-1}'s first stage of QM9 finetuning for multi-task molecule property prediction. While improving or marginally exchanging \textsc{Zatom-1}'s molecule property prediction performance, these results demonstrate that (generative) jointly pretrained \textsc{Zatom-1}'s \textit{multi-task} property predictions can effectively be applied to both 3D molecules and materials, to our knowledge a \textit{first-of-its-kind} result. In contrast, for materials property prediction in \cref{table:qm9_matbench_property_prediction_results} of the main text, we report zero-shot $\Delta\varepsilon$ results for the model weights performing best specifically for multi-task molecule property prediction, to illustrate the broad applicability of \textsc{Zatom-1}'s pretrained molecule and material embeddings.

\subsection{Extension to energies and forces}
\label{appendix:extension_to_energies_and_forces}
\cref{appendix:table:mptrj_omol25_finetuned_mlip_results} explores \textsc{Zatom-1}'s ability to predict energies and forces to serve as an MLIP. Through this preliminary set of experiments, we find that the performance of jointly pretrained \textsc{Zatom-1}'s material force predictions for MPtrj slightly surpasses that of Orb-v1, a state-of-the-art, pretrained model for this dataset. Results for OMol25 demonstrate that jointly pretrained \textsc{Zatom-1} can balance the performance of its force predictions for molecules and materials jointly, highlighting its potentially broad applicability in molecular and materials sciences. Despite these promising first steps, \textsc{Zatom-1}, when pretrained solely as an MLIP using OMol25, achieves considerably better energy predictions for OMol25 than jointly pretrained \textsc{Zatom-1}, suggesting the presence of negative (energy-specific) transfer learning between jointly pretrained \textsc{Zatom-1}'s small-sized QM9 and MP20 pretraining examples and OMol25's much larger pretraining molecules (n.b., as the energy of an atomic system is a function of the system's size). To characterize this observation further, using the generative modeling weights studied in Table \ref{appendix:table:omol25_dng}, we generatively pretrained \textsc{Zatom-1} and finetuned MLIP task-specific weights using OMol25 end-to-end, finding that \textit{generative} pretraining with OMol25 leads to better force predictions than \textit{MLIP} pretraining/finetuning with OMol25 (though energy predictions are improved with the latter approach). As a potential future direction, \textit{joint} molecule and material generative pretraining (beyond 80 epochs \citep{levine2025open}, up to convergence) using more structurally diverse datasets, such as LeMat-Bulk and OMol25, could improve \textsc{Zatom-1}'s ability to balance energy and force predictions effectively; we defer this to future work due to compute resource constraints.

\subsection{Finding an optimal method of finetuning}
In our initial approach to finetuning, we tried unfreezing \textsc{Zatom-1}'s entire pretrained Transformer trunk for downstream tasks, which led to significant generative performance degradation. Subsequently, we tried addressing this issue by adopting Low-Rank Adaptation (LoRA) for Transformer trunk finetuning \citep{hu2022lora}. In this setting, we used rank $r=8$, $\alpha=16$, and a scaling factor of $\frac{\alpha}{r}$, targeting the Transformer trunk's query, key, value, and output projection layers (totaling 500k parameters). As shown in Table \ref{appendix:table:qm9_dnmg_finetuning_results}, whole-trunk unfreezing results in catastrophic failure regarding 3D molecule generation validity as verified by RDKit/PoseBusters. Although LoRA remains significantly more robust in this regard than trunk unfreezing, we find that both trunk unfreezing and LoRA show further generative performance decay over extended finetuning. These findings support our current hypothesis: maintaining generative integrity while performing representation learning requires finetuning downstream task-specific weights rather than modifying the pretrained Transformer trunk.

\subsection{Comprehensively examining GEOM-Drugs molecule generation performance}
Table \ref{appendix:table:geom_dnmg_comprehensive_results} reports GEOM-Drugs molecule generation results for 10,000 samples across several additional metrics: Diversity and Novelty \citep{vonessen2026tabasco}; Quantitative Estimate of Drug-likeness (QED), Lipinski’s Rule of 5, and Crippen’s LogP \citep{buttenschoen2025an}; and PoseCheck’s Strain Energy \citep{harris2023posecheck}. As shown in Table \ref{appendix:table:geom_dnmg_comprehensive_results}, \textsc{Zatom-1} generates the most physically realistic molecules, matching the GEOM-Drugs reference dataset in PoseBusters validity (0.94). While TABASCO exhibits a slight advantage in chemical realism—specifically within Validity, Lipinski, and LogP scores—ADiT lags significantly in physical realism. Further, both \textsc{Zatom-1} and TABASCO achieve competitive Strain Energy bounds compared to ADiT. Overall, these results demonstrate that \textsc{Zatom-1} excels at producing physically realistic, drug-like molecules with chemical properties that closely mirror those of the GEOM-Drugs dataset.



\end{document}